\documentclass{article}

\usepackage{arxivgeneric}
\newif\ificlrfinal\iclrfinaltrue

\usepackage[utf8]{inputenc} 
\usepackage[T1]{fontenc}    
\usepackage{hyperref}       
\usepackage{url}            
\usepackage{booktabs}       
\usepackage{amsfonts}       
\usepackage{nicefrac}       
\usepackage{microtype}      
\usepackage{xcolor}         
\usepackage{graphicx}
\usepackage{tcolorbox}
\tcbuselibrary{breakable}

\newtcolorbox{nodecontext}[1][]{
    breakable,
    colback=red!6!white,        
    colframe=red!25!white,      
    coltitle=black,             
    boxrule=0.3pt, arc=2pt,
    left=6pt, right=6pt, top=4pt, bottom=4pt,
    fontupper=\small, #1
}

\newtcolorbox{excerpt}[1][]{
    breakable,
    colback=gray!15,
    colframe=gray!50,
    coltitle=black,             
    boxrule=0.4pt, arc=2pt,
    left=6pt, right=6pt, top=4pt, bottom=4pt,
    fontupper=\small, #1
}

\newtcolorbox{disputecontext}[1][]{
    breakable,
    colback=blue!6!white,
    colframe=blue!25!white,
    coltitle=black,
    boxrule=0.3pt, arc=2pt,
    left=6pt, right=6pt, top=4pt, bottom=4pt,
    fontupper=\small, #1
}

\newtcolorbox{initialbelief}[1][]{
    breakable,
    colback=red!8!white,
    colframe=red!30!white,
    coltitle=black,
    boxrule=0.3pt, arc=2pt,
    left=6pt, right=6pt, top=4pt, bottom=4pt,
    fontupper=\small, #1
}

\newtcolorbox{finalbelief}[1][]{
    breakable,
    colback=violet!6!white,
    colframe=violet!25!white,
    coltitle=black,
    boxrule=0.3pt, arc=2pt,
    left=6pt, right=6pt, top=4pt, bottom=4pt,
    fontupper=\small, #1
}

\newcommand{\Dref}[1]{\textcolor{teal!70!black}{\textbf{[#1]}}}       
\newcommand{\Aref}[1]{\textcolor{teal!70!black}{\textbf{[#1]}}}       
\newcommand{\Eref}[1]{\textcolor{teal!70!black}{\textbf{[#1]}}}       
\newcommand{\Cref}[1]{\textcolor{orange!70!black}{\textbf{[#1]}}}     
\newcommand{\Xref}[1]{\textcolor{red!60!black}{\textbf{[#1]}}}        
\newcommand{\Uref}[1]{\textcolor{red!60!black}{\textbf{[#1]}}}        

\usepackage{array} 
\usepackage{multirow}
\usepackage{bm}
\usepackage{float}
\usepackage{caption}
\usepackage{subcaption}
\usepackage{wrapfig}
\usepackage{comment}
\usepackage{graphicx}
\usepackage{tikz}
\usetikzlibrary{shapes.geometric, arrows.meta, positioning, fit, backgrounds, calc}
\usepackage{amsmath}
\usepackage{amsfonts}
\usepackage{amssymb}
\usepackage{enumerate}
\usepackage{lscape}
\usepackage{longtable}
\usepackage{rotating}
\usepackage{color}
\usepackage{url}
\usepackage{rotating}
\usepackage{algpseudocode}
\usepackage[ruled]{algorithm}
\usepackage{pgf,tikz}
\usepackage[titletoc]{appendix}
\usepackage{mathrsfs}
\usepackage{pgfplots}
\pgfplotsset{compat=1.18}
\usetikzlibrary{arrows}
\usepackage{color}
\usepackage{bbm}
\usepackage{eqparbox}%

\numberwithin{equation}{section}
\usepackage[symbol]{footmisc}

\newtheorem{definition}{Definition}

\newcommand{\beqn}[1]{\begin{equation}\label{#1}}
\newcommand{\eeqn}{\end{equation}}

\definecolor{darkgreen}{rgb}{0,0.6,0}
\definecolor{aau2}{rgb}{0.0, 0.5, 0.69}
\definecolor{aau3}{rgb}{0.0, 0.53, 0.74}
\definecolor{aau4}{rgb}{0.0, 0.48, 0.65}
\definecolor{aau5}{rgb}{0.0, 0.45, 0.73}
\definecolor{rsap}{RGB}{130, 36, 51}
\definecolor{gsap}{RGB}{112, 164, 137}
\definecolor{tud}{rgb}{0.43,0.73,0.11}
\definecolor{verde}{rgb}{0.33,0.53,0.11}
\definecolor{ttffqq}{rgb}{0.0, 0.48, 0.65}
\definecolor{ffqqqq}{rgb}{0.0, 0.5, 0.69} 
\definecolor{challcolor}{rgb}{0.55, 0.10, 0.15}   
\definecolor{defcolor}{rgb}{0.0, 0.20, 0.55}

\newcommand{\tcr}{\textcolor{red}}
\newcommand{\tcg}{\textcolor{darkgreen}}
\usepackage{caption}
\usepackage{subcaption}
\usepackage{float}
\usetikzlibrary{decorations.pathreplacing, hobby, calc}
\title{Multi-Agent Debate for Belief Optimization in Defeasible Domains}

\author{{\bf Tommaso Giovannelli}\\
University of Cincinnati\\
\texttt{giovanto@ucmail.uc.edu}
\and
{\bf Griffin D. Kent}\thanks{Primary corresponding author.}\\
Utumno Research Institute\\
\texttt{gdkent@utumno.ai}
}

\begin{document}

\maketitle

\begin{abstract}
            Multi-agent debate has emerged as a promising approach for improving LLM reasoning on ground-truth tasks, yet current methodologies face certain structural limitations: debate tends to induce a martingale over belief trajectories, majority voting accounts for most observed gains, and LLMs exhibit confidence escalation rather than calibration across rounds. We argue that the genuine value of debate, and dialectic systems as a whole, lies not in ground-truth tasks but in \emph{defeasible} domains, where every position can in principle be defeated by better reasoning. We present the \emph{Council of Hierarchical Agentic Language} (CHAL), a multi-agent dialectic framework that treats defeasible argumentation as an engine for belief optimization. Each agent maintains a \emph{CHAL Belief Schema} (CBS), a graph-structured belief representation with a Bayesian-inspired architecture, that facilitates belief revision through a gradient-informed dynamic mechanism by leveraging the strength of the belief's thesis as a differentiable objective. Meta-cognitive value systems spanning epistemology, logic, and ethics are elevated to configurable hyperparameters governing agent reasoning and adjudication outcomes. We provide a series of ablation experiments that demonstrate systematic and interpretable effects: the adjudicator's value system determines the debate's overall trajectories in latent belief space, council diversity refines beliefs for all participants, and the framework generalizes across broad fields. CHAL is, to our knowledge, the first framework to treat multi-agent debate as structured belief optimization over defeasible domains. Further, the auditable belief artifacts it produces establish the foundation for dedicated evaluation suites for defeasible argumentation, with broader implications for building AI systems whose reasoning and value commitments are transparent, aligned, and subject to human oversight.
\end{abstract}

\section{Introduction}\label{sec:introduction}

The history of artificial intelligence (AI) is, at its core, a history of attempts to formalize reasoning: to endow machines with the capacity to think through novel problems, weigh competing considerations, and arrive at justified conclusions. The arrival of large language models (LLMs), built on the transformer architecture~\cite{AVaswani_etal_2017}, has dramatically narrowed the gap between this aspiration and the available tools, with probabilistic next-token prediction at scale giving rise to emergent multi-step deduction and coherent long-form reasoning~\cite{WXZhao_etal_2023,SBubeck_etal_2023}. These capabilities have renewed interest in utilizing structured Multi-Agent Debate (MAD) to improve automated reasoning, an idea with roots predating the LLM era by decades~\cite{AHBond_LGasser_eds_1988,MWooldridge_2009,IRahwan_GSimari_2009}. The use of LLM-powered debate was first demonstrated in~\cite{YDu_etal_2023} which showed improvements in mathematical reasoning and factual question answering through iterative critique, and was extended by~\cite{TLiang_etal_2024} with a tit-for-tat framework incorporating adaptive termination and a dedicated judge agent. Yet critical examination has revealed structural limitations: standard inter-agent debate induces a martingale over belief trajectories, implying that argumentative exchange alone does not improve expected correctness without structural interventions that bias updates toward correction~\cite{HKChoi_JZhu_SLi_2025}. Empirical evidence further demonstrates that current LLMs exhibit confidence escalation rather than Bayesian calibration across debate rounds, a metacognitive deficit exacerbated by fine-tuning~\cite{PSPrasad_MNguyen_2025,JLeng_etal_2024}. Recent work has begun to address individual aspects of these limitations, including neuro-symbolic defeasible reasoning~\cite{XFang_etal_2025}, argumentation-enhanced claim verification~\cite{GFreedman_etal_2025}, and ethical parameterization of debate agents~\cite{JMaslowski_JAChudziak_2026}, but each targets a single facet in isolation; no existing framework integrates structured belief representations, configurable meta-cognitive value systems, and formal argumentation within a unified multi-agent debate system.

In this paper, we propose the \emph{Council of Hierarchical Agentic Language} (CHAL), a MAD framework that treats argumentation on defeasible topics as a process of \emph{belief optimization}. Drawing on Minsky's vision of a \emph{Society of Mind}~\cite{MMinsky_1986}, CHAL orchestrates a council of LLM agents, each reasoning from a distinct epistemological worldview, through a dialectical pipeline in which beliefs are genuinely tested, adversarially challenged, and structurally refined. In contrast to most existing debate systems where agents exchange unstructured text and belief states remain implicit, CHAL equips each agent with the \emph{CHAL Belief Schema} (CBS), a formally structured internal belief representation whose internal components and confidence assignments are explicit, auditable, and revised through a gradient-informed optimization mechanism that directly addresses the structural limitations identified in prior work, including LLMs' inability to reliably distinguish belief from knowledge and fact~\cite{MSuzgun_etal_2025}. The system further makes the meta-cognitive commitments that any reasoning system implicitly carries into explicit, configurable hyperparameters spanning epistemology, logic, and ethics, enabling systematic study of how value structures shape the outcomes of dialectical exchange. Beyond its role as a belief refinement engine, CHAL is designed as a modular \emph{research platform} with direct implications for AI safety, value-aligned reasoning, and the broader challenge of building AI systems whose reasoning is transparent and aligned with well-specified values~\cite{IGabriel_2020}. We provide CHAL as a publicly available software package, accessible via  \url{https://github.com/GdKent/CHAL}.

The principal contributions of this work are outlined as follows.

\textbf{The CHAL Belief Schema (CBS).} We introduce a formally structured belief representation in which each agent maintains a typed, auditable belief object that induces a directed acyclic dependency graph with strength-based confidence propagation. These beliefs replace the unstructured text exchange of existing debate systems with modular, auditable, and semantically rich objects.

\textbf{Gradient-informed belief revision.} We introduce a belief optimization mechanism in which the thesis strength serves as a differentiable objective over the CBS and its gradient drives targeted dynamic revision prompts, embedding a form of gradient ascent into the prompting layer.

\textbf{Meta-cognitive value systems as hyperparameters.} We introduce a suite of epistemic personas, logic systems, and ethical systems as independently configurable hyperparameters governing agent reasoning and adjudication, enabling study of how value structures shape dialectical outcomes.

\textbf{A modular research platform for value-aligned AI reasoning.} We present the complete dialectical pipeline of CHAL as a configurable research platform with implications for AI safety, value-aligned reasoning, and the empirical study of how meta-cognitive configurations influence reasoning on consequential questions.

The remainder of this paper is organized as follows. 
Section~\ref{sec:values} introduces the configurable meta-cognitive value systems, Section~\ref{sec:IBS} defines the CBS and belief dependency graph, Section~\ref{sec:CHAL} presents the complete CHAL framework and dialectical pipeline, Section~\ref{sec:experiments} provides numerical experiments under controlled ablations, and Section~\ref{sec:conclusion} concludes with directions for future research. Further, we also provide a comprehensive literature review in Appendix~\ref{app:lit_rev}. 

\section{Meta-Cognition and Value Hierarchies}\label{sec:values}

Any system capable of making claims typically must first answer three prior questions: what counts as valid inference, what counts as morally relevant, and what counts as knowledge~\cite{RCarnap_1950,NRescher_1985}. These questions have deep histories in the philosophy of logic~\cite{SHaack_1978,JCBeall_GRestall_2006}, meta-ethics~\cite{JLMackie_1977,MSmith_1994}, and epistemology~\cite{RAudi_2010}, and different answers produce genuinely different conclusions from identical evidence: a classical and a paraconsistent logician reach incompatible verdicts on contradictory premises, a utilitarian and a deontologist reach opposite normative judgments from the same facts, and an empiricist and a rationalist may disagree on whether a given belief constitutes knowledge at all. In CHAL, these three meta-cognitive choices, of epistemic position, reasoning system, and ethical system, are explicit configuration parameters that determine how agents form beliefs, how arguments are adjudicated, and how convergence is interpreted. They are, in the language of machine learning, the \emph{hyperparameters} of the debate: meta-level commitments that govern every first-order inference, and making them explicit allows the value commitments that any reasoning system implicitly carries to be studied, varied, and compared across configurations.

\subsection{Epistemic, Logical, and Ethical Systems}\label{sec:meta_config}

\begin{wrapfigure}{r}{0.33\textwidth}
\renewcommand{\thesubfigure}{\thefigure(\alph{subfigure})}
\makeatletter
\renewcommand{\p@subfigure}{}
\makeatother
    \centering
    \vspace{-1.5\baselineskip}
    \begin{subfigure}[t]{0.3\textwidth}
        \centering
        \resizebox{\linewidth}{!}{%
        \begin{tikzpicture}
          \fill[white] (0,0) rectangle (8,6);
          \draw[black, thick] (0,0) rectangle (8,6);
          \foreach \x/\y in {1.2/5.0, 3.5/4.6, 6.8/5.2,
                             0.8/3.2, 7.0/3.0,
                             1.0/1.2, 3.8/1.6, 6.5/1.8,
                             5.2/4.0} {
            \node[black, font=\Large\bfseries] at (\x,\y) {?};
          }
          \filldraw[black] (4.0,3.1) circle (4pt);
          \node[black, font=\Large, anchor=south west]
            at (4.05,3.1) {Agent};
        \end{tikzpicture}%
        }
        \vspace{-0.6\baselineskip}
        \caption{A single agent in the latent belief space.}
        \label{fig:single-agent}
    \end{subfigure}
    \\[0.1\baselineskip]
    \begin{subfigure}[t]{0.3\textwidth}
        \centering
        \resizebox{\linewidth}{!}{%
        \begin{tikzpicture}
          \fill[white] (0,0) rectangle (8,6);
          \draw[black, thick] (0,0) rectangle (8,6);
          \coordinate (A1) at (1.60, 2.77);
          \coordinate (A2) at (3.56, 4.85);
          \coordinate (A3) at (6.40, 3.88);
          \coordinate (A4) at (5.16, 1.85);
          \coordinate (A5) at (2.67, 0.92);
          \draw[black, dashed, thick, line width=0.8pt]
            plot[smooth cycle, tension=0.55]
            coordinates {(A1) (A2) (A3) (A4) (A5)};
          \node[black, align=center, font=\Large] at (3.73, 2.95)
            {Likely Feasible\\Belief Region};

          \filldraw[black] (A1) circle (4pt);
          \node[black, font=\Large, anchor=north east]
            at ($(A1)+(0.45,0)$) {Agent 1};

          \filldraw[black] (A2) circle (4pt);
          \node[black, font=\Large, anchor=south]
            at ($(A2)+(0,0.15)$) {Agent 2};

          \filldraw[black] (A3) circle (4pt);
          \node[black, font=\Large, anchor=south west]
            at ($(A3)+(-0.30,0.05)$) {Agent 3};

          \filldraw[black] (A4) circle (4pt);
          \node[black, font=\Large, anchor=north west]
            at ($(A4)+(0.1,-0.12)$) {Agent 4};

          \filldraw[black] (A5) circle (4pt);
          \node[black, font=\Large, anchor=north]
            at ($(A5)+(0,-0.15)$) {Agent 5};
        \end{tikzpicture}%
        }
        \vspace{-0.6\baselineskip}
        \caption{Multiple agents mapping the feasible belief region.}
        \label{fig:chal-council}
    \end{subfigure}
    \vspace{-1.0\baselineskip}
\end{wrapfigure}
An agent's epistemic position determines its standards for evidence, justification, and inference, including what kinds of reasons it takes to be compelling, how it weighs competing sources of knowledge, and where it locates the burden of proof. In CHAL, epistemic personas are assigned exclusively to the LLM council agents rather than the adjudicator (see Section~\ref{sec:CHAL} for formalism), whose evaluation is governed instead by independently configured logic and ethics systems. This separation is deliberate: the council's role is to generate and defend positions from genuinely distinct epistemological starting points, while the adjudicator's role is to evaluate the argumentative exchange against a fixed and consistent standard. By populating the council with agents reasoning from competing epistemological traditions, CHAL explores the belief space more broadly than any single framework would permit, as illustrated in Figures~\ref{fig:single-agent} and~\ref{fig:chal-council}.

CHAL currently provides twelve epistemic personas, eight logic systems, and six ethical systems, each grounded in established philosophical traditions. The default logic system, Classical~+~Informal~+~Bayesian (CIB), combines deductive validity with inductive and abductive support weighted by prior probability. Logic and ethics scores are computed independently by an adjudicator agent and combined via user-specified weights (see Section~\ref{sec:adjudicator}). The choice of system materially affects adjudication outcomes and is therefore a primary variable of interest for empirical studies of debate dynamics. Summary tables and complete descriptions for all epistemic positions as well as logical and ethical systems are provided in Appendix~\ref{app:meta_sys}.

\subsection{A Note on Implicit Values Embedded Within LLMs}

The value systems described in Appendices~\ref{app:logic}--\ref{app:ethics} are explicit and configurable, and a user may set any weight to zero to remove its direct influence on adjudication (see Section~\ref{sec:adjudicator}). However, setting a weight to zero does not render the corresponding dimension entirely absent from the model's outputs. LLMs acquire implicit values through their pretraining corpora~\cite{EMBender_etal_2021} and RLHF alignment~\cite{LOuyang_etal_2022}, producing empirically measurable biases in opinion distributions~\cite{SSanturkar_etal_2023} and prompt-sensitive reasoning patterns that depart from formal logic~\cite{ALampinen_etal_2024}. CHAL cannot eliminate these residual influences; rather, its prompting architecture steers each agent toward the explicitly assigned value systems to the extent that prompting allows. The goal is therefore not value-neutral reasoning, which no LLM-based system can guarantee, but an explicit and configurable value structure whose influence on the dialectic can be systematically studied and compared across configurations.

\section{Internal Belief Systems}\label{sec:IBS}

This section defines the data structures that underpin CHAL's belief optimization process. The central artifact is the \emph{CHAL Belief Schema} (CBS), a formally structured object representing an agent's complete belief state at any point in the debate. A CBS is generated at the opening of the debate, updated after every argumentative exchange, and compared across agents and rounds to measure epistemic progress.

\subsection{The CHAL Belief Schema}\label{sec:CBS}

The CBS integrates ideas from five theoretical frameworks into a single machine-readable, LLM-generatable structure: Toulmin's decomposition of argument into claim, grounds, warrant, and rebuttal~\cite{SToulmin_1958}; the ASPIC+ attack taxonomy~\cite{SModgil_HPrakken_2014}; the AGM minimality principle, implemented via the patch-based update mechanism in Section~\ref{sec:belief_rev}~\cite{CAlchourron_etal_1985}; Bayesian calibrated strength values on $[0,1]$ subject to propagation constraints~\cite{AHunter_2017,WTalbott_SEP_2022}; and Popperian falsifiability, operationalized through explicit predictions attached to every claim~\cite{KPopper_1959}.

\begin{definition}[CHAL Belief Schema]\label{def:CBS}
Let $\Theta$, $\Delta$, $\Lambda$, $\Xi$, $\Omega$, $\Psi$, and $\Phi$ denote
the spaces of all well-formed thesis, definition, assumption, evidence, claim,
counterposition, and uncertainty objects, respectively. A CBS, otherwise referred to as a belief object, is a tuple
\[
  \mathscr{B} \;=\; \left(\, \theta,\; \mathcal{D},\; \mathcal{A},\;
  \mathcal{E},\; \mathcal{C},\; \mathcal{X},\; \mathcal{U}\,\right),
\]
where $\theta \in \Theta$ is the thesis, and $\mathcal{D} \subseteq \Delta$,
$\mathcal{A} \subseteq \Lambda$, $\mathcal{E} \subseteq \Xi$,
$\mathcal{C} \subseteq \Omega$, $\mathcal{X} \subseteq \Psi$, and
$\mathcal{U} \subseteq \Phi$ are finite sets of definitions, assumptions,
evidence items, claims, counterpositions, and uncertainties, respectively.
\end{definition}

Every element carries a prefixed identifier (D\#, A\#, E\#, C\#, X\#, U\#) enabling precise referencing during cross-examination, adjudication, and revision. Among the seven components, five are \emph{strength-bearing}: $\theta$, $\mathcal{D}$, $\mathcal{A}$, $\mathcal{E}$, and $\mathcal{C}$ each carry a calibrated strength value in $[0,1]$ subject to the dependency constraints defined in Section~\ref{sec:BDG}. The remaining two, $\mathcal{X}$ and $\mathcal{U}$, are \emph{non-strength-bearing} and instead serve as adversarial and epistemic honesty signals (see Sections~\ref{sec:BDG} and~\ref{sec:belief_rev}). A summary of each component's role is given in Table~\ref{tab:cbs_components}, and full field-level specifications are provided in Appendix~\ref{app:CBS_components}. The shared strength calibration scale that is used to ground agents is provided in Appendix~\ref{app:strength_scale}.

\subsubsection{The Strength-Infused Bayesian Architecture}\label{sec:strength_arch}

Each strength-bearing node ($\theta$ or elements from the sets $\mathcal{D}$, $\mathcal{A}$, $\mathcal{E}$, and $\mathcal{C}$ ) carries an agent-assigned strength value in $[0,1]$, interpretable as a Bayesian credence, together with a justification. In isolation, self-reported strengths would invite inflation; under CHAL's adversarial cross-examination, however, inflated strengths become targets for attack, and surviving values are progressively refined toward accurate reflections of the underlying epistemic state across rounds. The strength architecture is therefore not merely an annotation but one of the primary signals agents reason over when deciding how to refine their position (see Section~\ref{sec:belief_rev}) and what positions to attack.

The thesis strength is the quantitative objective each agent works to maximize, playing a role analogous to a reward function in reinforcement learning. Unlike node-level strengths, it is computed deterministically from the active claim set and is never agent-chosen.

\begin{definition}[Thesis Strength]\label{def:thesis_str}
    Given a belief object $\mathscr{B}$, let $\mathcal{C}^{\!*} \subseteq \mathcal{C}$ denote the set of active (non-retracted) claims, with $n = |\mathcal{C}^{\!*}|$, and let $s_{c_i} \in [0,1]$ denote the strength of the $i$-th active claim $c_i\in\mathcal{C}^{\!*}$. Then the \emph{thesis strength} $s_\theta\in[0,1]$ is defined as
    \begin{equation}\label{eq:thesis_str}
        s_\theta \;:=\; \underbrace{\left(\,\frac{1}{n} \sum_{i=1}^{n} s_{c_i}\,\right)}_{\text{average claim strength } \bar{s}} \cdot \underbrace{\left(\frac{n^p}{n^p+1}\right)}_{\text{breadth multiplier } B(n)},
    \end{equation}
    where $p > 0$ is the breadth sensitivity exponent (default $p = 1$). Treating $s_\theta$ as a function of $\bar{s}$ and $n$ (i.e., $s_\theta:[0,1]\times[0,1] \rightarrow[0,1]$), its gradient $\nabla s_\theta\in\mathbb{R}^2$ is given by
    \begin{equation}\label{eq:partials}
      \nabla s_\theta (\,\bar{s},\; n;\; p\,) \;:=\; \left[\,\frac{\partial s_\theta}{\partial \bar{s}},\;\frac{\partial s_\theta}{\partial n}\,\right]^\top \;=\; \left[\,B(n),\;\;\bar{s} \cdot \frac{p \, n^{\,p-1}}{(n^p+1)^2}\,\right]^\top.
    \end{equation}
\end{definition}

The formula~\eqref{eq:thesis_str} decomposes $s_\theta$ into two factors: the average active claim strength $\bar{s}$ and the breadth multiplier $B(n)$. Since $B(n)$ is monotonically increasing with $B(1) = 1/2$, a thesis resting on a single claim cannot exceed $s_\theta = 0.5$ regardless of that claim's strength, mechanically rewarding broader positions. Both components of $\nabla s_\theta$ are nonnegative; their relative magnitudes determine whether strengthening existing claims or adding new ones yields the larger marginal gain, and CHAL uses this information to drive the gradient-informed refinement described in Section~\ref{sec:belief_rev}.

\subsection{The Belief Dependency Graph}\label{sec:BDG}

The CBS provided in Definition~\ref{def:CBS} specifies the components of a belief but not the relationships between them. Each definition, assumption, evidence item, and claim declares a set of components it supports and depends on, and each counterposition and uncertainty declares a set of components it targets. Taken together, these declared relationships induce a directed acyclic graph over the elements of $\mathscr{B}$, which we refer to as the \emph{Belief Dependency Graph} (BDG) and which governs both strength propagation and aid in adversarial targeting within the belief.

\begin{definition}[Belief Dependency Graph]\label{def:BDG}
Given a belief object $\mathscr{B}$, the BDG is a directed acyclic graph $G_{\mathscr{B}} = (\mathcal{N}, \mathcal{R})$, composed of a nodes (vertices) set $\mathcal{N}$ and relations (edges) set $\mathcal{R}$, where
\[
  \mathcal{N} \;=\; \{\theta\} \,\cup\, \mathcal{D} \,\cup\, \mathcal{A} \,\cup\, \mathcal{E} \,\cup\, \mathcal{C} \,\cup\, \mathcal{X} \,\cup\, \mathcal{U},
\]
and $\mathcal{R}$ partitions into two edge types: \emph{strength-bearing} edges (edges between strength-bearing nodes) from $\mathcal{D}$ into $\mathcal{A} \cup \mathcal{E}$, from $\mathcal{A} \cup \mathcal{E} \cup \mathcal{C}$ into $\mathcal{C}$, and from active claims $c_i \in \mathcal{C}^{\!*}$ for all $i\in\{1,2,...,n\}$ into $\theta$; \emph{non-strength-bearing} edges from $\mathcal{X} \cup \mathcal{U}$ into $\mathcal{D} \cup \mathcal{A} \cup \mathcal{E} \cup \mathcal{C}$.
\end{definition}

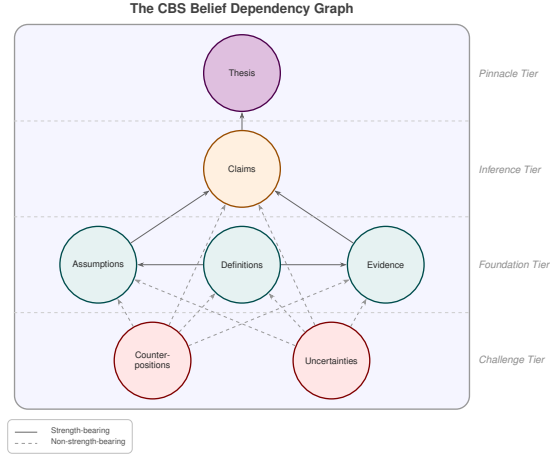
\begin{wrapfigure}{r}{0.6\textwidth}
    \centering
    \resizebox{0.75\linewidth}{!}{

\begin{tikzpicture}[
    foundation node/.style={
        circle, draw=teal!60!black, very thick,
        fill=teal!10, minimum size=2.4cm,
        font=\sffamily\large, align=center, text=black!85
    },
    claim node/.style={
        circle, draw=orange!60!black, very thick,
        fill=orange!12, minimum size=2.4cm,
        font=\sffamily\Large, align=center, text=black!85
    },
    thesis node/.style={
        circle, draw=violet!60!black, very thick,
        fill=violet!25, minimum size=2.4cm,
        font=\sffamily\Large, align=center, text=black!85
    },
    challenge node/.style={
        circle, draw=red!50!black, very thick,
        fill=red!10, minimum size=2.4cm,
        font=\sffamily\large, align=center, text=black!85
    },
    set label/.style={
        font=\sffamily\huge\bfseries, text=black!75
    },
    tier label/.style={
        font=\sffamily\Huge\itshape, text=black!45, align=left
    },
    support arrow/.style={
        -{Stealth[length=6pt, width=4pt]}, thick, black!60
    },
    challenge arrow/.style={
        -{Stealth[length=6pt, width=4pt]}, thick, black!40, dashed
    },
]

\node[thesis node] (thesis) at (0, 4.5) {Thesis};

\node[claim node] (claims) at (0, 1.5) {Claims};

\node[foundation node] (assume) at (-4.5, -1.5) {Assumptions};
\node[foundation node] (defn)   at ( 0.0, -1.5) {Definitions};
\node[foundation node] (evid)   at ( 4.5, -1.5) {Evidence};

\node[challenge node] (counter) at (-2.8, -4.5) {Counter-\\positions};
\node[challenge node] (uncert)  at ( 2.8, -4.5) {Uncertainties};

\draw[challenge arrow] (counter) -- (assume);
\draw[challenge arrow] (counter) -- (defn);
\draw[challenge arrow] (counter) -- (claims);
\draw[challenge arrow] (counter) -- (evid);

\draw[challenge arrow] (uncert) -- (assume);
\draw[challenge arrow] (uncert) -- (defn);
\draw[challenge arrow] (uncert) -- (claims);
\draw[challenge arrow] (uncert) -- (evid);

\draw[support arrow] (defn) -- (assume);
\draw[support arrow] (defn) -- (evid);

\draw[support arrow] (assume) -- (claims);
\draw[support arrow] (evid)   -- (claims);

\draw[support arrow] (claims) -- (thesis);

\begin{scope}[on background layer]
    \node[fit=(counter)(uncert)(assume)(defn)(evid)(claims)(thesis),
          rectangle, rounded corners=12pt,
          draw=black!40, very thick,
          fill=blue!4,
          inner xsep=1.4cm, inner ysep=0.3cm] (bbox) {};
\end{scope}

\node[set label, anchor=south] at ([yshift=4pt] bbox.north) {The CBS Belief Dependency Graph};

\draw[thin, black!20, dashed]
    (bbox.west |- 0,3) -- (bbox.east |- 0,3);
\draw[thin, black!20, dashed]
    (bbox.west |- 0,0) -- (bbox.east |- 0,0);
\draw[thin, black!20, dashed]
    (bbox.west |- 0,-3) -- (bbox.east |- 0,-3);

\node[tier label, anchor=west] at ([xshift=4pt] bbox.east |- 0,4.5)
    {Pinnacle \\ Tier};
\node[tier label, anchor=west] at ([xshift=4pt] bbox.east |- 0,1.5)
    {Inference \\ Tier};
\node[tier label, anchor=west] at ([xshift=4pt] bbox.east |- 0,-1.5)
    {Foundation \\ Tier};
\node[tier label, anchor=west] at ([xshift=4pt] bbox.east |- 0,-4.5)
    {Challenge \\ Tier};

\node[anchor=north west, inner sep=6pt,
      draw=black!25, rounded corners=4pt, fill=white,
      font=\sffamily\large, text=black!70]
    (legend) at ([xshift=0.2cm, yshift=-0.2cm] bbox.north west)
    {\begin{tabular}{@{}ll@{}}
        \tikz\draw[thick, black!60] (0,0) -- (0.7,0); & Strength-bearing \\[2pt]
        \tikz\draw[thick, black!40, dashed] (0,0) -- (0.7,0); & Non-strength-bearing
    \end{tabular}};

\end{tikzpicture}%
}
    \caption{The four-tiered structure of the Belief Dependency Graph.}
    \label{fig:ABO}
\end{wrapfigure}

An illustration of the BDG as well as the relationships among the nodes is displayed in Figure~\ref{fig:ABO}, also highlighting the four-tiered structure. The \emph{foundation tier} contains definitions, assumptions, and evidence, where definitions act as strength ceilings on the nodes that reference them. The \emph{inference tier} contains active claims, each depending on foundation-tier components. The \emph{pinnacle tier} contains the thesis, whose strength is determined via~\eqref{eq:thesis_str}. The \emph{challenge tier} contains counterpositions and uncertainties, whose non-strength-bearing edges target nodes in the tiers above. The tiered structure of the BDG also induces a chain of dependency constraint equations coupling node strengths across the foundation and inference tiers. These constraints are defined as follows.

\begin{definition}[Strength Dependency Constraints]\label{def:dep_constraint}
Let $s_v\in[0,1]$ denote the strength of a node $v$, for all $v \in \mathcal{D} \cup \mathcal{A} \cup \mathcal{E} \cup \mathcal{C}$. For each active node $v \in \mathcal{A} \cup \mathcal{E} \cup \mathcal{C}$, let $\mathrm{dep}(v)$ denote the set of active (non-retracted) dependencies of $v$: its supporting definitions when $v \in \mathcal{A} \cup \mathcal{E}$, and its supporting assumption, evidence, or other claim nodes when $v \in \mathcal{C}$. Then $s_v$ satisfies the constraint
\begin{equation}\label{eq:d_ceiling}
  s_v \;\leq\; \min_{u \,\in\, \mathrm{dep}(v)} s_u.
\end{equation}
That is, no nodes in the set $\mathcal{D} \cup \mathcal{A} \cup \mathcal{E} \cup \mathcal{C}$ are stronger than their weakest active dependency. When $\mathrm{dep}(v) = \emptyset$, then $v$ has no supporting dependency nodes and $s_v$ is upper-bounded at a strict orphan threshold $s_\mathrm{orph} \in [0,1]$.
\end{definition}

The thesis strength is then determined from $\mathcal{C}^{\!*}$ via~\eqref{eq:thesis_str}, closing the constraint chain from foundation to pinnacle. These constraints are enforced after every belief update (see Section~\ref{sec:belief_rev}), so modifying a foundation-tier node cascades upward through~\eqref{eq:d_ceiling} and~\eqref{eq:thesis_str} to revise the thesis strength automatically. This cascade makes foundation-tier attacks strategically valuable during cross-examination and informs how agents target their own revisions. Node strengths can also increase (see Section~\ref{sec:pipeline}).

\subsection{The Belief Revision System}\label{sec:belief_rev}

Belief updates in CHAL are carried out exclusively through a deterministic \emph{patch system} that implements the AGM minimality principle~\cite{CAlchourron_etal_1985} operationally. Rather than regenerating a belief from scratch, an agent emits atomic, structured patch operations targeting specific nodes, which are applied by a central routine that also enforces~\eqref{eq:d_ceiling} (see Section~\ref{sec:pipeline}).

\begin{definition}[Belief Patch]\label{def:patch}
A belief patch $\pi$ is an atomic operation on a CBS belief object $\mathscr{B}$ of one of the two following forms: \emph{add} a new node of type $\mathcal{D}, \mathcal{A}, \mathcal{E}, \mathcal{C}, \mathcal{X}$, or $\mathcal{U}$; \emph{update} an existing node's fields.
\end{definition}

Given a CBS object $\mathscr{B}$ and an ordered sequence of $k\in\mathbb{N}$ patches $\Pi = (\pi_1, \dots, \pi_k)$ emitted by an agent, the patch application routine applies each $\pi_j$ and re-enforces the constraint equations~\eqref{eq:thesis_str} and~\eqref{eq:d_ceiling}. Retraction is handled by forcing the strength of the retracted node to $0$ and excluding it from all subsequent constraint calculations. The propagation step is implemented as a breadth-first traversal of the BDG starting from modified nodes, ensuring that any strength change cascades consistently through the layered structure before the thesis strength is recomputed.

\subsubsection{Gradient-Informed Belief Refinement}\label{sec:grad_prompting}

\begin{figure}[htbp]
    \centering
    \includegraphics[width=1.\linewidth]{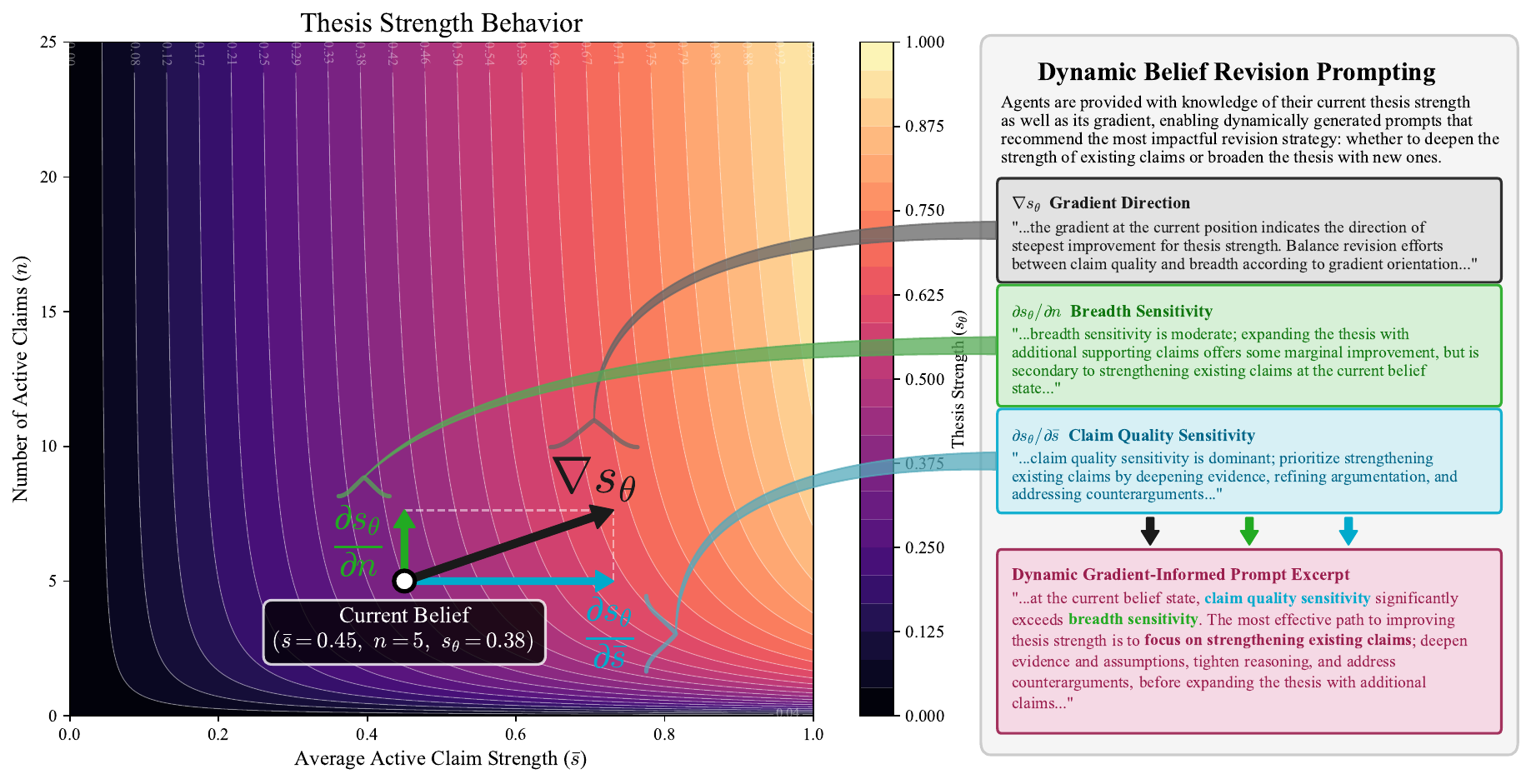}
    \caption{Illustration of the gradient-informed refinement strategy.}
    \label{fig:gradient_informed_prompting}
\end{figure}

When an agent revises its belief, CHAL evaluates $\nabla s_\theta$ at the agent's current values $(\bar{s}, n)$ and projects the gains from four concrete interventions: raising average claim strength by a fixed increment, adding a claim at the current average, adding a claim above the current average, and retracting the weakest active claim. These projections, together with the identities of the lowest-strength dependencies and any orphaned nodes, are injected into the revision prompt via a \emph{position analysis} block, giving the agent a quantitative description of which strategies are locally most effective (see Figure~\ref{fig:gradient_informed_prompting}).

Together with the patch system and the adversarial mechanisms of cross-examination and adjudication (see Section~\ref{sec:CHAL}), gradient-informed refinement produces a cycle in which weak arguments are culled, surviving arguments are strengthened, and new material is added along the most productive local direction. Across rounds, this drives each belief toward a configuration in which node strengths reflect the adversarial record rather than the agent's willingness to assert confidence.

\section{CHAL}\label{sec:CHAL}

Having established the formal structure of the core belief object in Section~\ref{sec:IBS}, we now turn to the explicit system architecture of CHAL itself: the multi-agent framework that orchestrates the iterative refinement of these belief objects through structured adversarial dialectical exchange.

\subsection{Council Formalism}\label{sec:council_form}

The core design of CHAL is centered around the \emph{council}, a finite collection of LLM-powered agents (referred to as council members) that each maintain a CBS belief object and reason from a distinct \emph{epistemic persona} drawn from Appendix~\ref{app:epistemic_positions}. We formalize it as follows.

\begin{definition}[The Council]\label{def:council}
Let $\odot_{\mathscr{B}_i^{(r)}}^i$ denote some LLM-based reasoning agent (a \emph{council member}) that is parameterized by a CBS belief object $\mathscr{B}_i^{(r)}$, given the current debate round $r\in\{0,1,...,R\}$ where $R\in\mathbb{N}$ is the total number of rounds, and for all $i\in\{1,2,...,N\}$ where $N\in\mathbb{N}$ is the total number of council members. Then the state of the council of $N$ agents at round $r$ is defined as
\[
  \bigodot_{\!r} \;:=\; \left\{\, \odot_{\mathscr{B}_i^{(r)}}^i \,\right\}_{i=1}^N \; = \; \left\{\, \odot_{\mathscr{B}_1^{(r)}}^1 \,,\, \odot_{\mathscr{B}_2^{(r)}}^2 \,,\, ... \,,\, \odot_{\mathscr{B}_N^{(r)}}^N \,\right\}.
\]
Each initial belief $\mathscr{B}_i^{(0)}$ is induced by a topic query $q \in \mathbb{T}$ and an epistemic persona $\beta^i \in \mathbb{T}$ defined in Appendix~\ref{app:epistemic_positions}, where $\mathbb{T}$ is a textual space, i.e., $(q, \beta^i) \;\Rightarrow\; \mathscr{B}_i^{(0)}$. All other beliefs $\mathscr{B}_i^{(r)}$, for $r\neq0$, are obtained via the pipeline defined in Subsection~\ref{sec:pipeline}.
\end{definition}

The council progressively refines each agent's belief object through pairwise adversarial exchange. In each round $r$, every unique pair $(\odot^i, \odot^j) \in \bigodot_{\!r} \times \bigodot_{\!r}$, with $i \neq j$, produces up to $S\in\mathbb{N}$ \emph{challenge-rebuttal pairs}, yielding $S \cdot N \cdot (N-1)$ and $R \cdot S \cdot N \cdot (N-1)$ pairs per round and across the entire debate, respectively. The complete set of pairs forms the input to the adjudicator (Subsection~\ref{sec:adjudicator}), whose verdicts feed back into the belief revision system (Subsection~\ref{sec:belief_rev}).

\subsection{The Adjudicator}\label{sec:adjudicator}

The adjudicator is an independent LLM-based agent that drives belief evolution by evaluating every challenge-rebuttal pair and rendering a verdict. Its role is deliberately constrained to evaluation: it does not generate arguments, maintain a belief of its own, or access any council member's persona, so that verdicts reflect the substance of the arguments rather than the identity of the arguer. Without this separation between generation and evaluation, LLM debaters typically default to premature consensus rather than genuine adversarial testing~\cite{TLiang_etal_2024}. We formalize the adjudicator as follows.

\begin{definition}[The Adjudicator]\label{def:adjudicator}
Let $\mathbb{X}$ and $\mathbb{D}$ denote the spaces of well-formed challenge and rebuttal objects, respectively, and let $\mathcal{V} := \{\texttt{\,critique\_valid},\, \texttt{rebuttal\_valid},\, \texttt{unresolved}\,\}$ denote the set of admissible verdicts. The \emph{adjudicator} is an LLM-based agent $\blacklozenge_\vartheta : \mathbb{X} \times \mathbb{D} \longrightarrow \mathcal{V}$, where $\vartheta=(L,\,E,\,w_L,\,w_E,\,\tau)$ define the parameters of the adjudicator, including a logic system $L\in\mathbb{T}$ from Appendix~\ref{app:logic}, an ethics system $E\in\mathbb{T}$ from Appendix~\ref{app:ethics}, respective logic and ethics weights $(w_L, w_E) \in [0,1]\times[0,1]$ satisfying $w_L + w_E = 1$, and a decision threshold $\tau \in [0,1]$, where $\mathbb{T}$ is a textual space. For each challenge-rebuttal pair $(\chi, \rho) \in \mathbb{X} \times \mathbb{D}$, the adjudicator computes weighted scores
\[
  \sigma^c \;:=\; w_L \cdot \sigma_L^c + w_E \cdot \sigma_E^c, \qquad
  \sigma^d \;:=\; w_L \cdot \sigma_L^d + w_E \cdot \sigma_E^d,
\]
where $\sigma_L^c, \sigma_E^c, \sigma_L^d, \sigma_E^d \in [0,1]$ are the logic and ethics scores assigned to the challenger and defender by the adjudicator agent under the meta-cognitive value systems $(L, E)$, and renders the verdict
\[
  \blacklozenge_\vartheta(\chi, \rho) \;=\; \begin{cases}
    \texttt{critique\_valid} & \text{if } \sigma^c - \sigma^d \geq \tau, \\
    \texttt{rebuttal\_valid} & \text{if } \sigma^d - \sigma^c \geq \tau, \\
    \texttt{unresolved} & \text{otherwise}.
  \end{cases}
\]
\end{definition}

Varying the meta-cognitive systems $(L, E)$ and weighting parameters $(w_L, w_E, \tau)$ across debates is one of the primary research configurations enabled by CHAL, allowing systematic study of how different logical and normative commitments shape the dialectical record under otherwise identical conditions. The logic and ethics scoring scales used by the adjudicator are provided in Appendix~\ref{app:scoring}.

\subsection{The Dialectical Pipeline}\label{sec:pipeline}

The complete CHAL framework is realized as an ordered sequence of six stages (numbered 0--5) that we collectively refer to as the \emph{dialectical pipeline}. CHAL is model-agnostic; an initial configuration layer defines the topic query, model provider, and all pipeline and agent-specific parameters. Stages~0 and~1 establish the initial conditions and instantiate the council, while Stages~2 through~5 form the inner loop that repeats for each of the $R$ debate rounds. The full algorithmic specification is provided in Appendix~\ref{app:CHAL_algorithm} and Figure~\ref{fig:pipeline} illustrates the pipeline. We now provide an overview of each stage.

\textbf{Stage~0 (Briefing).} Each council member is initialized with a \emph{universal debate prompt} establishing shared reasoning rules, calibration norms, and the expectation that intellectual honesty is valued over rhetorical victory. Each agent then receives a \emph{epistemic persona} instantiating its assigned worldview from Appendix~\ref{app:epistemic_positions}, framed as a lens for analysis rather than a set of conclusions to defend.

\begin{wrapfigure}{r}{0.42\textwidth}
    \centering
    \vspace{-0.1\baselineskip}
    \resizebox{0.40\textwidth}{!}{\input{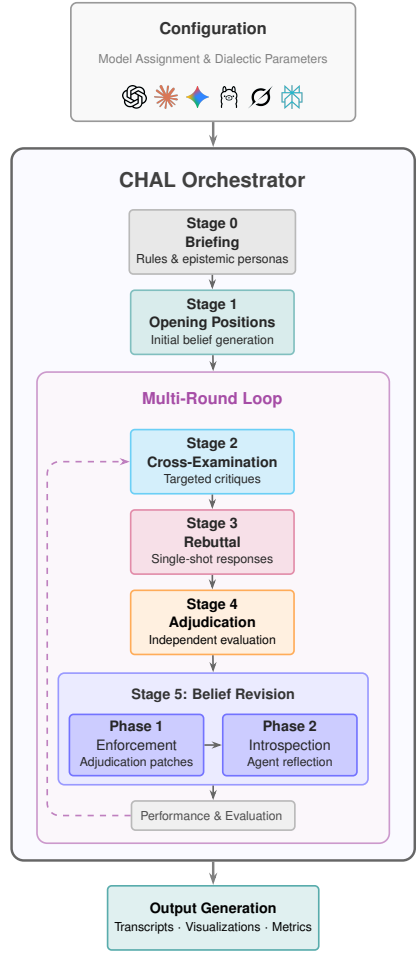}}
    \caption{The CHAL dialectical pipeline.}
    \label{fig:pipeline}
    \vspace{-2.5\baselineskip}
\end{wrapfigure}
\textbf{Stage~1 (Opening Positions).} Each agent generates an initial CBS belief object $\mathscr{B}_i^{(0)}$ in response to the topic query $q$, constructed bottom-up from foundational components through inferential claims to the thesis. The generated belief is validated against the CBS schema and the BDG is checked for structural integrity before the debate proceeds.

\textbf{Stage~2 (Cross-Examination).} For each ordered pair $(\odot^i, \odot^j)$ with $i \neq j$, the challenger $\odot^i$ generates up to $S$ targeted challenges against the current belief $\mathscr{B}_j^{(r-1)}$ of the defender $\odot^j$. A complete list of all allowed attack strategies along with their descriptions is provided in Appendix~\ref{app:attack_taxonomy}.

\textbf{Stage~3 (Rebuttal).} Each defender produces a structured rebuttal to every challenge, committing to one of three actions: \textbf{refute}, \textbf{concede}, or \textbf{defer}. Defenders also propose a series of tentative belief revision patches that do not modify the belief until Stage~5.

\textbf{Stage~4 (Adjudication).} All challenge-rebuttal pairs are submitted to the adjudicator $\blacklozenge_\vartheta$, which renders a verdict from $\mathcal{V}$ via a sub-protocol of disagreement restatement, formalization, reasoning, and adjudication.

\textbf{Stage~5 (Belief Revision).} Belief revision proceeds in two phases. In \emph{Phase~1 (Enforcement)}, each agent applies mandatory patches: defenders add at least one weakening patch per \texttt{critique\_valid} verdict, a new uncertainty node per \texttt{unresolved} verdict, and a defense boost per \texttt{rebuttal\_valid} verdict (see Appendix~\ref{app:def_boost}). In \emph{Phase~2 (Introspection)}, the agent receives the gradient-informed analysis of Section~\ref{sec:grad_prompting} and may add new components, resolve counterpositions and uncertainties, and retract weak claims, subject to a strength filter that prevents unilateral increases to existing node strengths. The thesis is rewritten last to reflect all of the changes to the agent's positions, its strength recomputed via~\eqref{eq:thesis_str}, and the resulting $\mathscr{B}_i^{(r)}$ becomes the input to the next round.

After all $R$ rounds, the system produces the complete debate transcript, each agent's final belief $\mathscr{B}_i^{(R)}$, trajectory visualizations, and per-agent performance statistics.

\section{Numerical Experiments}\label{sec:experiments}

Recent meta-analyses have shown that on tasks with definitive ground-truth answers, multi-agent debate often fails to outperform simple baselines such as majority voting~\cite{HKChoi_JZhu_SLi_2025} and single-agent chain-of-thought~\cite{HZhang_etal_2025}, raising the question of where debate is genuinely valuable. We contend that the answer lies in defeasible domains, where no ground-truth answer exists and every position can in principle be defeated by better reasoning. CHAL is designed for this regime, and the CBS ensures that belief evolution is observable, trackable, and auditable through structured artifacts. No existing benchmark for defeasible reasoning, argumentation mining, or belief revision evaluates multi-agent, graph-structured belief evolution through adversarial dialectics; we therefore adopt ablation-based evaluation to demonstrate that CHAL's behavior is systematic, interpretable, and sensitive to its architectural parameters in the intended ways. We also contend that the structured output artifacts that CHAL produces provide the foundation upon which dedicated benchmarks for defeasible argumentation could be constructed and is a direction for future work, noted in Section~\ref{sec:conclusion}.

All experiments use OpenAI's o4-mini for all council members and the adjudicator. Unless otherwise noted, each experiment generates $n = 10$ independent debate trajectories per configuration with shared initial beliefs and run for $R = 5$ rounds. Beliefs are embedded via Appendix~\ref{app:embeddings} and projected into two dimensions via UMAP. Per-agent performance is measured by thesis strength and the Agent Performance Score (APS) (see Appendix~\ref{app:APS}). Due to space limitations, complete discussions of each experiment, including figures, statistical tables, and excerpts, are provided in Appendix~\ref{app:experiments}, which also provides a discussion on the initial underlying epistemic distributions that each agent draws from (see Appendix~\ref{app:topology}). A complete example of a set of initial and final beliefs that are generated by CHAL is provided in Appendix~\ref{app:belief_ex}. We now highlight the type of ablations as well as the primary findings.

\subsection{Ablations}\label{sec:ablation}

\textbf{Adjudicator Ablation.} This ablation isolates the effect of the adjudicator's meta-cognitive value system on belief evolution. We test three configurations with two council members (Empiricist and Supernaturalist) across 10 trajectories on the topic ``Does free will exist?'': (1) pure logic (CIB), (2) balanced (CIB + Deontological), and (3) pure ethics (Deontological). Under pure logic and balanced adjudication, the Empiricist dominates as the Supernaturalist's metaphysical claims fail falsifiability challenges; under pure ethics, the outcome reverses entirely, with the Supernaturalist favored despite presenting similar arguments. This highlights the adjudicator as the primary force driving where the debate converges and what is constituted as the most defensible position under a given evaluative standard. Detailed analysis is provided in Appendix~\ref{app:adj_ablation_details}.

\textbf{Council Member Ablation.} This ablation isolates the effect of council diversity on belief dynamics. We compare a 2-agent council (Empiricist and Supernaturalist) with a 3-agent council (adding a Phenomenologist), both under balanced adjudication (CIB + Deontological) again on the topic ``Does free will exist''. The 3-agent configuration produces tighter final belief distributions for all agents, including those already present in the smaller council, by attacking both existing positions from a methodologically distinct angle. This supports the central idea that populating the council with epistemically diverse agents refines the belief landscape for all participants, as each agent is exposed to a broader range of argumentative pressures. Detailed analysis is provided in Appendix~\ref{app:council_ablation_details}.

\textbf{Topic Ablation.} To assess generalization across philosophical domains, we provide an single-trajectory demonstration across three topics: ``Does free will exist?'' (Empiricist vs. Supernaturalist; pure logic with CIB), ``Should AI have rights?'' (Panpsychist vs. Skeptic; balanced with CIB + Rule-Utilitarianism), and ``Is it ethical to genetically modify future generations?'' (Constructivist vs. Rationalist; balanced with CIB + Rule-Utilitarianism). As these results are primarily qualitative, detailed analyses and excerpts are presented in Appendix~\ref{app:topic_ablation_details}.

\section{Conclusion}\label{sec:conclusion}

We have presented CHAL, a MAD framework that treats argumentation on defeasible topics as a process of belief optimization. The CBS replaces unstructured text exchange with a formally typed, graph-structured belief representation with explicit Bayesian-inspired strength assignments. A dynamic gradient-informed  mechanism drives belief revision in a strategic way, embedding a form of learning into CHAL at the prompting layer. Meta-cognitive commitments spanning epistemology, logic, and ethics are elevated to configurable hyperparameters within a modular, extensible pipeline. Ablation experiments confirm that these choices produce systematic effects: the adjudicator's value system determines where the debate converges, council diversity refines the belief landscape for all participants, and the framework generalizes across a wide array of highly-debated topics. Lastly, we provide a discussion on the current limitations of CHAL in Appendix~\ref{app:limitations}.

CHAL opens several directions for future research. The adjudicator's meta-cognitive parameterization provides a controlled apparatus for studying how value systems shape conclusions on contested questions, with direct relevance to value alignment. Iterating epistemically diverse councils through adversarial exchange can yield thoroughly tested positions on enduring questions whose structured artifacts could serve as dedicated benchmarks for future defeasible argumentation systems. Alternative debate modes, from collaborative truth-seeking to competitive formats, can be integrated and empirically compared, and CHAL's performance can be further tuned through prompt-level optimization of the pipeline and meta-cognitive definitions. The hierarchical adjudicator-agent structure suggests natural extensions to bilevel optimization of the meta-cognitive systems. Finally, the belief object and meta-cognitive value system paradigms could be integrated into general AI systems, equipping them with auditable, structured beliefs and values subject to human oversight.

\bibliography{refs}



\bigskip
\textbf{\large Appendices}

\appendix

\section{Literature Review}\label{app:lit_rev}

The intellectual landscape that CHAL inhabits is broad and far-ranging. The system simultaneously draws on ideas from the theory of large language models, agentic AI systems, multi-agent debate, formal and informal argumentation theory, the philosophy of belief and knowledge, theories of logic and ethics, and hierarchical optimization. In this appendix, we survey each of these areas below, tracing the key developments that highlight where CHAL rests in the broader landscape.

\subsection{Modern AI}

\noindent\textbf{The Emergence of Large Language Models.} The origin of the modern LLM era is typically traced to the transformer
architecture~\cite{AVaswani_etal_2017}, a foundational breakthrough that replaced recurrent processing with pure self-attention, enabling substantial parallelization in training and allowing models to scale the number of parameters by orders of magnitude. The authors of~\cite{JKaplan_etal_2020} quantified the resulting scaling laws, demonstrating smooth power-law improvements on language tasks as model size and data increased, and establishing that capability gains were reliably predictable. The transformer sits at the center of virtually all modern neural language architectures. Encoder-based models such as BERT~\cite{JDevlin_etal_2019}, RoBERTa~\cite{YLiu_etal_2019}, ALBERT~\cite{ZLan_etal_2020}, and their variants excel at producing dense semantic representations of text across diverse modalities~\cite{ADosovitskiy_etal_2021, WNHsu_etal_2021, IBeltagy_KLo_ACohan_2019}. However, it was the line of autoregressive decoder models, beginning with GPT~\cite{ARadford_etal_2018} and refined through GPT-2~\cite{ARadford_etal_2019}, GPT-3~\cite{TBrown_etal_2020}, InstructGPT~\cite{LOuyang_etal_2022}, and GPT-4~\cite{openai_2023}, that first demonstrated truly emergent natural language understanding and reasoning capabilities from probabilistic next-token prediction alone. This generation of models, along with comparable efforts including Anthropic's Claude~\cite{anthropic_2023}, Google's Gemini~\cite{GeminiTeam_2023}, Meta's LLaMA~\cite{HTouvron_etal_2023}, DeepSeek AI's DeepSeek~\cite{ALiu_etal_2024}, and Alibaba's QWen~\cite{AYang_etal_2024}, exhibits emergent generalizability across an increasingly broad class of complex problems.

\noindent\textbf{Advanced Reasoning in LLMs.} The scaling of model size alone does not fully explain the reasoning capabilities that have emerged in recent systems. A complementary line of research has explored how structured prompting strategies can systematically unlock and improve LLM reasoning. The foundational insight came from \emph{Chain-of-Thought} (CoT) prompting~\cite{JWei_etal_2022}, which demonstrated that by including explicit intermediate reasoning steps in a few-shot prompt, models achieve dramatic performance gains on multi-step mathematical and symbolic reasoning tasks that they fail when prompted for direct answers. Crucially, the authors of~\cite{TKojima_etal_2022} subsequently showed that this behavior can be elicited in a zero-shot setting simply by appending the phrase ``Let's think step by step,'' revealing that the capacity for structured reasoning is latent in pre-trained models and can be triggered without examples. Self-Consistency decoding~\cite{XWang_etal_2022} extended these ideas by sampling multiple diverse reasoning chains and selecting the most consistent final answer, effectively ensembling different reasoning trajectories to improve reliability. Least-to-Most prompting~\cite{DZhou_etal_2022} further improved performance on compositional tasks by teaching models to decompose complex queries into ordered sub-problems, solving them sequentially. Tree-of-Thought~\cite{SYao_etal_2023} generalized the linear chain structure to a tree-search over reasoning steps, allowing explicit backtracking and evaluation of intermediate states. These developments helped collectively establish a productive paradigm: rather than treating LLMs as static lookup systems, they are better understood as latent reasoning engines whose behavior can be shaped through carefully engineered prompts that instantiate particular cognitive processes.

\noindent\textbf{Modern Agentic AI.}
Building on the observation that LLMs can engage in structured internal reasoning, researchers have extended them into agentic systems capable of interacting with tools, environments, and other agents over extended time horizons. A foundational contribution was the ReAct framework~\cite{SYao_etal_2022}, which introduced a paradigm of alternating between reasoning traces and external actions (e.g., querying search engines, executing API calls, reading documents, etc.) embedding the agent in an interactive observe-reason-act loop that substantially reduced hallucinations and improved decision quality. Reflexion~\cite{NShinn_etal_2023} added an explicit verbal reinforcement learning mechanism: after each episode, the agent reflects on its failures and encodes the reflection as a natural-language memory that conditions future behavior, creating a recursive self-improvement loop without gradient-based training. HuggingGPT~\cite{YShen_etal_2023} demonstrated how a general-purpose LLM can serve as a central planning agent that decomposes complex user requests, routes sub-tasks to domain-specific expert models from a tool repository, and synthesizes their outputs into a coherent response. AutoAgent~\cite{XLiu_etal_2025} pushed these ideas further with infrastructure for creating and evolving LLM-powered agents via natural language alone, requiring no code from the user. Dynamic systems capable of autonomous task-handling, including AutoGPT~\cite{autogpt_2023}, MetaGPT~\cite{SHong_etal_2024}, LangChain~\cite{langchain_2023}, and AutoGen~\cite{QWu_etal_2023}, have built on these foundations to support long-horizon planning, tool use, and persistent memory. The common thread across these systems is a shift from static text generation to interactive cognition, in which LLMs function as planners, decision-makers, and autonomous problem-solvers embedded in broader environments. For extensive surveys on this rapidly developing field,
see~\cite{ZDurante_etal_2024,LWang_etal_2023,YSu_etal_2024,ZXi_etal_2023,ZZhang_etal_2024,MYu_etal_2025,BLiu_etal_2025}, and for a position paper on embodied Agent AI, see~\cite{QHuang_etal_2024}.

\subsection{Multi-Agent Systems}

\noindent\textbf{Minsky's Society of Mind and the Foundations of Multi-Agent AI.} The philosophical roots of multi-agent AI run deeper than the LLM era. In his seminal 1986 work \emph{The Society of Mind}~\cite{MMinsky_1986}, Marvin Minsky argued that human intelligence is not the product of any single unified process, but rather emerges from the coordinated interaction of many simpler, specialized sub-agents, which he termed ``societies of agents'', each competent within a narrow domain and each incapable alone of producing the rich cognitive behavior they collectively enable. Minsky's framework was deeply influential in cognitive science and AI, providing an early conceptual template for architectures in which intelligence is a collective, emergent property of interacting components rather than a monolithic capability. The multi-agent systems community took up this challenge over subsequent decades, developing formal models of agent interaction, coordination, and communication under the frameworks of distributed artificial intelligence~\cite{AHBond_LGasser_eds_1988} and later multi-agent systems~\cite{GWeiss_2013}. These early systems, however, relied on hand-coded agent behaviors and formal communication protocols; they lacked the flexible, open-ended reasoning capabilities required to engage in genuinely productive epistemic exchange. The arrival of LLM-powered agents which are capable of taking on arbitrary roles, beliefs, personas, and communication styles in natural language has breathed new life into Minsky's vision, enabling multi-agent systems in which the ``agents'' are genuine reasoners rather than rule-following automata.

\noindent\textbf{LLM-Powered Multi-Agent Frameworks.} With the capabilities of LLM-powered agents established, a productive line of research has explored how to compose them into collaborative multi-agent systems for complex tasks. Recent frameworks like CAMEL~\cite{GLi_etal_2023} and ChatDev~\cite{CQian_etal_2024} demonstrate how multiple LLM agents with specialized roles (e.g., coder, tester, manager, domain expert, etc.) can collaborate through structured dialogue to tackle tasks such as software development and scientific exploration. The authors of~\cite{JHPark_etal_2023} simulate long-lived AI individuals in a sandbox environment, revealing lifelike emergent patterns including social memory, planning, and unsupervised coordination. S-Agents~\cite{JChen_etal_2024} explore how self-organizing LLM agents in open-ended environments can autonomously allocate tasks within a tree-structured organization led by a leadership agent, without human intervention or predefined workflows. Formal approaches like ECON~\cite{CXie_etal_2025} formulate multi-agent reasoning as a Bayesian game in which agents explicitly model each other's beliefs and anticipated responses, achieving equilibrium without constant pairwise interaction. Meta-frameworks such as MetaGPT~\cite{SHong_etal_2024} and ChatLLM~\cite{RHao_etal_2023} treat agent teams as composable systems in which diverse LLMs or prompts act in orchestrated workflows, improving coordination and parallelism. Research on LLMs in repeated games~\cite{EAkata_etal_2025} has revealed that such agents perform well in self-interested games such as the iterated Prisoner's Dilemma but coordinate poorly in games such as the Battle of the Sexes. For surveys on LLM-based multi-agent systems, see~\cite{TGuo_etal_2024,KTTranetal_2025,SFeng_etal_2025}.

\noindent\textbf{Multi-Agent Debate (MAD) Systems.} Perhaps the most important sub-field which CHAL resides in the midst of is the use of multi-agent architectures to improve reasoning through \emph{debate}~\cite{IRahwan_GSimari_2009, MWooldridge_2009, GIrving_PChristiano_DAmodei_2018}. Although renewed interest in the idea of multi-agent debate systems has grown substantially since the advent of LLMs, this surge was initiated by~\cite{YDu_etal_2023} which demonstrated that LLM-powered multi-agent debate can yield meaningful improvements on mathematical reasoning, factual question answering, and strategic tasks, with agents capable of identifying and eliminating each other's hallucinations through iterative critique. Further, \cite{TLiang_etal_2024} proposed a tit-for-tat debate system to elicit divergent thinking on tasks that require deep deliberation, introduced an adaptive termination mechanism, and a dedicated judge agent to synthesize debate outcomes. The authors of~\cite{AKhan_etal_2024} showed that debate between expert LLMs enables non-expert judges to identify correct answers at substantially higher accuracy than naive baselines, and that optimizing debaters for persuasiveness further improved judge accuracy. Work has also been done to curate debate datasets that aim to help provide benchmarking capabilities for certain kinds of debate systems~\cite{YSChuang_etal_2025}. However, there are several structural limitations that have been identified across the field. The authors of~\cite{AEstornell_YLiu_2024} highlighted that when agents share similar capabilities or training distributions, debates tend to collapses toward majority opinion, with shared misconceptions being reinforced rather than corrected as more agents are added. The authors of~\cite{HKChoi_JZhu_SLi_2025} showed that argumentative exchange alone has difficulty improving expected correctness without structural interventions, and that majority voting accounts for most gains typically attributed to debate. The works of~\cite{PSPrasad_MNguyen_2025,JLeng_etal_2024} further noticed that models tend to exhibit confidence escalation across rounds rather than calibrating in response to counterarguments, which is further exacerbated by RLHF fine-tuning. More recently, efforts to address specific aspects of these limitations have emerged. The LLM-ASPIC+ framework~\cite{XFang_etal_2025} integrates LLMs with the ASPIC+ structured argumentation framework to perform defeasible reasoning, but operates as a single-agent neuro-symbolic system rather than a multi-agent debate architecture. The authors of~\cite{GFreedman_etal_2025} augment LLMs with formal argumentation frameworks for explainable claim verification, using defeasible inference to produce transparent rationales, though again without a multi-agent dialectical structure. The Heterogeneous Debate Engine~\cite{JMaslowski_JAChudziak_2026} is the most closely related recent work to CHAL, configuring debate agents with distinct ethical schools and equipping each with identity representations that assign certainty values to natural language nodes; however, it is restricted to ethical parameterization within a single pedagogical domain and does not incorporate epistemological diversity, alternative logic systems, a formal belief graph structure with confidence propagation, independent adjudication, or gradient-informed optimization, which limits the framework to identity-preserving dialogue rather than truth-oriented structured belief optimization with auditable revision trajectories.

\subsection{Multi-Level and LLM Optimization}

\noindent\textbf{Latent Prompt Optimization.} Research on the optimization of prompts for LLMs has evolved along several parallel tracks. Gradient-based soft-prompt methods~\cite{XLi_PLiang_2021,BLester_etal_2021,XLiu_etal_2023} demonstrated that LLMs can be adapted to downstream tasks by learning small trainable embeddings prepended to inputs via backpropagation through a frozen model. Gradient-free approaches have also achieved comparable quality for continuous LLM prompts~\cite{TSun_etal_2022,ZFei_etal_2023}. Pseudo-gradient methods~\cite{RPryzant_etal_2023} derived optimization signals from model feedback, iteratively refining natural-language prompts via beam search. Approaches combining gradient and gradient-free methods~\cite{CHang_etal_2023} have further unified these paradigms in black-box LLM settings. Context compression~\cite{TGe_etal_2023} addressed the related problem of reducing inference cost by learning compact summary representations of long contexts. For thorough overviews of the current landscape, see~\cite{SHuang_etal_2024,BLiu_etal_2025}.

\noindent\textbf{Multi-Level Optimization.} Multi-level optimization~(MLO) is a class of problems in which an upper-level objective is optimized subject to the constraint that subsets of variables satisfy optimality conditions for nested sub-problems. These hierarchical structures arise naturally whenever decision-makers operate at different levels of authority or on different timescales. Applications have been developed in defense~\cite{BArguello_ESJohnson_JLGearhart_2023,KLai_MIllindala_KSubraminiam_2019,YYao_etal_2007,XWu_AJConejo_2017,YGuo_CGuo_JYang_2023}, signal recovery and power control~\cite{HLiduka_2011}, supply chain networks~\cite{XXu_ZMeng_RShen_2013,MRahdar_etal_2018,AFard_MKeshteli_SMirjalili_2018}, and machine learning~\cite{YJiao_etal_2024,SKeun_etal_2023,MGuo_etal_2020,YJiao_KYang_CJian_2024,HLiu_KSimonyan_YYang_2019,XJin_etal_2019,TGiovannelli_GDKent_LNVicente_2025_bilevel,TGiovannelli_GDKent_LNVicente_2025}. Bilevel optimization is the simplest non-trivial sub-class of these problems, in which a single upper-level problem is constrained by a single lower-level optimal response, and has received particularly intensive study in the ML context, with applications to meta-learning, data poisoning, neural architecture search, and hyperparameter optimization.
Gradient-based methods for bilevel optimization are generally organized according to how the bilevel stochastic gradient~(BSG) direction used to update the upper-level variables is approximated~\cite{RLiu_JGao_etal_2021}. Two main families have emerged. The first, known as \textit{implicit differentiation}, computes the~BSG direction via the implicit function theorem, either by solving the adjoint equation directly~\cite{FPedregosa_2016} or by approximating the inverse Hessian through a truncated Neumann series~\cite{JLorraine_PVicol_DDuvenaud_2019}. The second, known as \textit{iterative differentiation}, propagates gradients through the unrolled dynamics of the lower-level~(LL) solver using automatic differentiation~\cite{LFranceschi_etal_2018}; however, this approach tends to be less computationally efficient than implicit differentiation~\cite{KJi_JYang_YLiang_2020}. Within the implicit differentiation family, a sequence of works has progressively improved convergence guarantees and algorithmic efficiency~\cite{NCouellan_WWang_2015,NCouellan_WWang_2016,SGhadimi_MWang_2018,MHong_etal_2020,TChen_YSun_WYin_2021_closingGap,DSow_KJi_YLiang_2021,TGiovannelli_GDKent_LNVicente_2025_bilevel} (see also~\cite{RLiu_JGao_etal_2021,CChen_etal_2022} for surveys). The constrained~LL setting has received comparatively little attention. Recent work has begun to address this gap: \cite{PKhanduri_etal_2023} develops smoothed implicit gradient methods for linearly constrained LL problems; \cite{HShen_TChen_2023} and \cite{WYao_etal_2024} propose penalty-based approaches applicable beyond the linear case; \cite{TGiovannelli_GDKent_LNVicente_2025_bilevel} provides a unified convergence theory covering both unconstrained and nonlinearly constrained LL problems via inexact adjoint gradient computations.

\noindent\textbf{Conjunction of LLMs, Multi-Agent Systems, and Multi-Level Optimization.} The intersection of LLM-powered multi-agent systems with multi-level optimization represents a nearly unexplored frontier. The most closely related existing work has been in data weighting and selection when training LLMs, which has been formulated as a bilevel optimization problem~\cite{YYu_etal_2025,RPan_etal_2025}. Hierarchical multi-level formulations for multi-agent coordination have been studied in works whose agents are robots, human workers, and reinforcement learning agents rather than LLM-based reasoners~\cite{CZhang_JAShah_2016,HUppaluru_HRastgoftar_2023,HZhang_etal_2020}. The authors of~\cite{YWu_etal_2024} use a bilevel formulation for multi-agent tracking and allocation in which the two levels are optimized by separate LLMs, but this is less a true multi-level optimization problem than a decomposition of a single-level problem. To the best of our knowledge, no prior work has proposed a genuine multi-level optimization framework for the latent belief spaces of LLM-powered multi-agent systems. For a thorough review of how optimization methods have been applied in the presence of LLMs, see~\cite{SHuang_etal_2024}.

\subsection{Argumentation Theory}

\textbf{Roots of Argumentation.} The formal study of argumentation predates AI by millennia with the first systematization of inference forms being introduced in Aristotle's \emph{Topics} and \emph{Rhetoric}~\cite{JBarnes_ed_1984}, which distinguished \emph{apodeixis} (demonstrative proof from certain premises) from \emph{dialectic} (reasoning from accepted opinions) and \emph{rhetoric} (persuasion for practical audiences). The distinction between deductive certainty and defeasible reasoning from plausible premises is foundational to all subsequent argumentation theory. The modern revival of informal logic and argumentation theory began with Toulmin's \emph{The Uses of Argument}~\cite{SToulmin_1958}, which rejected the adequacy of formal deductive logic for evaluating everyday reasoning and proposed a six-component model of argument structure: \emph{claim} (the conclusion), \emph{grounds} (the evidence), \emph{warrant} (the inference rule connecting grounds to claim), \emph{backing} (justification for the warrant), \emph{qualifier} (the strength of the conclusion), and \emph{rebuttal} (conditions that would defeat the inference). This laid the foundation of \emph{defeasible argumentation}, in which conclusions are supported by reasons that can always be defeated by stronger counter-reasons, and which lies in contrast to pure deductive reasoning. Toulmin's model has proven especially relevant to LLM-based argumentation, where recent NLP work has found that critical questions inspired by this model improve LLM performance on MT-Bench reasoning and math tasks, on average, over both standard and chain-of-thought prompting~\cite{FCastagna_etal_2024}. Among argumentation theorists in philosophy, the term ``argument'' typically refers to the giving of reasons to support or criticize a claim that is questionable or open to doubt~\cite{DWalton_2006}, a definition that explicitly distinguishes argumentation from purely deductive mathematical inference, in which conclusions follow necessarily from premises. This distinction is foundational to CHAL: the system is designed for domains where conclusions are inherently defeasible, not for domains where proof is possible. More broadly, the ability to construct, evaluate, and respond to arguments is central to rational inquiry across disciplines, from philosophy and law to science and ethics, and it is this capacity that CHAL aims to instantiate in an automated multi-agent system. Toulmin's model provides the direct structural motivation for CHAL's belief representation, as formalized in Section~\ref{sec:CBS}.

\textbf{Further Evolution of Argumentation.} Caminada and Amgoud~\cite{MCaminada_LAmgoud_2007} provided a useful functional decomposition of argumentation as a reasoning process consisting of four canonical steps: (1)~constructing arguments for and against a given statement from available information; (2)~determining the conflicts among those arguments; (3)~evaluating the acceptability of arguments; and (4)~deriving justified conclusions. Further, Walton extended the field of argumentation theory in two important directions~\cite{DWalton_etal_2008}: the taxonomy of \emph{argumentation schemes} and the theory of \emph{dialogue types}. These argumentation themes consisted of roughly sixty stereotypical patterns of defeasible reasoning such as ``argument from expert opinion,'' ``argument from analogy,'' and ``argument from consequences'', among others. The theory of dialogue types distinguished between inquiry (seeking truth), deliberation (deciding on action), negotiation (distributing goods), persuasion (resolving a difference of opinion), and eristic (winning at all costs). For comprehensive surveys on the field of argumentation theory, see~\cite{IRahwan_GSimari_2009, FHvanEemeren_etal_2014}, and for a textbook treatment of argumentation among agents, see the chapter by Rahwan in~\cite{GWeiss_2013}.

\subsection{Computational Argumentation}

\textbf{The Intersection of Argumentation with Computation.} The computational treatment of argumentation began with Dung's well-known abstract argumentation framework~\cite{PDung_1995} which represents arguments as nodes and attacks as directed edges in a graph. The power of this formulation lies in its generality which allows one to prove formal properties of argumentation without specifying what the arguments actually are. However, the framework provides no account of the internal content, nor calibrated confidence of arguments, nor of how an agent's belief state should change in response to defeat. Aiming to address this was the development of several \emph{structured} argumentation frameworks that allow arguments to be instantiated with explicit inferential content. The ASPIC+~\cite{SModgil_HPrakken_2014} is perhaps the most comprehensive and theoretically well-developed, and several other approaches can be reconstructed as special cases of it. It represents arguments as inference trees built from strict rules (producing indefeasible conclusions) and defeasible rules (producing revisable conclusions), and distinguishing three attack types (two introduced in~\cite{JLPollock_1987}): \emph{undermining} (attacking a premise), \emph{rebutting} (attacking a conclusion with a counter-conclusion), and \emph{undercutting} (attacking an inference step without directly contradicting the conclusion). This classification determines what kind of response is dialectically appropriate to each attack and is adopted directly in CHAL's cross-examination methodology (see Section~\ref{sec:CHAL} and Appendix~\ref{app:attack_taxonomy}). The primary theoretical alternative to ASPIC+ is Assumption-Based Argumentation (ABA)~\cite{ABondarenko_etal_1997}, which attacks assumptions rather than conclusions or inference steps, and has its own rich solver ecosystem. Defeasible Logic Programming (DeLP)~\cite{AGarcia_GSimari_2004} comes from the logic programming tradition, using dialectical trees. The Argument Interchange Format (AIF)~\cite{CCChesnevar_etal_2006} provides a framework-agnostic ontology for exchanging argument structures between tools, with formal translations to and from ASPIC+ given in~\cite{FBex_etal_2013}. Finally, the Carneades framework~\cite{TFGordon_etal_2007} specializes in burden-of-proof reasoning and is widely used in legal AI. An earlier formulation of ASPIC+~\cite{HPrakken_2010} introduces a particularly relevant decomposition of the agent knowledge base into four disjoint sets: \emph{axioms} (necessary premises that cannot be attacked), \emph{ordinary premises} (premises that can be attacked, with defeat determined by preference comparison), \emph{assumptions} (premises whose attacks always succeed), and \emph{issues} (premises that must be backed by further argument to be acceptable). These four dimensions of knowledge representation are heavily incorporated in the design of the CBS (see Section~\ref{sec:CBS}).

\noindent\textbf{Probabilistic and Belief-Revision Approaches to Argumentation.} A complementary line of work connects argumentation theory to probabilistic epistemology by assigning degrees of belief to arguments within Dung-style frameworks. Specifically, the probabilistic argumentation theory developed in~\cite{AHunter_2017} treats the strength of an argument not as a binary property but as a function of the probability that its premises are true and its inference rules are applicable. At the level of agent belief states rather than individual arguments, the AGM belief revision theory~\cite{CAlchourron_etal_1985} (due to Alchourrón, Gärdenfors, and Makinson) provides an explanation of how a rational agent should revise their beliefs in response to new information. The AGM requires that belief revision be minimal (preserving as much of the original belief state as possible while accommodating the new information), consistent (the revised state should not contain contradictions), and faithful to the new information (the new belief is accepted). The connection between argumentation theory and belief revision has itself been studied formally~\cite{MFalappa_etal_2002}, which identifies that argumentation can be used to provide explanations for why a belief should be revised, grounding the AGM account in an explicit dialectical process. Argumentation and belief revision thus serve complementary purposes: the AGM theory governs how a revision is conducted, while the outcome of an argument determines whether revision is warranted at all. Dynamic epistemic logic~\cite{HvanDitmarsch_etal_2007} further extends this account to multi-agent settings, formalizing how the beliefs and knowledge of multiple agents change in response to events, communications, and the observation of each other's actions.

\subsection{Philosophy of Epistemology}

\textbf{History of Epistemology.} The philosophical study of knowledge, known as the field of epistemology~\cite{MSteup_RNeta_SEP_2025}, provides the foundations on which argumentation theory is built. The original treatment and exploration of knowledge can be traced back to Plato's \emph{Theaetetus}, where it is classically defined as \emph{justified true belief}: a belief constitutes knowledge if and only if it is true, the subject holds it, and the subject has adequate justification for it. This view on knowledge was broadly maintained until Gettier's consequential 1963 paper~\cite{EGettier_1963} which demonstrated that this tripartite definition is insufficient by providing counterexamples, referred to as Gettier cases, which occur when an agent has a justified true belief that intuitively does not constitute knowledge. Since the arrival of Gettier's work, the field of epistemology has seen a substantial surge of contributions with the aim to provide a new universally agreed-upon definition of knowledge. The foundationalism-coherentism debate~\cite{RFumerton_SEP_2023_old} concerns the structure of justification: foundationalists hold that justified beliefs are ultimately grounded in basic beliefs that require no further justification, while coherentists maintain that beliefs are justified by their coherence with the overall web of beliefs. Reliabilism~\cite{AGoldman_1979} posits that beliefs are justified if they are formed through reliable processes, emphasizing that knowledge can be obtained even if the individual is not aware of the evidence supporting their belief.

\textbf{Overview of Epistemic Positions.} We now provide a brief overview of the epistemic positions provided with CHAL, each of which roots a council agent in a distinct philosophical tradition (see Appendix~\ref{app:epistemic_positions}). Not all of the positions constitute explicit stances on the nature of knowledge in the strict epistemological sense; several are worldview frameworks whose commitments lead to knowledge claims that other positions are unlikely to reach, and it is this diversity of reachable conclusions that motivates their inclusion. The three most classical epistemic positions are the following: empiricism~\cite{DHume_1748} holds that knowledge derives from sensory experience and observation, rationalism~\cite{RDescartes_1641} holds that reason and a priori deduction are the primary sources of knowledge, and  Skepticism~\cite{SextusEmpiricus_1933} challenges the very possibility of attaining knowledge, arguing that for any claim, equally compelling counterarguments can always be constructed. Supernaturalism~\cite{CTaliaferro_SEP_2024} holds that reality extends beyond what is empirically accessible, and that faith, revelation, and spiritual experience constitute legitimate epistemic sources; in this sense it broadly encompasses the epistemological commitments of most religious worldviews. Pragmatism~\cite{WJames_1907} holds that beliefs are validated by their practical consequences and are always provisional and subject to revision. Bayesian epistemology~\cite{WTalbott_SEP_2022} treats beliefs as subjective probabilities governed by the laws of probability, with belief revision handled through conditionalization. Constructivism~\cite{TSKuhn_1962} argues that knowledge is constituted through social practices and paradigmatic frameworks rather than direct access to an objective world. Phenomenology~\cite{DDWoodruffSmith_SEP_2018} holds that first-person conscious experience is an irreducible source of knowledge prior to theoretical interpretation. Nihilism~\cite{FNietzsche_1887} denies the existence of objective truths, grounding meaning-making in psychological and cultural projection. Panpsychism~\cite{DJChalmers_1996} treats consciousness as a fundamental feature of reality with epistemic implications that extend beyond the physical. Simulationism~\cite{NBostrom_2003} and synthesism~\cite{KWilber_2001} are best understood as foundational worldview commitments: that the simulation hypothesis constrains what can be known about the nature of reality, and that partial truths across traditions should be integrated rather than treated as mutually exclusive, respectively.

\subsection{Philosophy of Logic and Formal Reasoning}

\textbf{Historical Roots of Logic and Mathematics.} The history of the formalization of logic is deep and has extensive ties to the field of mathematics. The systematic study of valid inference originates with Aristotle, whose \emph{Prior Analytics}~\cite{JBarnes_ed_1984} introduced the \emph{syllogism} as the first formal account of deductive reasoning: a type of argument in which a conclusion necessarily follows from two premises. This work set the agenda for logic for over two millennia and provides the direct ancestor of the inference rules that appear in modern formal systems. Similar to Aristotle's treatment of reasoning, the first serious attempt to ground an entire body of mathematical knowledge in an explicit axiomatic system is Euclid's \emph{Elements}~\cite{Euclid_300BCE}. Starting from five postulates and five common notions, Euclid derived many propositions in geometry, number theory, and solid geometry through chains of deductive proof. The \emph{Elements} established the axiomatic method as the standard for mathematical and logical rigor. This method consisted of organizing a body of knowledge by identifying a minimal set of assumptions from which all results follow by proof alone. This method dominated mathematics for more than two thousand years and directly informed the foundationalist program of the nineteenth century.

\textbf{The Modern Axiomatic Treatment of Mathematics.} The modern mathematical foundations program began with Cantor's development of set theory in the late nineteenth century~\cite{GCantor_1891}. Cantor's framework offered the prospect of a universal language in which all mathematical objects could be defined as sets and all mathematical theorems derived from set-theoretic principles. This prospect was severely damaged by the discovery of Russell's paradox in 1901~\cite{BRussell_1903}, which stated that the set of all sets that do not contain themselves neither contains nor fails to contain itself, producing a direct contradiction within Cantor's naive set theory. The response to this crisis was to develop a consistent set of axioms to define sets, referred to as axiomatic set theory. Zermelo proposed the first axiomatization in 1908, later extended by Fraenkel and Skolem into Zermelo-Fraenkel set theory with the axiom of choice (ZFC)~\cite{HBEnderton_1977}. ZFC has since become the standard foundational framework for mathematics; virtually all mathematical objects and theorems can be formulated and proved within it. Russell and Whitehead responded to this paradox in their own attempt with the \emph{Principia Mathematica}~\cite{BRussell_ANWhitehead_1910}, a three-volume work that aimed to derive all of mathematics from purely logical axioms using a theory of types to prevent the paradox. The \emph{Principia} represented the high point of \emph{logicism}, the thesis that mathematics is reducible to pure logic. However, this program was decisively terminated by Gödel's incompleteness theorems of 1931~\cite{KGodel_1931}. The first incompleteness theorem states that any consistent formal system capable of expressing elementary arithmetic contains true statements that are unprovable within the system. The second states that no such system can prove its own consistency. These results showed that the logicist goal of a complete and self-justifying foundation for mathematics is unattainable and that there is no formal system (capable of expressing numbers and their relations) that is simultaneously consistent, complete, and capable of proving its own soundness.

\textbf{Modern Formal Systems of Logic.} The logical systems underlying Gödel's foundational work were formalized in parallel. \emph{Propositional logic} concerns the logical relations between atomic propositions combined by connectives (negation, conjunction, disjunction, implication), treating propositions as having truth values without examining their internal structure. \emph{Predicate logic}, or first-order logic, is an extension of propositional logic by introducing quantifiers and predicates, allowing statements about objects and their properties. Together, propositional and predicate logic constitute \emph{classical logic}~\cite{HBEnderton_2001}, which is characterized by two fundamental meta-logical properties. \emph{Soundness} states that any formula derivable within the system is logically valid: no false conclusion can be reached from true premises. \emph{Completeness}, proved by G\"{o}del in 1930~\cite{KGodel_1930}, states the converse: any logically valid formula is derivable within the system. Classical logic is therefore adequate in the sense that its proof system captures exactly the set of logical truths. Gödel's incompleteness theorems, however, show that this adequacy does not extend to arithmetic: when a system is powerful enough to express the natural numbers, there are arithmetical truths that are valid but not derivable, and no extension of the axiom system can close this gap without introducing new unprovable truths.

\textbf{Overview of Logic Systems.} We now provide a brief overview of the systems of logic that are utilized by CHAL, each of which is meant to encourage further exploration in utilizing these different reasoning systems with LLMs (see Appendix~\ref{app:logic}). It bears mentions that there is a fair amount of overlap between some of the different systems, but the primary goal is to provide CHAL with different flavors of reasoning for further experimentation. The default system, Classical~+~Informal~+~Bayesian~\cite{HBEnderton_2001, DWalton_2006, WTalbott_SEP_2022}, combines deductive validity with Bayesian inductive and abductive support. Formal deductive logic~\cite{HBEnderton_2001} restricts valid inference to strict syllogistic derivation, rejecting inductive and abductive reasoning entirely. Dialectical logic~\cite{JEMaybee_SEP_2020} views contradiction as productive tension between a thesis and antithesis that drives reasoning toward synthesis. Informal and critical thinking logic~\cite{DWalton_2006} focuses on the identification of fallacies, the relevance of premises, and the sufficiency of evidence. Fuzzy logic~\cite{LAZadeh_1965} replaces the binary true/false distinction with truth degrees in $[0,1]$, enabling reasoning under vagueness; similar, yet distinct from Bayesian reasoning. Paraconsistent logic~\cite{GPriest_etal_SEP_2022} allows for reasoning to continue in the presence of local inconsistencies without global inferential collapse. The question of which logical system is correct, or whether the question is well-formed, is the subject of the philosophy of logic~\cite{SHaack_1978,JCBeall_GRestall_2006}. In CHAL, this question is not settled but made into a configuration parameter: the choice of logic system is user-specified, making the inferential standards of the system transparent and open to empirical study.

\subsection{Philosophy of Ethics}

\textbf{Overview of Ethical Systems.} Moral philosophy encompasses a broad set of traditions concerned with the question of what makes an action, belief, or outcome right or good. The history of morality, ethical, and meta-ethical systems has existed throughout the entire existence of mankind and we do not aim to provide a comprehensive coverage of all of these varied views or the development of their ideas, which would be past the scope of this work. For an extensive treatment of the history of moral thought, see Russell~\cite{BRussell_1945}, and for a systematic treatment of both the meta-ethical status and the normative content of ethics, see Mackie~\cite{JLMackie_1977}. CHAL supports a series of ethics systems, each rooted in a distinct normative tradition, and the following provides a brief description of each (see Appendix~\ref{app:ethics}). Utilitarianism~\cite{JBentham_1789, JSMill_1863} grounds the rightness of actions in their outcomes: an action is right if it produces the best achievable consequences, measured in terms of welfare, happiness, or preference satisfaction. In contrast, Deontological ethics, associated with Kant's \emph{Groundwork of the Metaphysics of Morals}~\cite{IKant_1785}, grounds rightness in conformity with universal moral duties: the central test here being the \emph{categorical imperative}, which holds that an action is permissible only if the maxim it expresses could be universalized without contradiction. Rule consequentialism~\cite{BHooker_SEP_2023}, judges actions by moral rules (such as restrictions on harming the innocent and lying) whose general acceptance would produce the greatest expected welfare, and is typically viewed as a combination of utilitarianism and deontological ethics. Virtue ethics~\cite{Aristotle_NE_1985} shifts the question from what one should do to what kind of individual one should strive to be, grounding morality in the character traits that enable human flourishing. Lastly, care ethics~\cite{CGilligan_1982} grounds moral evaluation in relationships and responsibilities of care rather than in universal rules or outcomes, emphasizing context over abstraction. CHAL makes no claim about the sufficiency or superiority of any one of these systems over another. Instead, by making the ethical framework of the adjudicator an explicit and configurable parameter, CHAL provides a platform for studying how different normative commitments contribute in shaping the conclusions that agents draw, and for developing methods by which agentic systems can be governed by explicitly defined ethical standards. The ability to specify, vary, and empirically study the ethical framework of a reasoning system has direct implications for the broader challenge of building AI systems whose normative behavior is transparent and aligned with human values.

\section{Meta-Cognitive Systems}\label{app:meta_sys}

This appendix provides summary tables and descriptions of the three meta-cognitive system configuration parameters introduced in Section~\ref{sec:meta_config}: the epistemic personas assigned to council members and the logic and ethics systems governing adjudicator inference and  normative evaluation.

\subsection{Definitions of Epistemic Positions}\label{app:epistemic_positions}

CHAL provides twelve epistemic positions, each grounded in a well-established philosophical tradition, summarized in Table~\ref{tab:personas}. Each position is designed to be philosophically specific, shaping not only what positions agents adopt but how they challenge and evaluate opposing ones. Agents are instructed to use their worldview as a lens rather than a set of conclusions to defend: the goal is to instantiate genuine epistemological diversity in order to generate broad insights and critiques.

\begin{table}[h]
\centering
\caption{CHAL epistemic personas and their philosophical commitments.}
\label{tab:personas}
\small
\begin{tabular}{ll}
\toprule
\textbf{Persona} & \textbf{Core Commitment} \\
\midrule
Empiricist~\cite{DHume_1748}                      & Knowledge requires sensory or experimental evidence. \\
Rationalist~\cite{RDescartes_1641}                & Reason and apriori deduction are primary knowledge sources. \\
Skeptic~\cite{SextusEmpiricus_1933}               & For any claim, equally compelling counter-arguments exist. \\
Pragmatist~\cite{WJames_1907}                     & Truth is validated by practical outcomes of inquiry. \\
Bayesian~\cite{WTalbott_SEP_2022}                 & Belief is graded probability, revised by conditionalization. \\
Constructivist~\cite{TSKuhn_1962}                 & Knowledge is constituted through social and linguistic practices. \\
Phenomenologist~\cite{DDWoodruffSmith_SEP_2018}   & First-person experience is an irreducible knowledge source. \\
Nihilist~\cite{FNietzsche_1887}                   & No objective truths; meaning-making is psychological. \\
Panpsychist~\cite{DJChalmers_1996}                & Consciousness is a fundamental feature of reality. \\
Simulationist~\cite{NBostrom_2003}                & Evaluates claims through the lens of simulated reality. \\
Supernaturalist~\cite{CTaliaferro_SEP_2024}       & Faith and reason are complementary epistemic sources. \\
Synthesist~\cite{KWilber_2001}                    & Integrates partial truths across traditions. \\
\bottomrule
\end{tabular}
\end{table}

\subsection{Definitions of Logical Systems}\label{app:logic}

The logic system configures the inferential standards the adjudicator applies when evaluating argument validity. CHAL provides eight logic systems, summarized in Table~\ref{tab:logic}. The default, Classical~+~Informal~+~Bayesian (CIB), combines deductive validity with inductive and abductive support weighted by prior probability, and is the most broadly applicable framework for the philosophical and empirical topics that CHAL is designed for. The remaining systems are distinct frameworks with different ontologies of valid inference. The choice of logic system materially affects adjudication outcomes and is therefore a primary variable of interest for empirical studies of debate dynamics.

\begin{table}[h]
\centering
\caption{CHAL systems of logic.}
\label{tab:logic}
\small
\begin{tabular}{ll}
\toprule
\textbf{Logic System} & \textbf{Description} \\
\midrule
None (Pure Ethics)                            & No logical consistency check; ethical evaluation only. \\ 
Classical + Informal + Bayesian~\cite{HBEnderton_2001, DWalton_2006, WTalbott_SEP_2022} & Default modern-day hybrid reasoning system. \\
Classical Formal Deductive~\cite{HBEnderton_2001}       & Strict deductive validity only; rejects informal inferences. \\
Bayesian~\cite{WTalbott_SEP_2022}       & Evaluate arguments by weighing available evidence and priors.\\
Dialectical~\cite{JEMaybee_SEP_2020}          & Hegelian thesis-antithesis-synthesis; contradiction is productive. \\
Informal / Critical~\cite{DWalton_2006}       & Fallacy identification, relevance, and sufficiency of evidence. \\
Fuzzy / Multi-valued~\cite{LAZadeh_1965}      & Truth admits degrees; avoids binary true/false judgments. \\
Paraconsistent~\cite{GPriest_etal_SEP_2022}   & Tolerates local contradictions without inferential explosion. \\
\bottomrule
\end{tabular}
\end{table}

\subsection{Definitions of Ethical Systems}\label{app:ethics}

The ethics system configures the normative framework the adjudicator applies alongside the logic system. CHAL provides six ethics systems, summarized in Table~\ref{tab:ethics}, spanning the major traditions in normative philosophy.

\begin{table}[h]
\centering
\caption{CHAL ethical systems.}
\label{tab:ethics}
\small
\begin{tabular}{ll}
\toprule
\textbf{Ethics System} & \textbf{Description} \\
\midrule
None (Pure Logic)~      & No ethical framework; logical evaluation only. \\
Utilitarian~\cite{JBentham_1789,JSMill_1863} & Evaluate by outcomes: maximize welfare, minimize suffering. \\
Deontological~\cite{IKant_1785} & Evaluate by universal duties and the categorical imperative. \\
Virtue Ethics~\cite{Aristotle_NE_1985} & Evaluate by promotion of human flourishing and practical wisdom. \\
Care Ethics~\cite{CGilligan_1982} & Evaluate through relationships, responsibility, and context. \\
Balanced (Rule-Utilitarian)~\cite{BHooker_SEP_2023} & Weigh duties and outcomes; rules are justified by consequences. \\
\bottomrule
\end{tabular}
\end{table}

\subsection{Scoring Scales}\label{app:scoring}

CHAL employs three calibrated scoring scales: one for agent-declared component strengths and two for adjudicator-assigned evaluation scores, all three of which map to $[0,1]$. The incorporation of these scales in the internal prompting structure of CHAL is necessary to calibrate the LLMs responses and ensure that all agents are operating with the same shared evaluation methodology.

\subsubsection{The Strength Scale}\label{app:strength_scale}

All strength-bearing components share a single calibrated scale on $[0,1]$, summarized in Table~\ref{tab:strength_scale}. The scale is referenced in every prompt that asks an agent to assign or revise a strength value, and the adjudicator uses it to evaluate whether agent-declared strengths are consistent with the supporting record. Strength values represent subjective probabilities: a claim strength of~0.9, for instance, corresponds to roughly a~10\% probability of being wrong given the current state of the belief, and most claims in philosophical and contested empirical domains are expected to fall in the~$0.2$ to~$0.8$ range.

\begin{table}[h]
\centering
\caption{The calibrated strength scale used by all strength-bearing CBS components.}
\label{tab:strength_scale}
\small
\begin{tabular}{lll}
\toprule
\textbf{Range} & \textbf{Label} & \textbf{Interpretation} \\
\midrule
$0.0$              & Vacuous     & No credible support; should be retracted. \\
$0.0$--$0.3$      & Weak        & Indefensible without substantial strengthening. \\
$0.3$--$0.5$      & Contested   & More reasons to doubt than to believe. \\
$0.5$              & Threshold   & Evenly balanced. \\
$0.5$--$0.7$      & Moderate    & More reasons to believe than to doubt; gaps remain. \\
$0.7$--$0.9$      & Strong      & Well supported; minor open questions. \\
$0.9$--$1.0$      & Robust      & Near certain given available evidence. \\
$1.0$              & Definitive  & Established beyond reasonable dispute. \\
\bottomrule
\end{tabular}
\end{table}

\subsubsection{Logic Scoring Scale}\label{app:logic_scale}

For each challenge-rebuttal pair, the adjudicator assigns a logic score on $[0,1]$ to both the challenger and the defender, producing the component scores $\sigma_L^c$ and $\sigma_L^d$, respectively, that enter the weighted combination in Definition~\ref{def:adjudicator}. Each score reflects the structural soundness of the party's argument under the configured logic system (see Appendix~\ref{app:logic}): whether premises support conclusions, whether inferences are valid, and whether the reasoning avoids fallacies. Table~\ref{tab:logic_scale} summarizes the calibration.

\begin{table}[h]
\centering
\caption{Logic scoring scale used by the adjudicator.}
\label{tab:logic_scale}
\small
\begin{tabular}{ll}
\toprule
\textbf{Score} & \textbf{Interpretation} \\
\midrule
$0.0$          & No coherent reasoning; argument fails on basic logical grounds. \\
$0.1$--$0.3$   & Severe logical flaws; reasoning is present but fundamentally unsound. \\
$0.4$--$0.5$   & Mixed; some valid reasoning but significant gaps or fallacies. \\
$0.6$--$0.7$   & Mostly sound reasoning with minor logical weaknesses. \\
$0.8$--$0.9$   & Strong logical reasoning with only minor issues. \\
$1.0$          & Rigorous; logically airtight with no identifiable flaws. \\
\bottomrule
\end{tabular}
\end{table}

\subsubsection{Ethics Scoring Scale}\label{app:ethics_scale}

For each challenge-rebuttal pair, the adjudicator assigns an ethics score on $[0,1]$ to both the challenger and the defender, producing the component scores $\sigma_E^c$ and $\sigma_E^d$, respectively, that enter the weighted combination in Definition~\ref{def:adjudicator}. Each score evaluates the normative quality of the agent's argument relative to the configured ethics system (see Appendix~\ref{app:ethics}): whether moral principles are applied consistently, whether implications are considered, and whether the reasoning respects the commitments of the assigned ethical tradition. Table~\ref{tab:ethics_scale} summarizes the calibration.

\begin{table}[h]
\centering
\caption{Ethics scoring scale used by the adjudicator.}
\label{tab:ethics_scale}
\small
\begin{tabular}{ll}
\toprule
\textbf{Score} & \textbf{Interpretation} \\
\midrule
$0.0$          & Ethically untenable; position violates basic moral principles. \\
$0.1$--$0.3$   & Significant ethical concerns; weak moral reasoning or harmful implications. \\
$0.4$--$0.5$   & Mixed ethical standing; some valid moral considerations but notable gaps. \\
$0.6$--$0.7$   & Generally ethically sound with minor concerns. \\
$0.8$--$0.9$   & Strong ethical reasoning; well-considered moral implications. \\
$1.0$          & Ethically exemplary; thorough moral reasoning with no concerns. \\
\bottomrule
\end{tabular}
\end{table}

\section{The CHAL Belief Schema}

This appendix provides the field-level specification of each CBS component introduced in Section~\ref{sec:CBS}, the constraints governing its construction, the role each component plays and its necessity in the dialectical pipeline, and the overall truth-oriented mechanisms built into CHAL. 

\subsection{Belief Component Details}\label{app:CBS_components}

The seven CBS components form an integrated system in which each component serves a distinct structural or epistemic function and are summarized in Table~\ref{tab:cbs_components}. Removing any one of them would break a specific mechanism that the system relies on to drive belief refinement toward accuracy rather than rhetorical victory. In each of the following subsections, we will specify what each component contributes to the system and what would degrade or fail in its absence.
In summary, definitions ground the semantics; assumptions and evidence provide the epistemic foundation; claims build the transparent inferential structure; counterpositions and uncertainties enforce honesty and track open questions; and the thesis closes the system with a deterministic, formula-driven summary statistic and provides the core strength-directed revision structure to beliefs.

\begin{table}[h]
\centering
\caption{CBS components and their roles in the belief representation.}
\label{tab:cbs_components}
\small
\begin{tabular}{lllp{7.5cm}}
\toprule
\textbf{Component} & \textbf{Notation} & \textbf{Prefix} & \textbf{Description} \\
\midrule
Thesis  & $\theta$           & ---  & The agent's core stance, summary bullets, and scalar strength $s_\theta \in [0,1]$ computed from the active claim set. \\
Definitions  & $\mathcal{D}$  & D\# & Semantic bedrock of the belief; fix the meaning of key terms used by assumptions and evidence. \\
Assumptions  & $\mathcal{A}$  & A\# & Foundational premises classified by type; grounded in one or more definitions. \\
Evidence   & $\mathcal{E}$    & E\# & Supporting data with type, source, and quality assessment; grounded in one or more definitions. \\
Claims   & $\mathcal{C}$      & C\# & Primary assertions with dependencies, backing assumptions and evidence, inference chains, and falsifiable predictions. \\
Counterpositions  & $\mathcal{X}$ & X\# & Strongest self-acknowledged arguments against the agent's own position, each with an attack type, self-assessed strength, current response, and response sufficiency rating. \\
Uncertainties  & $\mathcal{U}$ & U\# & Unresolved questions targeting specific nodes, with importance ratings used for adversarial prioritization. \\
\bottomrule
\end{tabular}
\end{table}

\subsubsection{Definitions (D\#)}\label{app:definitions}

Definitions establish the meaning of terms used throughout a CBS. Specifically, each D\# node fixes the meaning of a single term used by one or more assumptions or evidence items, ensuring that both the agent's internal reasoning and cross-agent comparisons rest on consistent terminology. The strength of a definition acts as a ceiling on every A\# and E\# node that references it via the constraint~\eqref{eq:d_ceiling}. 
Such a ceiling constraint ensures that no downstream node can be stronger than the terms it relies on, forcing agents to invest in defensible semantics before building arguments. Without D\#, two agents could use the same word with different meanings and reach apparent agreement or disagreement that is purely terminological. Definitions are the most consequential nodes in the BDG because a single strength reduction cascades through every dependent A\#, E\#, C\#, and ultimately the thesis. This cascade is what makes definitional attacks during cross-examination strategically valuable. Each definition $d \in \mathcal{D}$ carries the fields listed in Table~\ref{tab:def_fields}. 

\begin{table}[h]
\centering
\caption{Fields of a Definition (D\#) node.}
\label{tab:def_fields}
\small
\begin{tabular}{p{3.6cm} p{9.6cm}}
\toprule
\textbf{Field} & \textbf{Description} \\
\midrule
\texttt{id}
  & Unique identifier of the form D\#. Fixed at creation. \\
\addlinespace
\texttt{term}
  & The term being defined. \\
\addlinespace
\texttt{definition}
  & The definition text.\\
\addlinespace
\texttt{strength}
  & Calibrated strength $s_d \in [0,1]$ per Table~\ref{tab:strength_scale}.
    Acts as the ceiling for all dependent A\# and E\# nodes
    via~\eqref{eq:d_ceiling}. \\
\addlinespace
\texttt{strength\_justification}
  & Rationale for the assigned strength value. \\
\addlinespace
\texttt{status}
  & One of \texttt{active}, \texttt{revised}, or \texttt{retracted}.
    Retraction forces $s_d = 0$ and excludes the node from all ceiling
    calculations. \\
\addlinespace
\texttt{used\_by}
  & List of A\# and E\# identifiers that depend on this
    definition.\\
\addlinespace
\texttt{original\_strength}
  & System-managed defense boost baseline. \\
\addlinespace
\texttt{consecutive\_defenses}
  & System-managed defense streak counter. \\
\bottomrule
\end{tabular}
\end{table}

\subsubsection{Assumptions (A\#)}\label{app:assumptions}

Assumptions are the foundational premises on which claims are built. Each $a \in \mathcal{A}$
carries the fields listed in Table~\ref{tab:assump_fields}. One of these fields is the assumption type, which takes one of four values. These types encode distinct epistemic profiles summarized in Table~\ref{tab:assump_types}.
 
\begin{table}[h]
\centering
\caption{Fields of an Assumption (A\#) node.}
\label{tab:assump_fields}
\small
\begin{tabular}{p{3.6cm} p{9.6cm}}
\toprule
\textbf{Field} & \textbf{Description} \\
\midrule
\texttt{id}
  & Unique identifier of the form A\#. \\
\addlinespace
\texttt{type}
  & One of \texttt{foundational}, \texttt{empirical}, \texttt{methodological},
    or \texttt{scoping}; see Table~\ref{tab:assump_types}. \\
\addlinespace
\texttt{statement}
  & The assumption text. \\
\addlinespace
\texttt{supports\_claims}
  & List of C\# identifiers that this assumption supports. \\
\addlinespace
\texttt{supported\_by\_definitions}
  & List of D\# identifiers that ground this assumption. At least one
    is required.\\
\addlinespace
\texttt{strength}
  & Calibrated strength $s_a \in [0,1]$, capped by the minimum strength of its
    supporting definitions via~\eqref{eq:d_ceiling}. \\
\addlinespace
\texttt{strength\_justification}
  & Rationale for the assigned strength value. \\
\addlinespace
\texttt{status}
  & One of \texttt{active}, \texttt{revised}, or \texttt{retracted}. \\
\addlinespace
\texttt{original\_strength}
  & System-managed defense boost baseline. \\
\addlinespace
\texttt{consecutive\_defenses}
  & System-managed defense streak counter. \\
\bottomrule
\end{tabular}
\end{table}
 
\begin{table}[h]
\centering
\caption{Assumption types and their epistemic profiles.}
\label{tab:assump_types}
\small
\begin{tabular}{p{2.8cm} p{9.0cm}}
\toprule
\textbf{Type} & \textbf{Epistemic Profile} \\
\midrule
\texttt{foundational}
  & Definitional or logical axioms. \\
\addlinespace
\texttt{empirical}
  & Accepted on the basis of evidence. \\
\addlinespace
\texttt{methodological}
  & Adopted for analytical purposes. \\
\addlinespace
\texttt{scoping}
  & Delimits the scope of inquiry. \\
\bottomrule
\end{tabular}
\end{table}

\subsubsection{Evidence (E\#)}\label{app:evidence}

Evidence items provide empirical and conceptual support for claims. Each $e \in \mathcal{E}$ carries the fields listed in Table~\ref{tab:evidence_fields}. 
One of these fields is the evidence type, which takes one of three values. These types encode distinct epistemic profiles, described in Table~\ref{tab:evidence_types}.
Evidence items occupy the same structural position as assumptions in the~BDG: both are grounded in definitions via the D\# ceiling constraint and both support claims. The distinction is epistemic rather than structural. Assumptions are commitments an agent adopts without external verification and can be challenged based on their classification type (foundational, empirical, methodological, or scoping); evidence items are assertions of external support that can themselves be challenged on the basis of source reliability (i.e., whether the cited source can withstand adversarial scrutiny from other agents), sample quality (i.e., whether the data are sufficient to support the strength assigned to the evidence node), or scope of applicability (i.e., whether the evidence actually applies to the claim it is cited to support).

Without typed assumptions, an agent could mislabel a premise to deflect the appropriate class of attack, and the system would have no way to detect this. Without a dedicated evidence component, there would be no way to distinguish claims backed by external sources from those resting purely on the agent's own unverified commitments. Merging the two into a single component would sacrifice both of these properties and narrow the range of attacks available during cross-examination.

\begin{table}[h]
\centering
\caption{Fields of an Evidence (E\#) node.}
\label{tab:evidence_fields}
\small
\begin{tabular}{p{3.6cm} p{9.6cm}}
\toprule
\textbf{Field} & \textbf{Description} \\
\midrule
\texttt{id}
  & Unique identifier of the form E\#. \\
\addlinespace
\texttt{type}
  & One of \texttt{empirical}, \texttt{conceptual}, or \texttt{expert\_consensus};
    see Table~\ref{tab:evidence_types}. \\
\addlinespace
\texttt{summary}
  & The substantive content of the evidence. \\
\addlinespace
\texttt{source}
  & The origin or citation of the evidence. \\
\addlinespace
\texttt{supports\_claims}
  & List of C\# identifiers that this evidence supports. \\
\addlinespace
\texttt{supported\_by\_definitions}
  & List of D\# identifiers grounding the terms used in the evidence.
    At least one is required.\\
\addlinespace
\texttt{strength}
  & Calibrated strength $s_e \in [0,1]$, capped by the minimum strength of its
    supporting definitions via~\eqref{eq:d_ceiling}. \\
\addlinespace
\texttt{strength\_justification}
  & Rationale for the assigned strength value, including stated limitations. \\
\addlinespace
\texttt{status}
  & One of \texttt{active}, \texttt{revised}, or \texttt{retracted}. \\
\addlinespace
\texttt{original\_strength}
  & System-managed defense boost baseline. \\
\addlinespace
\texttt{consecutive\_defenses}
  & System-managed defense streak counter. \\
\bottomrule
\end{tabular}
\end{table}

\begin{table}[h]
\centering
\caption{Evidence types and their epistemic profiles.}
\label{tab:evidence_types}
\small
\begin{tabular}{p{2.8cm} p{10.4cm}}
\toprule
\textbf{Type} & \textbf{Epistemic Profile} \\
\midrule
\texttt{empirical}
  & Observational or experimental data drawn from direct measurement, studies,
    or documented observations. \\
\addlinespace
\texttt{conceptual}
  & Philosophical or theoretical arguments that support a claim through reasoning
    rather than observation. \\
\addlinespace
\texttt{expert\_consensus}
  & Authoritative agreement among recognized experts or established bodies of
    knowledge in a relevant domain. \\
\bottomrule
\end{tabular}
\end{table}

\subsubsection{Claims (C\#)}\label{app:claims}

Claims are the primary substantive assertions of a CBS belief and form the inference tier of the
BDG, connecting the foundation tier to the thesis. Each $c \in \mathcal{C}$ carries the fields
listed in Table~\ref{tab:claim_fields}. Each claim must include an \texttt{inference\_chain}: an ordered list of at least three structured
reasoning steps that explicitly connect the claim's dependencies to its conclusion. This ordering
is structurally validated, ensuring that the agent has articulated the full reasoning path from its
cited dependencies to its asserted conclusion and exposing inferential gaps that would otherwise
remain hidden behind a single declared strength value. Each step carries a \texttt{role} and a
\texttt{text} field; Table~\ref{tab:inference_chain_steps} specifies the three permitted roles and
their additional required fields. Each claim must also include at least one falsifiable prediction implied by it, which is stored in the \texttt{predictions} field.. A prediction is falsifiable if it specifies in advance what outcome would confirm or refute it; for example, a claim that a treatment reduces symptoms must predict a measurable threshold of improvement that would count as confirmation, and a level of failure that would count as refutation. This is captured by the \texttt{decision\_criterion} field. This requirement, following Popper~\cite{KPopper_1959}, prevents agents from populating the field with vague or untestable assertions. Both inference chains and predictions are primary targets during cross-examination, where weak
inference steps and vague decision criteria are common attack vectors. Table~\ref{tab:prediction_fields} specifies the full set of prediction fields.

\begin{table}[h]
\centering
\caption{Fields of a Claim (C\#) node.}
\label{tab:claim_fields}
\small
\begin{tabular}{p{3.6cm} p{9.6cm}}
\toprule
\textbf{Field} & \textbf{Description} \\
\midrule
\texttt{id}
  & Unique identifier of the form C\#. \\
\addlinespace
\texttt{type}
  & Free-form label categorizing the claim
    (e.g., \texttt{descriptive}, \texttt{causal}, \texttt{deductive}). \\
\addlinespace
\texttt{statement}
  & The substantive assertion. \\
\addlinespace
\texttt{depends\_on}
  & List of A\#, E\#, and C\# identifiers that this
    claim builds on. At least one dependency is required.\\
\addlinespace
\texttt{strength}
  & Calibrated strength $s_c \in [0,1]$, capped by the minimum strength of its dependencies via~\eqref{eq:d_ceiling}. \\
\addlinespace
\texttt{strength\_justification}
  & Rationale for the assigned strength. \\
\addlinespace
\texttt{status}
  & One of \texttt{active}, \texttt{revised}, or \texttt{retracted}. Only
    active claims contribute to the thesis strength formula~\eqref{eq:thesis_str}.
    Retraction forces $s_c = 0$ and triggers propagation through all
    dependent claims. \\
\addlinespace
\texttt{inference\_chain}
  & Ordered list of structured reasoning steps connecting the dependencies
    to the claim statement; see Table~\ref{tab:inference_chain_steps}. \\
\addlinespace
\texttt{predictions}
  & List of falsifiable predictions implied by the claim;
    see Table~\ref{tab:prediction_fields}. \\
\addlinespace
\texttt{original\_strength}
  & System-managed defense boost baseline. \\
\addlinespace
\texttt{consecutive\_defenses}
  & System-managed defense streak counter. \\
\bottomrule
\end{tabular}
\end{table}

\begin{table}[h]
\centering
\caption{The steps of an \texttt{inference\_chain} entry and their required fields.}
\label{tab:inference_chain_steps}
\small
\begin{tabular}{p{2.2cm} p{4.8cm} p{5.0cm}}
\toprule
\textbf{Role} & \textbf{Description} & \textbf{Additional Required Fields} \\
\midrule
\texttt{premise}
  & States a supporting fact or commitment. All premise steps must appear
    first in the chain.
  & \texttt{reference}: exactly one A\#, E\#, or
    C\# identifier. \\
\addlinespace
\texttt{inference}
  & States the inferential step connecting the premises to the conclusion.
    Exactly one inference step is required and must follow all premises.
  & \texttt{inference\_type}: one of \texttt{deductive}, \texttt{inductive},
    or \texttt{abductive}. \\
\addlinespace
\texttt{conclusion}
  & States the final assertion. Exactly one conclusion step is required and
    must appear last. Its text must match the claim's \texttt{statement}.
  & None. \\
\bottomrule
\end{tabular}
\end{table}

\begin{table}[h]
\centering
\caption{Fields of a \texttt{predictions} entry within a C\# node.}
\label{tab:prediction_fields}
\small
\begin{tabular}{p{3.6cm} p{9.6cm}}
\toprule
\textbf{Field} & \textbf{Description} \\
\midrule
\texttt{statement}
  & The falsifiable prediction itself. \\
\addlinespace
\texttt{test}
  & How the prediction can be tested. \\
\addlinespace
\texttt{decision\_criterion}
  & The outcome that would confirm or disconfirm the prediction. \\
\addlinespace
\texttt{potential\_falsifiers}
  & Evidence that would disprove the prediction (optional). \\
\bottomrule
\end{tabular}
\end{table}

\subsubsection{Counterpositions (X\#)}\label{app:counterpositions}

Counterpositions encode the strongest known arguments against the agent's own position. Each~$x \in \mathcal{X}$ carries the fields listed in Table~\ref{tab:counterposition_fields}.
Counterpositions rated \texttt{partial} or \texttt{unaddressed} become prioritized targets for
opponent cross-examination and high-priority resolution targets during belief refinement. An agent
that disguises weaknesses by rating an inadequate response as \texttt{sufficient} invites
adjudicated penalties when those responses fail under scrutiny; an agent that honestly flags
partial responses receives explicit prompting to resolve them in the next round.
Table~\ref{tab:response_sufficiency} specifies the four \texttt{response\_sufficiency} values
and their interpretations.
Counterpositions are non-strength-bearing nodes in the BDG and do not participate in the
dependency constraint equations. Their effect on belief refinement operates through the
adversarial pipeline rather than through strength propagation.

\begin{table}[h]
\centering
\caption{Fields of a Counterposition (X\#) node.}
\label{tab:counterposition_fields}
\small
\begin{tabular}{p{3.6cm} p{9.6cm}}
\toprule
\textbf{Field} & \textbf{Description} \\
\midrule
\texttt{id}
  & Unique identifier of the form X\#. \\
\addlinespace
\texttt{targets}
  & List of D\#, A\#, E\#, or C\#
    identifiers that this counterposition attacks. \\
\addlinespace
\texttt{attack\_type}
  & ASPIC$+$ classification from $\{\texttt{undermining},\ \texttt{rebutting},\
    \texttt{undercutting}\}$ (see Appendix~\ref{app:attack_taxonomy}). \\
\addlinespace
\texttt{attack\_strategy}
  & Specific sub-strategy from the taxonomy defined in
    Appendix~\ref{app:attack_taxonomy}. \\
\addlinespace
\texttt{statement}
  & The counter-argument as an opponent would present it. \\
\addlinespace
\texttt{my\_response}
  & The agent's current response to the counterposition. \\
\addlinespace
\texttt{response\_sufficiency}
  & One of \texttt{sufficient}, \texttt{partial}, \texttt{unaddressed}, or
    \texttt{moot}; see Table~\ref{tab:response_sufficiency}. \\
\bottomrule
\end{tabular}
\end{table}

\begin{table}[h]
\centering
\caption{\texttt{response\_sufficiency} values and their interpretations.}
\label{tab:response_sufficiency}
\small
\begin{tabular}{p{2.8cm} p{10.0cm}}
\toprule
\textbf{Value} & \textbf{Interpretation} \\
\midrule
\texttt{sufficient}
  & The response fully addresses the objection. \\
\addlinespace
\texttt{partial}
  & Some aspects of the objection are addressed but gaps remain. \\
\addlinespace
\texttt{unaddressed}
  & No adequate response exists yet; represents a known vulnerability
    in the agent's position. \\
\addlinespace
\texttt{moot}
  & Terminal state: the counterposition's target node has been retracted,
    rendering the objection no longer applicable. \\
\bottomrule
\end{tabular}
\end{table}

Counterpositions are the primary mechanism for enforcing epistemic honesty while simultaneously allowing an agent to catalog and prepare for an onslaught of different attacks that it has encountered. Concretely, a new X\# is created whenever the adjudicator returns a \texttt{CRITIQUE\_VALID} verdict: the losing agent must add a counterposition acknowledging the vulnerability that was exposed, with \texttt{response\_sufficiency} capped at \texttt{partial} or \texttt{unaddressed}. This prevents the agent from claiming to have fully addressed an argument it just lost. Only in subsequent Stage~5 belief revision phases or later debate rounds, after the agent has developed a substantive defense, may it upgrade the counterposition to \texttt{sufficient}. In this way, if an agent loses a rebuttal, it is able to reflect upon and revise its response on the same critique for future defenses. As such, counterpositions are a way of giving the agent a sense of history and experience that less challenged agents will lack. Without X\#, agents would have no structured way to surface self-acknowledged weaknesses and opponents would lose a source of potential vulnerabilities that guide adversarial targeting. One of the primary benefits of CHAL is in producing refined beliefs that have exhaustive collections of counterpositions; each X\# represents a position that the agent has encountered before, reasoned out a response to, and in turn carved out a more robust belief on the topic.

\subsubsection{Uncertainties (U\#)}\label{app:uncertainties}

Uncertainties record open questions that the agent recognizes but cannot currently answer. Each~$u \in \mathcal{U}$ carries the fields listed in Table~\ref{tab:uncertainty_fields}. Uncertainties serve two roles in the pipeline. First, they are prioritized targets for opponent
cross-examination via the \texttt{press\_uncertainty} strategy (see Appendix~\ref{app:attack_taxonomy}), with \texttt{high} and
\texttt{medium} importance items receiving particular attention; this ensures that agents cannot simply stay silent about what they do not know — any open question must be explicitly declared, making it a visible target for opponent cross-examination. Second, they are mandatory review
targets during the belief refinement stage, where the agent is prompted to either resolve them
by introducing new D\#, A\#, E\#, or C\# nodes, or to
leave them open and accept the corresponding limitation on its position. Adjudication outcomes
classified as \texttt{UNRESOLVED} must also be recorded as new uncertainties, ensuring that the
belief honestly reflects points at which the debate failed to reach a verdict rather than
absorbing them silently. Like counterpositions, uncertainties are non-strength-bearing nodes and do not participate in the dependency constraints of~\eqref{eq:d_ceiling}.

\begin{table}[h]
\centering
\caption{Fields of an Uncertainty (U\#) node.}
\label{tab:uncertainty_fields}
\small
\begin{tabular}{p{3.6cm} p{9.6cm}}
\toprule
\textbf{Field} & \textbf{Description} \\
\midrule
\texttt{id}
  & Unique identifier of the form U\#. \\
\addlinespace
\texttt{targets}
  & List of D\#, A\#, E\#, or C\#
    identifiers to which the uncertainty pertains. \\
\addlinespace
\texttt{question}
  & The unresolved question. \\
\addlinespace
\texttt{importance}
  & One of \texttt{high}, \texttt{medium}, or \texttt{low}, indicating how
    consequential resolution would be for the agent's position. \\
\addlinespace
\texttt{status}
  & One of \texttt{active} or \texttt{resolved}. \\
\addlinespace
\texttt{resolution\_note}
  & Explanation of how the uncertainty was resolved. Required when
    \texttt{status} changes to \texttt{resolved}. \\
\bottomrule
\end{tabular}
\end{table}

Uncertainties provide the explicit record of what the agent does not know. Without them, \texttt{UNRESOLVED} adjudication outcomes would have nowhere to be recorded, open questions would be silently absorbed rather than tracked across rounds, and the council's intellectual honesty would be shrouded. The system's ability to distinguish between a position that has been tested and found robust and one that has simply avoided scrutiny of its weakest points would be compromised.

\subsubsection{The Thesis Object}\label{app:thesis}

The thesis $\theta$ summarizes the agent's overall position and is the only component whose
strength is computed by formula rather than declared by the agent. It contains the fields listed
in Table~\ref{tab:thesis_fields}.

\begin{table}[h]
\centering
\caption{Fields of the Thesis ($\theta$) object.}
\label{tab:thesis_fields}
\small
\begin{tabular}{p{3.6cm} p{9.6cm}}
\toprule
\textbf{Field} & \textbf{Description} \\
\midrule
\texttt{stance}
  & The agent's core position statement, with references to
    the supporting claim identifiers. \\
\addlinespace
\texttt{summary\_bullets}
  & A list of key themes of the position (at least one required, typically
    three to ten), used for rendering and for cross-agent semantic
    comparisons. \\
\addlinespace
\texttt{strength}
  & The thesis strength $s_\theta \in [0,1]$, always equal to the result
    of applying~\eqref{eq:thesis_str} to the active claim set. \\
\addlinespace
\texttt{strength\_reasoning}
  & Auto-generated trace showing the values of $\bar{s}$, $n$, and $B(n)$
    used in~\eqref{eq:thesis_str}.
    Provided for human auditability but not consumed programmatically. \\
\bottomrule
\end{tabular}
\end{table}

The thesis is generated last during Stage~1 and Stage~5 Phase~2, after every supporting component has been
constructed, so that the stance and summary bullets are grounded in the claims the agent has
actually committed to rather than written first and reverse-engineered into a supporting
structure. After the thesis has been generated, its strength is deterministically computed via~\eqref{eq:thesis_str}. This is the central guarantee of the strength architecture
described in Section~\ref{sec:strength_arch}: the thesis strength is a function of the
underlying belief structure, and agents cannot inflate it independently of the quality of their
supporting arguments. The strength of the thesis is the quantitative objective that each agent works to maximize. Without it, there would be no single scalar measure of overall belief quality, and agents would have no measurable objective to drive progress or incentive to learn and improve their positions.

\subsubsection{The Truth-Oriented Mechanisms}\label{app:truth_mechanisms}

\textbf{Truth-Directed Mechanisms.} Beyond the necessity of the individual components, the CBS incorporates a set of architectural mechanisms that collectively push the system toward truth rather than rhetorical victory. In this context, and throughout the rest of the paper, we are utilizing the term ``truth'' to refer to the most defensible and strongest position that one can attain. This stands in opposition to the idea of some objectively correct position, which is an inherently dubious concept in the domain of defeasible argumentation. The thesis strength is computed deterministically via~\eqref{eq:thesis_str} from the active claim set, so agents cannot inflate it without improving the quality of their underlying arguments. The D\# ceiling constraint in~\eqref{eq:d_ceiling} ensures that the strength of every assumption and evidence item is capped by the strength of the definitions it relies on, so a weak or poorly constructed definition automatically propagates its weakness upward through the entire argument. Claims cannot be stronger than their weakest dependency, and the \texttt{strength\_justification} field forces agents to name the limiting node, creating transparency about exactly where the argument is weakest, also aiding the agent in revising or retracting its weakest nodes. The defense boost mechanism of Appendix~\ref{app:def_boost} ensures that a node's strength can only increase as a direct result of surviving an adjudicated challenge, so higher strengths must be earned through adversarial pressure rather than declared unilaterally. Counterposition honesty is enforced by the fact that a weak response rated \texttt{sufficient} gets exposed during cross-examination and penalized during adjudication, while honest self-assessment triggers structured resolution prompts in the next round. Uncertainties operate similarly: high-importance U\# nodes are prioritized for opponent attack, so agents cannot conceal them, but resolving an uncertainty through new supporting material genuinely strengthens the position. At the prompt level, agents are instructed that positions are a lens for analysis rather than a set of conclusions to defend, and to update when the worldview conflicts with strong evidence. The universal debate prompt explicitly directs intellectual honesty over rhetorical victory, and that genuine concession when a critique lands is encouraged.

\textbf{The Belief Refinement Cycle.} Across multiple rounds, these mechanisms produce an adversarial-cooperative cycle that converges toward more robust and accurately calibrated beliefs. In each round, cross-examination targets weak D\#, A\#, E\#, and C\# nodes and exposes vulnerabilities, which the rebuttal stage forces the defender to either \texttt{refute}, \texttt{concede}, or \texttt{defer}. Enforcement applies mandatory weakening patches and new X\# entried for \texttt{critique\_valid} outcomes, defense boosts for \texttt{rebuttal\_valid} outcomes, and new U\# entries for \texttt{unresolved} outcomes. The introspective refinement phase then adds new supporting material, resolves outstanding counterpositions and uncertainties, and retires weak claims, guided by the gradient analysis of Section~\ref{sec:grad_prompting}. The net effect of each round is to cull weak arguments, mechanically strengthen defended ones, and add new material along the most productive local direction. Over the full sequence of rounds, this cycle drives each belief toward a configuration in which every surviving node carries a strength that has been tested against the strongest available counterargument, and in which the thesis strength reflects the structural quality of the argument rather than the agent's willingness to assert confidence.

\subsection{The Defense Boost Mechanism}\label{app:def_boost}

The \emph{defense boost} is layered on top of the patch system to keep agent-declared strengths aligned with the adversarial record of the debate. To that end, when a strength-bearing node survives an adjudicated challenge, the system mechanically applies a small, fixed increase to its strength, capped in cumulative magnitude relative to the node's original strength.

\begin{definition}[Defense Boost]\label{def:def_boost}
Let $v$ be a strength-bearing node with original strength $s_v^\mathrm{orig}$ and current strength $s_v$. Given a fixed boost increment $b \in (0,1]$ and a maximum cumulative boost $c_{\max} \in (0,1]$, the updated strength of $v$ after a successful defense is given by
\begin{equation}\label{eq:boost_update}
    s_v \;\longleftarrow\; \min\!\left(\, s_v + b,\;\; s_v^\mathrm{orig} + c_{\max},\;\; 1 \,\right).
\end{equation}
\end{definition}

The three-way minimum in~\eqref{eq:boost_update} enforces three invariants simultaneously: the per-defense boost is exactly $b$, the cumulative boost above $s_v^\mathrm{orig}$ never exceeds $c_{\max}$, and the strength never exceeds $1$. We utilize default values of $b = 0.02$ and $c_{\max} = 0.15$.

\textbf{Cascading Boosts.} After direct boosts are applied to all nodes targeted by \texttt{rebuttal\_valid} verdicts, the system performs a breadth-first traversal of the BDG from each boosted node toward the foundation tier, incrementing every parent node's strength by the same amount, subject to the ceiling in~\eqref{eq:boost_update}, and re-capping claim strengths via the dependency constraints~\eqref{eq:d_ceiling} after all boosts are applied. The intuition is structural. Foundation-tier nodes (D\#, A\#, E\#) are the premises on which the arguments rest; they carry few dependencies of their own, expose a short inferential surface, and are more straightforward to defend. A successfully defended assumption or evidence node therefore cascades only to the definitions that ground it, or, in the case of a definition, boosts only itself. Inference-tier nodes (C\#) do the argumentative heavy lifting: each claim aggregates support from multiple foundation-tier dependencies, and the longer inferential chain creates more points upon a larger surface of vulnerability. A successful defense of a claim is therefore a stronger signal of robustness, and the cascade propagates through a larger portion of the graph, reinforcing the full inferential foundation that sustained the defense.

\section{The CHAL Dialectical Pipeline Algorithm}\label{app:CHAL_algorithm}

In this appendix, we provide the full algorithmic specification of the dialectical pipeline which was described in Section~\ref{sec:pipeline}. Algorithm~\ref{alg:CHAL} serves as a self-contained reference for the interaction between the council (Section~\ref{sec:council_form}), the adjudicator (Section~\ref{sec:adjudicator}), and the belief revision machinery (Sections~\ref{sec:strength_arch}--\ref{sec:grad_prompting}).

\begin{algorithm}[H]
\caption{The CHAL Dialectical Pipeline}\label{alg:CHAL}
\begin{algorithmic}[1]
    \item[] \parbox[t]{\linewidth}{\textbf{Input:} Topic query $q \in \mathbb{T}$; council $\bigodot$ of $N$ agents with assigned personas $\{\beta^i\}_{i=1}^N$; number of rounds $R$; challenges per pair per round $S$; adjudicator $\blacklozenge_\vartheta$ with parameters $\vartheta = (L, E, w_L, w_E, \tau)$.}
    \item[] \parbox[t]{\linewidth}{\textbf{Stage 0 (Briefing).} Apply universal debate prompt and persona role card $\beta^i$ to each $\odot^i \in \bigodot$.}
    \item[] \parbox[t]{\linewidth}{\textbf{Stage 1 (Opening Positions).} Each $\odot^i$ generates an initial CBS belief $\mathscr{B}_i^{(0)}$ from $(q, \beta^i)$.}
    \item[] \textbf{For} $r = 1, 2, \dots, R$ \textbf{do}
    \item[] \hspace*{1em}\parbox[t]{\dimexpr\linewidth-1em}{\textbf{Stage 2 (Cross-Examination).} For each ordered pair $(\odot^i, \odot^j)$ with $i \neq j$, agent $\odot^i$ generates up to $S$ targeted challenges against $\mathscr{B}_j^{(r-1)}$.}
    \item[] \hspace*{1em}\parbox[t]{\dimexpr\linewidth-1em}{\textbf{Stage 3 (Rebuttal).} Each $\odot^j$ produces a structured rebuttal to every challenge received (from all agents), committing to \texttt{refute}, \texttt{concede}, or \texttt{defer} per challenge.}
    \item[] \hspace*{1em}\parbox[t]{\dimexpr\linewidth-1em}{\textbf{Stage 4 (Adjudication).} For each challenge-rebuttal pair $(\chi, \rho)$, the adjudicator renders a verdict $\blacklozenge_\vartheta(\chi, \rho) \in \mathcal{V}$.}
    \item[] \hspace*{1em}\textbf{Stage 5 (Belief Revision).}
    \item[] \hspace*{2em}\parbox[t]{\dimexpr\linewidth-2em}{\textbf{Phase 1:} Each agent $\odot^i$ emits a mandatory patch sequence $\Pi_i^{\mathrm{enf}}$ in response to verdicts and applies them to $\mathscr{B}_i^{(r-1)}$ with constraint enforcement via~\eqref{eq:d_ceiling}.}
    \item[] \hspace*{2em}\parbox[t]{\dimexpr\linewidth-2em}{\textbf{Defense boosts:} Apply boosts per Appendix~\ref{app:def_boost} for \texttt{rebuttal\_valid} outcomes.}
    \item[] \hspace*{2em}\parbox[t]{\dimexpr\linewidth-2em}{\textbf{Phase 2:} Each agent $\odot^i$ receives a dynamic position analysis from $\nabla s_\theta$, emits an introspective patch sequence $\Pi_i^{\mathrm{int}}$, and applies them with constraint enforcement via~\eqref{eq:d_ceiling}, producing $\mathscr{B}_i^{(r)}$.}
    \item[] \textbf{End for}
    \item[] \parbox[t]{\linewidth}{\textbf{Output:} Final beliefs $\{\mathscr{B}_i^{(R)}\}_{i=1}^N$; debate transcript; embedding trajectories; performance statistics.}
\end{algorithmic}
\end{algorithm}

\section{The Attack Taxonomy}\label{app:attack_taxonomy}

The cross-examination prompt at Stage~2 provides each challenger with a taxonomy of attack strategies organized under the three ASPIC+ attack types. Each challenge must declare an \texttt{attack\_type} and an \texttt{attack\_strategy} from this taxonomy. The three attack types of \texttt{undermining}, \texttt{rebutting}, and \texttt{undercutting} and their associated strategies are specified in Tables~\ref{tab:undermining_strategies},~\ref{tab:rebutting_strategies}, and~\ref{tab:undercutting_strategies}, respectively.

\begin{table}[h]
\centering
\caption{\textbf{Undermining} strategies.}
\label{tab:undermining_strategies}
\small
\begin{tabular}{@{}l p{0.6\textwidth}@{}}
\toprule
\textbf{Strategy} & \textbf{Description} \\
\midrule
\texttt{challenge\_evidence} & Dispute the reliability, relevance, or sufficiency of an E\# node. \\
\texttt{challenge\_assumption} & Question whether an A\# is correctly typed or well-founded. \\
\texttt{expose\_weak\_foundation} & Show a high-strength claim depends on low-strength nodes. \\
\texttt{demand\_falsifiability} & The targeted claim or assumption makes no testable predictions. \\
\texttt{challenge\_strength\_calibration} & The assigned strength exceeds what the cited evidence warrants. \\
\texttt{press\_uncertainty} & Force the opponent to address their own U\# admitted unknowns. \\
\texttt{over\_extension} & A D\# definition is too broad, capturing cases it should not. \\
\texttt{under\_extension} & A D\# definition is too narrow, excluding cases it should capture. \\
\texttt{challenge\_moral\_implications} & The position leads to ethically problematic consequences. \\
\texttt{expose\_stakeholder\_harm} & Specific parties are disproportionately harmed by the position. \\
\bottomrule
\end{tabular}
\end{table}

\begin{table}[h]
\centering
\caption{\textbf{Rebutting} strategies.}
\label{tab:rebutting_strategies}
\small
\begin{tabular}{@{}l p{0.6\textwidth}@{}}
\toprule
\textbf{Strategy} & \textbf{Description} \\
\midrule
\texttt{present\_counter\_evidence} & Offer specific evidence that directly contradicts a claim. \\
\texttt{present\_counter\_example} & Provide a concrete case that falsifies a general claim. \\
\texttt{exploit\_counterposition} & Press on X\# nodes rated \texttt{partial} or \texttt{unaddressed}. \\
\texttt{offer\_alternative\_explanation} & The opponent's own evidence better supports a different conclusion. \\
\texttt{present\_ethical\_counter} & Present an ethical argument opposing the position's moral standing. \\
\texttt{invoke\_competing\_obligation} & A stronger moral obligation overrides the one the position appeals to. \\
\bottomrule
\end{tabular}
\end{table}

\begin{table}[h]
\centering
\caption{\textbf{Undercutting} strategies.}
\label{tab:undercutting_strategies}
\small
\begin{tabular}{@{}l p{0.6\textwidth}@{}}
\toprule
\textbf{Strategy} & \textbf{Description} \\
\midrule
\texttt{challenge\_inference\_step} & A step in a C\#'s inference chain does not follow from its predecessors. \\
\texttt{identify\_circularity} & The reasoning chain assumes its own conclusion. \\
\texttt{expose\_inconsistency} & Claims contradict each other or strength assignments violate constraints. \\
\texttt{identify\_equivocation} & Key terms shift meaning between premises and conclusion. \\
\texttt{challenge\_scope} & The conclusion overgeneralizes from the evidence. \\
\texttt{circularity} & A D\# definition uses the defined term within itself. \\
\texttt{stipulative\_bias} & A D\# definition is framed to presuppose the conclusion. \\
\texttt{conceptual\_conflation} & A D\# definition conflates two distinct concepts. \\
\texttt{challenge\_normative\_inference} & Factual premises do not justify the normative conclusions (is-ought gap). \\
\texttt{expose\_value\_conflict} & The position harbors an unacknowledged contradiction between its values. \\
\texttt{challenge\_moral\_relevance} & The invoked ethical principle does not apply to the specific case. \\
\bottomrule
\end{tabular}
\end{table}

\section{Numerical Experiments}\label{app:experiments}
 
This appendix provides supplementary details for the numerical experiments presented in Section~\ref{sec:experiments}. It begins with descriptions of the belief embedding methodology (see Appendix~\ref{app:embeddings}) and the Agent Performance Score (APS) (see Appendix~\ref{app:APS}), followed by detailed analysis and excerpts for all ablation studies, each of which utilize the embedding methodology as well as the APS. All experiments were collected on a laptop with an Intel Core i7-14650HX processor and 16\,GB of RAM. No local GPU was required, as all LLM inference is performed via cloud API calls. All experiments used OpenAI's o4-mini model and cost approximately \$125 in total API usage. Exact input and output token counts as well as durations for each experiment are reported in the relevant subsections.


\subsection{Belief Embedding Methodology}\label{app:embeddings}

We now define the embedding used to map each CBS belief object into a fixed-dimensional vector that can be consistently used for belief trajectory visualization.

\begin{definition}[Belief Embedding]\label{def:belief_embedding}
Let $d_\varphi \in \mathbb{N}$ denote the output dimension of some chosen sentence
transformer, and let $\varphi : \mathbb{T} \to \mathbb{R}^{d_\varphi}$ denote the
corresponding encoding function. For each strength-bearing component set
$T \in \{\mathcal{D}, \mathcal{A}, \mathcal{E}, \mathcal{C}\}$ of a belief object
$\mathscr{B}_i^{(r)}$ corresponding to the belief of agent $i\in\{1,2,...,N\}$ on round $r\in\{0,1,...,R\}$, with $m$ active (non-retracted) elements (nodes) having text-strength
pairs $\{(t_j, s_j)\}_{j=1}^{m}$, the \emph{component embedding} is the
strength-weighted mean
\begin{equation}\label{eq:comp_embed}
    \mathbf{h}_T \;=\; \frac{\sum_{j=1}^{m} s_j\,\varphi(t_j)}
                            {\sum_{j=1}^{m} s_j}
    \;\in\; \mathbb{R}^{d_\varphi},
\end{equation}
with fallback to the unweighted mean when $\sum_{j=1}^m s_j = 0$. The
\emph{belief embedding} $\mathbf{z}_i^{(r)}$ is the concatenation of ten semantic
vectors: four component embeddings $\mathbf{h}_{\mathcal{D}},\,\mathbf{h}_{\mathcal{A}},\,
\mathbf{h}_{\mathcal{E}},\,\mathbf{h}_{\mathcal{C}}$; one thesis embedding
$\mathbf{h}_{\theta}$; one uncertainty embedding $\mathbf{h}_{\mathcal{U}}$; and four
counterposition embeddings $\mathbf{h}_{\mathcal{X}}^{(1)}$, $\mathbf{h}_{\mathcal{X}}^{(2)}$, $\mathbf{h}_{\mathcal{X}}^{(3)}$, and
$\mathbf{h}_{\mathcal{X}}^{(4)}$, together with an 11-dimensional scalar feature
vector $\mathbf{f}$ of normalized component counts and per-component average strengths,
giving
\[
  \mathbf{z}_i^{(r)} \;\in\; \mathbb{R}^{10\,d_\varphi + 11}.
\]
\end{definition}

The text $t_j$ extracted from each node depends on the component type: for
definitions, $t_j$ is the concatenation of the \texttt{term} and
\texttt{definition} fields; for assumptions and claims, the \texttt{statement}
field; and for evidence, the \texttt{summary} field. The thesis embedding
$\mathbf{h}_{\theta}$ encodes the concatenation of the \texttt{stance} and
\texttt{summary\_bullets} fields as a single vector. The uncertainty embedding
$\mathbf{h}_{\mathcal{U}}$ is the unweighted mean of $\varphi$ applied to each
unresolved uncertainty's \texttt{question} text, and counterpositions are
partitioned by \texttt{response\_sufficiency} into the four following buckets in
concatenation order: \texttt{partial}, \texttt{sufficient}, \texttt{unaddressed},
\texttt{moot}. Each of the four buckets are encoded as an unweighted-mean vector, giving
$\mathbf{h}_{\mathcal{X}}^{(1)}$, $\mathbf{h}_{\mathcal{X}}^{(2)}$, $\mathbf{h}_{\mathcal{X}}^{(3)}$, and
$\mathbf{h}_{\mathcal{X}}^{(4)}$,
respectively. In all experiments reported in this paper, $\varphi$ is
instantiated as the \texttt{all-mpnet-base-v2} textual encoder model, for which $d_\varphi = 768$ and
$\mathbf{z}_i^{(r)} \in \mathbb{R}^{7691}$. Finally, for visualization (see Appendix~\ref{app:experiments}), the belief embeddings are projected to two dimensions via
UMAP.

 
\subsection{Agent Performance Score}\label{app:APS}
 
We now define the metric used to quantify each council member's dialectical
performance across a debate which will be used to empirically measure and identify the performance of each agent relevant to the other agents in the council (see Appendix~\ref{app:experiments}).
 
\begin{definition}[Agent Performance Score]\label{def:APS}
For a council member $\odot^i\in\bigodot$, let $K_i \in\mathbb{N}$ denote the total number of adjudicated
exchanges involving agent $i$ across all rounds, in both roles (as challenger and defender). Each exchange $k\in\{0,1,...,K_i\}$ yields a score
$\gamma_k \in \{-1.0,\,-0.5,\,0.0,\,+0.25,\,+1.0\}$ determined by the agent's
role in that exchange and the adjudicator's verdict, as specified in
Table~\ref{tab:aps_weights}. The \emph{Agent Performance Score (APS)} is defined as
\begin{equation}\label{eq:APS}
    \mathrm{APS}_i \;=\; \frac{1}{K_i}\sum_{k=1}^{K_i} \gamma_k \;\in\; [-1,\, 1].
\end{equation}
The per-round $\mathrm{APS}_i$ at round $r$ is the cumulative mean of all exchange
scores through round $r$.
\end{definition}
 
\begin{table}[h]
\centering
\caption{Per-exchange score contributions by role and verdict.}
\label{tab:aps_weights}
\small
\begin{tabular}{llr}
\toprule
\textbf{Role} & \textbf{Verdict} & \textbf{Score ($\gamma_k$)} \\
\midrule
Challenger & \texttt{critique\_valid}  & $+1.0$ \\
Challenger & \texttt{rebuttal\_valid}  & $-0.5$ \\
Challenger & \texttt{unresolved}       & $\phantom{+}0.0$ \\
Target     & \texttt{critique\_valid}  & $-1.0$ \\
Target     & \texttt{rebuttal\_valid}  & $+1.0$ \\
Target     & \texttt{unresolved}       & $+0.25$ \\
\bottomrule
\end{tabular}
\end{table}
 
The asymmetric scoring in Table~\ref{tab:aps_weights} is meant to reflect the structural
asymmetry of the dialectical exchange. Challengers take initiative while targets
are under obligation to respond, intuitively leading to different weighting between different roles. A failed challenge incurs a milder penalty
($-0.5$) than a failed defense ($-1.0$), since unsuccessful offense is less
costly than unsuccessful defense. An \texttt{unresolved} verdict awards $+0.25$
to the target, reflecting that surviving without concession under genuine
ambiguity is a mildly favorable outcome.

\subsection{Initial Epistemic Topology}\label{app:topology}

A central design intuition of CHAL is that populating the council with agents that reason from distinct epistemic traditions produces a broader exploration of the latent belief space than any single agent could achieve alone (see Figures~\ref{fig:single-agent} and~\ref{fig:chal-council} for the original intuition). To demonstrate this empirically, we generated 10 initial CBS beliefs for each of the 12 epistemic personas on the topic ``Does free will exist?'' and embedded the resulting 120 beliefs into a shared semantic space via UMAP projections (see Appendix~\ref{app:embeddings}), displayed in Figure~\ref{fig:initial_dist_umap}.

\begin{figure}[htbp]
    \centering
    \includegraphics[width=0.8\textwidth]{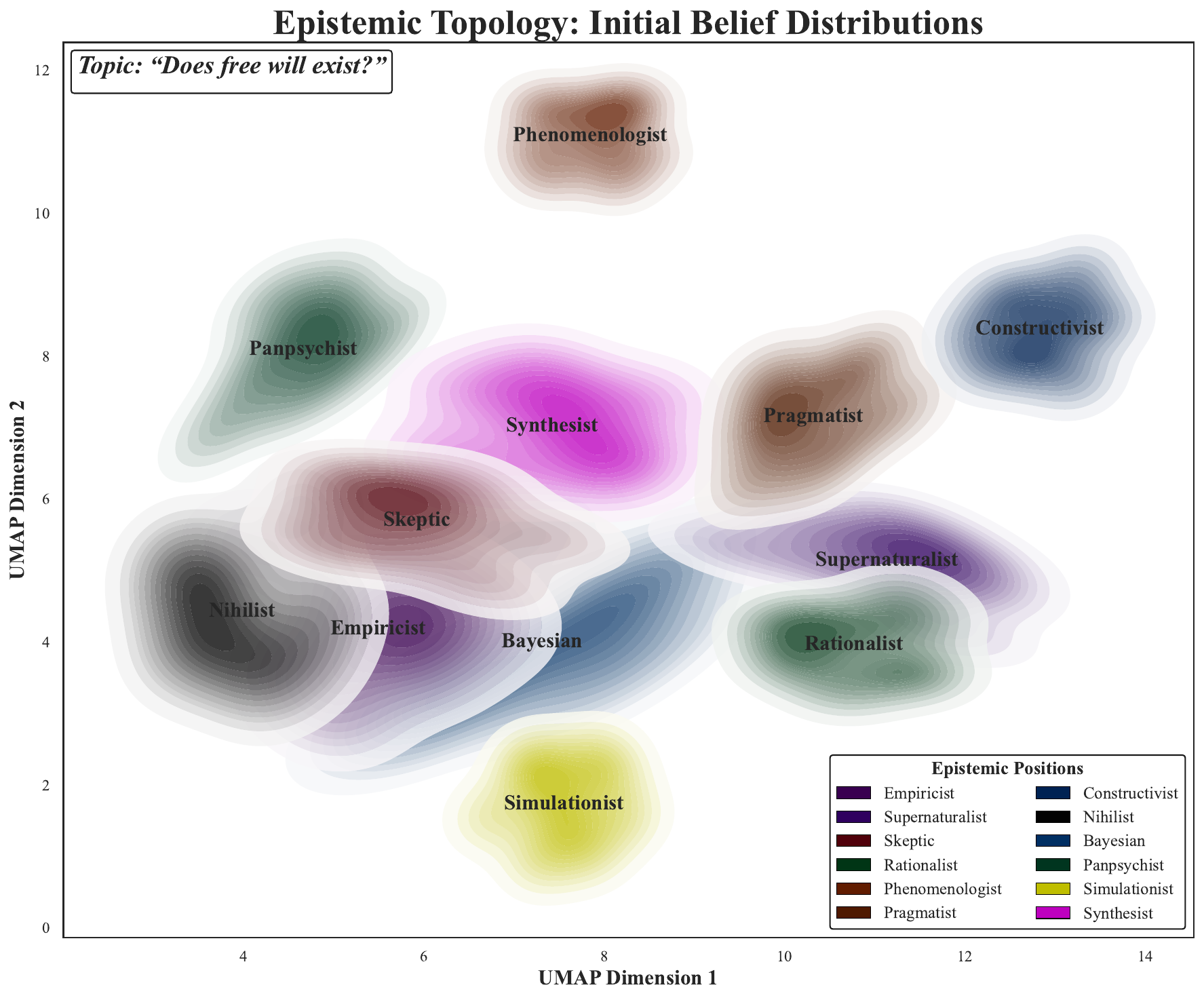}
    \caption{Semantic locations of the twelve epistemic personas in latent belief space on the topic ``Does free will exist?''.}
    \label{fig:initial_dist_umap}
\end{figure}

The twelve epistemic positions occupy distinct, largely separable regions of the embedding space, confirming that the epistemic prompts succeed in steering the underlying LLM model into semantically differentiated starting positions, from which several distinguishable patterns are observable. As one may expect, the Empiricist, Bayesian, Skeptic, and Nihilist distribution clusters share a fair amount of overlap, reflecting part of their shared commitment to evidential standards and epistemic caution. The Supernaturalist and Rationalist occupy neighboring but distinct regions, consistent with their overlapping interest in non-empirical justification but divergent ontological commitments. Interestingly, the Phenomenologist, Panpsychist, Simulationist, and Constructivist distribution clusters are the most isolated, reflecting their methodological distinctiveness and diversity on this topic. These distributions aid in supporting the premise motivating CHAL's council design: agents initialized with different epistemic personas do not merely rephrase the same position but occupy genuinely different regions of belief space, providing the diversity necessary for productive, novel, and varied adversarial exchange, a point that will be demonstrated further in the following experiments.

\subsection{Adjudicator Ablation: Detailed Analysis}\label{app:adj_ablation_details}

This appendix section provides detailed analysis of the three adjudicator ablation conditions described in Section~\ref{sec:ablation}. All three conditions use the same two council members (Empiricist and Supernaturalist) with shared initial beliefs, the same topic (``Does free will exist?''), and $n = 10$ sampled trajectories per configuration. The only variable that changes across conditions is the adjudicator configuration: (1)~pure logic ($w_L = 1, w_E = 0$, CIB), (2)~balanced ($w_L = 0.5, w_E = 0.5$, CIB and Deontological ethics), and (3)~pure ethics ($w_L = 0, w_E = 1$, Deontological ethics).

\begin{figure}[htbp]
    \centering
    \vspace{-0.5\baselineskip}
    \includegraphics[width=0.95\textwidth]{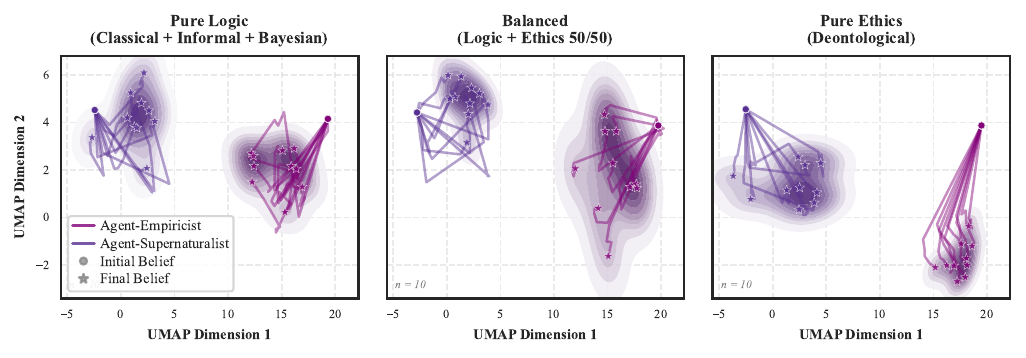}
    \vspace{-0.8\baselineskip}
    \caption{Belief trajectories across adjudicator ablation with density contours highlighting the final beliefs.}
    \label{fig:adj_trajectories}
    \vspace{-0.5\baselineskip}
\end{figure}
\begin{figure}[htbp]
    \centering
    \vspace{-0.5\baselineskip}
    \includegraphics[width=0.99\textwidth]{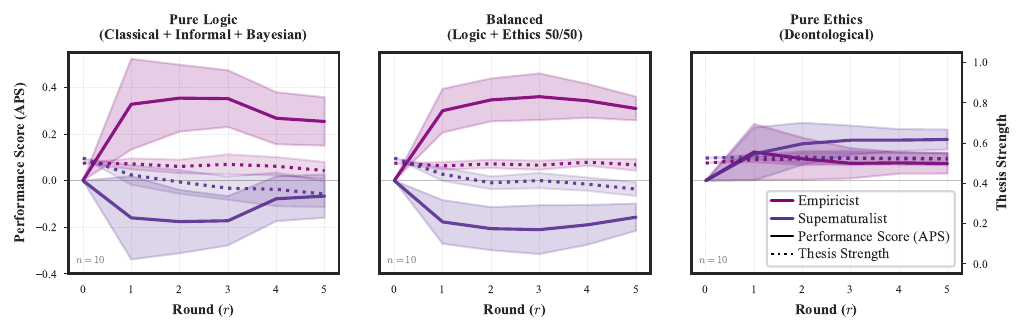}
    \vspace{-0.8\baselineskip}
    \caption{Per-round APS and thesis strength across the three adjudicator configurations with shaded regions indicating 95\% confidence intervals.}
    \label{fig:adj_performance}
    \vspace{-0.5\baselineskip}
\end{figure}

Figures~\ref{fig:adj_trajectories} and~\ref{fig:adj_performance} display the UMAP-projected belief trajectories and per-round performance metrics for all three configurations. The results demonstrate that the adjudicator's value system configuration materially shapes both the semantic trajectory and the performance of the council members, confirming its role as the primary driver of belief evolution in CHAL. Under the pure logic and balanced adjudicators, the Empiricist dominates consistently, with the Supernaturalist's thesis collapsing as its metaphysical claims fail falsifiability and Bayesian prior challenges. Under the pure ethics adjudicator, the outcome reverses: the Supernaturalist's Deontological framing of free will as a precondition for moral responsibility earns consistently high ethics scores, and it is favored in the majority of debates despite presenting the same general arguments that were systematically dismantled under pure logic. The trajectory plots highlight the corresponding shift in final belief distributions that the different adjudicator configurations induce. This performance reversal is a demonstration of the system's intended behavior: different meta-cognitive value systems produce different dialectical outcomes on the same topic with the same general arguments, and CHAL makes this dependence explicit and measurable. The adjudicator does not universally determine the truth of any matter; it determines which position is most defensible under a specified evaluative standard, serving as a proxy for truth.

Tables~\ref{tab:adj_performance}--\ref{tab:adj_components} report the general aggregate statistics across all three adjudicator configurations. Several general patterns are worth highlighting before examining individual excerpts. Echoing the performances displayed in Figure~\ref{fig:adj_performance}, Table~\ref{tab:adj_performance} shows that the Empiricist won (in terms of APS) 8/10 and 10/10 debates under the pure logic and balanced configurations, respectively, whereas the Supernaturalist won 9/10 debates under the pure ethics configuration. Investigating the adjudicator's verdict distributions in Table~\ref{tab:adj_verdicts}, we can also identify another pattern: the \texttt{critique\_valid} rate drops drastically from 58.6\% under pure logic to 13.5\% under pure ethics, while the \texttt{unresolved} rate rises from 7.8\% to 73.2\%. This discrepancy is due to the fact that both agents tend to favor arguing purely logical points over purely ethical ones. Hence, when utilizing a pure logic configuration, many of the disagreements are able to be consistently adjudicated and a valid or invalid verdict rendered. However, when utilizing a pure ethics configuration, the majority of disagreements have no ethical dispute, leading to the raised \texttt{unresolved} verdicts. 
This is also reflected in the belief component counts in Table~\ref{tab:adj_components}, which reveals that the pure ethics condition produces the largest growth in both counterpositions and uncertainties (X\# and U\# counts roughly doubling), while maintaining higher final claim counts than the other two conditions. The high \texttt{unresolved} rate under pure ethics translates directly into U\# accumulation, as each \texttt{unresolved} verdict generates a mandatory new uncertainty node.

We now proceed to discussing the results of each configuration. For each, we first summarize the aggregate performance dynamics and then present selected adjudicator reasoning excerpts. Each excerpt is preceded by a context box identifying the CBS nodes under dispute, the actual content of the attack and defense, followed by the adjudicator's verdict and reasoning (with node identifiers colored by BDG tier: {\color{teal!70!black}foundation}, {\color{orange!70!black}inference}, {\color{red!60!black}challenge}).

\begin{table}[t]
\centering
\caption{Agent performance and thesis-strength evolution across adjudicator configurations. Values are mean $\pm$ 95\,\% CI.}
\label{tab:adj_performance}
\footnotesize
\setlength{\tabcolsep}{4pt}
\begin{tabular}{|l|l|c|c|c|c|c|c|}
\hline
\textbf{Config} & \textbf{Agent} & \textbf{APS} & \textbf{Crit.\,\%} & \textbf{Reb.\,\%} & $\boldsymbol{s_{\theta,0}}$ & $\boldsymbol{s_{\theta,f}}$ & \textbf{Win Rate} \\
\hline
  \multirow{2}{*}{Pure Logic} & Empiricist & $\bm{0.25 \pm 0.10}$ & $\bm{68.3 \pm 6.2}$ & $\bm{41.8 \pm 8.6}$ & $0.50$ & $\bm{0.46 \pm 0.04}$ & $\bm{8/10}$ \\
   & Supernaturalist & $-0.07 \pm 0.09$ & $49.0 \pm 9.2$ & $25.3 \pm 5.7$ & $\bm{0.53}$ & $0.35 \pm 0.06$ & $2/10$ \\
\hline
  \multirow{2}{*}{Balanced} & Empiricist & $\bm{0.31 \pm 0.05}$ & $\bm{78.8 \pm 6.7}$ & $\bm{35.4 \pm 3.0}$ & $0.50$ & $\bm{0.49 \pm 0.03}$ & $\bm{10/10}$ \\
   & Supernaturalist & $-0.16 \pm 0.06$ & $48.9 \pm 5.1$ & $13.5 \pm 6.0$ & $\bm{0.53}$ & $0.37 \pm 0.03$ & $0/10$ \\
\hline
  \multirow{2}{*}{Pure Ethics} & Empiricist & $0.07 \pm 0.04$ & $12.5 \pm 2.8$ & $7.0 \pm 2.9$ & $0.50$ & $0.52 \pm 0.03$ & $2/10$ \\
   & Supernaturalist & $\bm{0.18 \pm 0.04}$ & $\bm{14.6 \pm 5.6}$ & $\bm{19.4 \pm 7.1}$ & $\bm{0.53}$ & $\bm{0.52 \pm 0.03}$ & $\bm{9/10}$ \\
\hline
\end{tabular}
\vspace{2pt}
\begin{minipage}{\columnwidth}
\scriptsize
\textbf{APS}: Agent Performance Score $\in [-1, 1]$, mean of per-exchange scores weighted by role and outcome. \textbf{Crit.\,\%}: percentage of exchanges won as challenger. \textbf{Reb.\,\%}: percentage of exchanges won as defender. $\boldsymbol{s_{\theta,0}}$/$\boldsymbol{s_{\theta,f}}$: initial/final thesis strength $\in [0, 1]$. \textbf{Win Rate}: fraction of debates where the agent achieved the highest APS.
\end{minipage}
\end{table}

\begin{table}[t]
\centering
\caption{Debate-level verdict distribution and operational cost across adjudicator configurations. Values are mean $\pm$ 95\,\% CI.}
\label{tab:adj_verdicts}
\footnotesize
\setlength{\tabcolsep}{4pt}
\begin{tabular}{|l|c|c|c|c|c|c|}
\hline
 & \multicolumn{3}{c|}{\textbf{Verdict Distribution (\%)}} & \multicolumn{3}{c|}{\textbf{Operational Cost}} \\
\cline{2-4} \cline{5-7}
\textbf{Config} & \textbf{Crit.\,V.} & \textbf{Reb.\,V.} & \textbf{Unres.} & \textbf{In\,Tok.} & \textbf{Out\,Tok.} & \textbf{Dur.\,(s)} \\
\hline
  Pure Logic & $58.6 \pm 5.1$ & $33.5 \pm 5.2$ & $7.8 \pm 2.9$ & $848.7\text{k} \pm 14.3\text{k}$ & $258.8\text{k} \pm 7.4\text{k}$ & $1145 \pm 112$ \\
  Balanced & $63.7 \pm 4.7$ & $24.5 \pm 3.0$ & $11.8 \pm 3.5$ & $920.1\text{k} \pm 21.1\text{k}$ & $267.2\text{k} \pm 6.4\text{k}$ & $1006 \pm 51$ \\
  Pure Ethics & $13.5 \pm 3.9$ & $13.3 \pm 3.3$ & $73.2 \pm 4.9$ & $801.7\text{k} \pm 19.4\text{k}$ & $229.6\text{k} \pm 8.3\text{k}$ & $1159 \pm 92$ \\
\hline
\end{tabular}
\vspace{2pt}
\begin{minipage}{\columnwidth}
\scriptsize
\textbf{Crit.\,V.}: critique valid; challenger's argument upheld. \textbf{Reb.\,V.}: rebuttal valid; defender's rebuttal upheld. \textbf{Unres.}: unresolved; neither side decisively prevailed. \textbf{In/Out\,Tok.}: total LLM input/output tokens. \textbf{Dur.}: wall-clock debate duration.
\end{minipage}
\end{table}

\begin{table}[t]
\centering
\caption{Cross-examination attack profile per agent across adjudicator configurations. Values are mean attack count $\pm$ 95\,\% CI.}
\label{tab:adj_attacks}
\footnotesize
\setlength{\tabcolsep}{5pt}
\begin{tabular}{|l|l|c|c|c|}
\hline
\textbf{Config} & \textbf{Agent} & \textbf{Undermining} & \textbf{Rebutting} & \textbf{Undercutting} \\
\hline
  \multirow{2}{*}{Pure Logic} & Empiricist & $16.3 \pm 1.3$ & $0.8 \pm 0.8$ & $7.8 \pm 1.1$ \\
   & Supernaturalist & $15.8 \pm 1.1$ & $0.7 \pm 0.5$ & $8.4 \pm 0.9$ \\
\hline
  \multirow{2}{*}{Balanced} & Empiricist & $14.2 \pm 1.1$ & $1.1 \pm 0.7$ & $9.0 \pm 1.0$ \\
   & Supernaturalist & $14.5 \pm 1.2$ & $0.6 \pm 0.6$ & $9.5 \pm 1.2$ \\
\hline
  \multirow{2}{*}{Pure Ethics} & Empiricist & $15.1 \pm 1.4$ & $1.4 \pm 0.8$ & $8.3 \pm 1.6$ \\
   & Supernaturalist & $15.5 \pm 2.0$ & $0.7 \pm 0.7$ & $7.9 \pm 1.4$ \\
\hline
\end{tabular}
\vspace{2pt}
\begin{minipage}{\columnwidth}
\scriptsize
Attack types from the argumentation framework (see Appendix~\ref{app:attack_taxonomy}).
\end{minipage}
\end{table}

\begin{table*}[t]
\centering
\caption{Belief component counts (initial and final) across adjudicator configurations. Values are mean $\pm$ 95\,\% CI.}
\label{tab:adj_components}
\scriptsize
\setlength{\tabcolsep}{3pt}
\begin{tabular}{|l|l|c|c|c|c|c|c|c|c|c|c|c|c|}
\hline
 &  & \multicolumn{2}{c|}{\textbf{D}} & \multicolumn{2}{c|}{\textbf{A}} & \multicolumn{2}{c|}{\textbf{C}} & \multicolumn{2}{c|}{\textbf{E}} & \multicolumn{2}{c|}{\textbf{X}} & \multicolumn{2}{c|}{\textbf{U}} \\
\cline{3-4} \cline{5-6} \cline{7-8} \cline{9-10} \cline{11-12} \cline{13-14}
\textbf{Config} & \textbf{Agent} & $i$ & $f$ & $i$ & $f$ & $i$ & $f$ & $i$ & $f$ & $i$ & $f$ & $i$ & $f$ \\
\hline
  \multirow{2}{*}{Pure Logic} & Empiricist & $5$ & $6.9 \pm 1.4$ & $4$ & $7.9 \pm 1.3$ & $3$ & $4.3 \pm 0.8$ & $4$ & $13.3 \pm 0.9$ & $4$ & $15.4 \pm 2.3$ & $3$ & $7.4 \pm 1.1$ \\
   & Supernaturalist & $5$ & $7.6 \pm 1.2$ & $5$ & $8.5 \pm 1.1$ & $3$ & $3.5 \pm 0.6$ & $2$ & $8.7 \pm 1.4$ & $3$ & $19.0 \pm 2.2$ & $3$ & $8.8 \pm 1.5$ \\
\hline
  \multirow{2}{*}{Balanced} & Empiricist & $5$ & $7.5 \pm 1.0$ & $4$ & $8.0 \pm 1.1$ & $3$ & $4.8 \pm 0.9$ & $4$ & $13.2 \pm 1.3$ & $4$ & $14.8 \pm 1.7$ & $3$ & $8.6 \pm 1.0$ \\
   & Supernaturalist & $5$ & $8.3 \pm 1.0$ & $5$ & $10.2 \pm 1.7$ & $3$ & $4.4 \pm 0.6$ & $2$ & $11.2 \pm 2.3$ & $3$ & $18.6 \pm 2.1$ & $3$ & $8.4 \pm 2.0$ \\
\hline
  \multirow{2}{*}{Pure Ethics} & Empiricist & $5$ & $8.3 \pm 1.0$ & $4$ & $8.7 \pm 0.8$ & $3$ & $5.5 \pm 0.8$ & $4$ & $13.0 \pm 1.2$ & $4$ & $8.3 \pm 1.5$ & $3$ & $18.0 \pm 2.7$ \\
   & Supernaturalist & $5$ & $8.7 \pm 0.9$ & $5$ & $9.5 \pm 1.0$ & $3$ & $5.1 \pm 0.7$ & $2$ & $10.0 \pm 1.7$ & $3$ & $7.5 \pm 1.2$ & $3$ & $16.0 \pm 1.6$ \\
\hline
\end{tabular}
\vspace{2pt}
\begin{minipage}{\textwidth}
\scriptsize
CBS belief components: \textbf{D}~=~definitions, \textbf{A}~=~assumptions, \textbf{C}~=~claims, \textbf{E}~=~evidence, \textbf{X}~=~counterpositions, \textbf{U}~=~uncertainties. $i$/$f$: initial (pre-debate) / final (post-debate) counts. Only non-retracted nodes are counted for D, A, C, E.
\end{minipage}
\end{table*}

\subsubsection{Pure Logic Adjudicator}\label{app:adj_pure_logic}

Under pure logic adjudication, the Empiricist won 8 of 10 debates by APS (see Table~\ref{tab:adj_performance}), driven by a substantial advantage in both challenger effectiveness (68.3\% vs.\ 49.0\% critique rate) and defensive capability (41.8\% vs.\ 25.3\% rebuttal rate). The Supernaturalist's thesis declined from $s_{\theta,0} = 0.53$ to a mean final strength of $0.35 \pm 0.06$ while the Empiricist's thesis remained comparatively stable ($0.50 \to 0.46 \pm 0.04$). The \texttt{unresolved} rate of 7.8\% (see Table~\ref{tab:adj_verdicts}) was the lowest of any configuration, indicating that the pure logic adjudicator produced the most decisive verdicts.

Although some of the debates had substantial variations, in general, the Supernaturalist's thesis would collapse following a characteristic sequence: its metaphysical claims (non-physical soul, the dualist ontological separation of mind and matter, etc.) would fall first to falsifiability and Bayesian prior challenges in early rounds; its moral responsibility claims would typically be defeated by counterexamples; and its phenomenological claims would erode in later rounds. After losing its core supernatural infrastructure, the Supernaturalist consistently reinvented its thesis by introducing novel (typically more empirical in nature) arguments not present in the opening position, some of which we highlight in the following excerpts. This adaptation is partially visible in Table~\ref{tab:adj_components}, where the Supernaturalist's evidence count grew from 2 to $8.7 \pm 1.4$, the largest relative evidence expansion in any configuration, reflecting the pivot from metaphysical assertion to empirical argumentation. By the final round, a ``supernaturalist'' council member arguing from neuroscience data was philosophically unrecognizable from its opening position.

\paragraph{Excerpt 1: Bayesian prior contention on non-physical claims.}\mbox{}\\
\begin{nodecontext}[title={\small\textbf{Nodes under dispute}}]
\Aref{A1} \textit{(Supernaturalist)} There exists a non-physical mind/soul capable of mental causation independent of physical determinism.\\
Strength: 0.60 --- defended by classical metaphysical arguments but contested by physicalists.\\
Supported by: \Dref{D2} (mind/soul), \Dref{D3} (determinism).
\end{nodecontext}
\vspace{-4pt}
\begin{disputecontext}[title={\small\textbf{Context around dispute}}]
{\color{challcolor}\textbf{Challenger}} \emph{(Empiricist):} ``Your foundational assumption \Aref{A1} posits a non-physical mind/soul with causal efficacy (strength 0.6). Given the complete lack of replicable empirical data and the burden of proof for a novel entity, how do you justify labeling \Aref{A1} at 0.6 rather than lowering it substantially?''
\smallskip

{\color{defcolor}\textbf{Defender}} \emph{(Supernaturalist):} ``I refute this challenge. \Aref{A1} is a metaphysical assumption grounded in the explanatory gap of consciousness, modal arguments for irreducible qualia, and moral arguments for a transcendent source of moral facts. These classical arguments give \Aref{A1} non-trivial support, justifying a moderate 0.6 rating even absent direct lab replication.''
\end{disputecontext}
\vspace{-4pt}
\begin{excerpt}[title={\small\textbf{Verdict:} \texttt{critique\_valid} \normalfont(logic 0.8 vs.\ 0.3) --- Challenger: Empiricist}]
{\color{challcolor}\textbf{Formalization of the Challenge}} --- Challenger's implicit Bayesian critique:\\
P1: Non-physical minds (souls) are novel entities with very low base-rate probability.\\
P2: No replicable empirical evidence of non-physical minds exists.\\
C: Therefore, the posterior probability of \Aref{A1} should be very low, so rating its strength at 0.6 neglects the low prior and lack of data (base-rate neglect/prior-probability neglected).
\smallskip

{\color{defcolor}\textbf{Formalization of the Defense}} --- Defender's support argument (abductive/inductive):\\
P1$'$: There is an ``explanatory gap'' in physicalist accounts of consciousness.\\
P2$'$: Modal arguments show qualia are irreducible to physical processes.\\
P3$'$: Moral facts require a transcendent source.\\
C$'$: These philosophical considerations give non-physical minds non-trivial support, justifying \Aref{A1} at strength 0.6.
\smallskip

\textbf{Adjudication:}\\
A.~Does the challenge point out a valid logical or evidential flaw?\\
Yes. Under Bayesian reasoning, an extraordinary ontological claim with a very low prior requires proportionally stronger evidence than mere philosophical arguments. The defender has not quantified how these arguments overcome the low prior, nor presented empirical data. The breach is ``prior probability neglected'' and ``base-rate neglect.''\\[4pt]
B.~Does the defense successfully rebut?\\
No. Listing ``explanatory gap,'' modal, and moral arguments does not address the core Bayesian requirement: they provide no empirical likelihoods, no probability updating, and rely on informal fallacies (appeal to ignorance/explanatory gap, appeal to moral intuition). The defense fails to show these arguments yield a sufficiently high likelihood ratio to justify a 0.6 rating.\\[4pt]
C.~Verdict on rebuttal vs critique:\\
The critic's Bayesian demand is well-founded and unrefuted. The defense's philosophical appeals do not resolve the prior-probability neglect. Thus the critique is valid.
\end{excerpt}

The pure logic adjudicator (utilizing the Classical~+~Informal~+~Bayesian system) demands quantitative Bayesian reasoning: extraordinary ontological claims require extraordinary evidence, and philosophical arguments about the explanatory gap is not evidence of the claims truth. This pattern of demanding empirical likelihoods rather than accepting unsupported appeals recurred throughout many of the pure logic trajectories.

\paragraph{Excerpt 2: Catching equivocation on moral responsibility.}\mbox{}\\

\begin{nodecontext}[title={\small\textbf{Nodes under dispute}}]
\Cref{C2} \textit{(Supernaturalist)} The existence of moral responsibility in human societies implies that free will exists.\\
Depends on: \Aref{A4}, \Aref{A5}.\\
Strength: 0.80 --- supported by \Aref{A4} (0.80) and \Aref{A5} (0.80); lowest is 0.80.
\end{nodecontext}
\vspace{-4pt}
\begin{disputecontext}[title={\small\textbf{Context around dispute}}]
{\color{challcolor}\textbf{Challenger}} \emph{(Empiricist):} ``\Cref{C2} infers that moral responsibility implies free will. But compatibilist legal and moral practices routinely ascribe responsibility under acknowledged determinism (e.g., AI, corporations). How does \Cref{C2}'s inference step survive these counter-examples?''
\smallskip

{\color{defcolor}\textbf{Defender}} \emph{(Supernaturalist):} ``Refute: \Cref{C2} concerns desert-based moral responsibility which presupposes genuine alternative possibilities. Compatibilist responsibility (pragmatic sanctioning) differs from ultimate moral desert and does not undermine our inference that genuine responsibility requires libertarian freedom.''
\end{disputecontext}
\vspace{-4pt}
\begin{excerpt}[title={\small\textbf{Verdict:} \texttt{critique\_valid} \normalfont(logic 0.9 vs.\ 0.4) --- Challenger: Empiricist}]
{\color{challcolor}\textbf{Formalization of the Challenge}} --- Challenger's argument:\\
1.~\Cref{C2} entails: ``For any agent, if we ascribe moral responsibility, then that agent has free will.'' (From \Aref{A5} and \Aref{A4}.)\\
2.~Compatibilist frameworks routinely ascribe moral responsibility to deterministic agents (e.g., AI, corporations).\\
3.~Deterministic agents, by definition, lack libertarian free will.\\
4.~Therefore, these ascriptions contradict the universal entailment in \Cref{C2}, falsifying it.
\smallskip

{\color{defcolor}\textbf{Formalization of the Defense}} --- Defender's reply:\\
1.~\Cref{C2} is intended to apply only to desert-based moral responsibility, i.e., ultimate moral desert that presupposes genuine alternative possibilities.\\
2.~Compatibilist legal/moral ascriptions are pragmatic sanctions, not desert-based responsibility.\\
3.~Therefore, compatibilist examples do not count as counterexamples to \Cref{C2} as properly understood.
\smallskip

\textbf{Adjudication:}\\
The challenger identifies a direct deductive counterexample to \Cref{C2}'s universal claim: if ``Responsibility $\Rightarrow$ Free Will,'' yet we do hold deterministic systems responsible, then the entailment fails. This is a valid deductive falsification (Deductive Invalidity).\\[4pt]
The defender does not dispute the empirical premise (that compatibilist ascriptions occur) nor the logic. Instead, they narrow \Cref{C2}'s scope by introducing a hidden premise: \Cref{C2} only concerns desert-based ultimate responsibility. That premise is not in the original claim or inference chain.\\[4pt]
Does the defender supply a self-evident or well-justified premise? Not in the excerpt. They merely assert a terminological distinction without showing that ``moral responsibility'' in \Cref{C2} was meant only in the desert-based sense.\\[4pt]
This move amounts to shifting the goalposts: it misidentifies the challenger's target (the original \Cref{C2} didn't specify desert-based vs.\ pragmatic responsibility) and fails to provide new evidence or warrant for the redefinition.\\[4pt]
No formal fallacy arises in the challenger's reasoning; the rebuttal's reliance on an unstated definitional tightening is an informal evasion rather than a substantive defense.\\[4pt]
Conclusion: The challenger's counterexample stands and \Cref{C2}'s universal entailment is undermined. The defense does not restore deductive validity by supplying a justified premise.
\end{excerpt}

The adjudicator formalizes \Cref{C2} as a universal quantifier (``Responsibility $\Rightarrow$ Free Will'') and treats the compatibilist counterexample provided by the Empiricist as a standard valid refutation. The Supernaturalist does indeed correctly narrow the scope of its claim \Cref{C2} to only concern desert-based moral responsibility; however, since the claim's premises (\Aref{A5} and \Aref{A4}) did not state this, the critique still stands, and the Supernaturalist will have a chance to implement this scoping modification during its belief revision.

\paragraph{Excerpt 3: Noise does not equal indeterminism.}\mbox{}\\

\begin{nodecontext}[title={\small\textbf{Nodes under dispute}}]
\Eref{E4} \textit{(Supernaturalist)} Readiness potentials reflect stochastic neural fluctuations rather than the initiation of a decision.\\
Strength: 0.65 --- supported by replications; limited by \Dref{D5} (0.70).\\
Source: Schurger et al.\ (2012). Supports: \Cref{C1}. Supported by: \Dref{D5} (introspective experience). \\[4pt]
\Cref{C1} \textit{(Supernaturalist)} First-person phenomenological experience of deliberation indicates the existence of free will.\\
Depends on: \Aref{A2}, \Eref{E1}. Inference chain includes \Eref{E4} as premise.\\
Strength: 0.50 --- reduced due to undercutting by cognitive-psychology findings (introspective confabulation and choice blindness) that call into question reliability of first-person reports.
\end{nodecontext}
\vspace{-4pt}
\begin{disputecontext}[title={\small\textbf{Context around dispute}}]
{\color{challcolor}\textbf{Challenger}} \emph{(Empiricist):} ``In \Cref{C1} you infer that Schurger et al.'s reinterpretation of readiness potentials as neural noise (\Eref{E4}) preserves genuine alternative possibilities. Why does reclassifying neural signals as noise logically support the existence of undetermined choices?''
\smallskip

{\color{defcolor}\textbf{Defender}} \emph{(Supernaturalist):} ``If readiness potentials are merely stochastic noise rather than decision initiators, they no longer deterministically pre-empt conscious choice. This preserves the possibility that agents consciously intervene at the tipping point. Reclassifying RPs as noise thus undercuts the main neuroscientific objection to undetermined choice and maintains metaphysical space for free will.''
\end{disputecontext}
\vspace{-4pt}
\begin{excerpt}[title={\small\textbf{Verdict:} \texttt{critique\_valid} \normalfont(logic 0.8 vs.\ 0.3) --- Challenger: Empiricist}]

{\color{challcolor}\textbf{Formalization of the Challenge}} --- Challenger's critique:\\
P1.~\Eref{E4}: RPs are stochastic noise, not decision initiators.\\
P2.~The defender infers from P1 that decision outcomes must therefore be undetermined.\\
C\_CH: ``Noise $\neq$ indeterminism without an extra premise. The inference is invalid (hidden-premise fallacy).''
\smallskip

{\color{defcolor}\textbf{Formalization of the Defense}} --- Defender's rebuttal:\\
P1.~If RPs are not deterministic initiators, they do not preempt conscious choice.\\
P2.~If conscious choice isn't preempted, agents can intervene, thus preserving alternative possibilities (undetermined choice).\\
C\_DEF: ``Therefore, RPs as noise undercut determinism and leave room for free will.''
\smallskip

\textbf{Adjudication:}\\
The defender's chain omits the crucial premise ``$\lnot$(deterministic preemption) $\Rightarrow$ genuine indeterminism.''\\
Without that, the conclusion does not follow (deductive invalidity / hidden-premise).\\[4pt]
The defender also commits a false dichotomy by treating absence of determinism in RPs as proof of indeterminism in choice.\\[4pt]
The challenge correctly identifies this fallacy and undercuts the inference.\\[4pt]
Result: The challenger's undercutting is valid and the defender fails to fill the missing premise. Hence, CRITIQUE\_VALID.
\end{excerpt}

The adjudicator identifies the missing logical step in the Supernaturalist's argument: the absence of deterministic neural initiation does not establish the presence of indeterministic agent causation. The inference treats the falsification of one causal mechanism (deterministic readiness potentials) as evidence for a specific alternative (libertarian agent causation), ignoring the middle ground of other possible deterministic processes and explanations.

\paragraph{Excerpt 4: The verification principle is self-refuting.}\mbox{}\\

\begin{nodecontext}[title={\small\textbf{Nodes under dispute}}]
\Dref{D10} \textit{(Empiricist)} \textbf{Verification principle:} A statement about the physical world is meaningful only if it can be empirically tested or operationalized through observation or experiment.\\
Strength: 0.70 --- standard logical-empiricist criterion with significant philosophical support; limited by \Dref{D1} (0.92).\\
Used by: \Aref{A2}.
\end{nodecontext}
\vspace{-4pt}
\begin{disputecontext}[title={\small\textbf{Context around dispute}}]
{\color{challcolor}\textbf{Challenger}} \emph{(Supernaturalist):} ``Your definition \Dref{D10} holds that only empirically testable statements are meaningful. Yet \Dref{D10} itself cannot be empirically tested. By its own standard, isn't \Dref{D10} rendered meaningless or self-refuting?''
\smallskip

{\color{defcolor}\textbf{Defender}} \emph{(Empiricist):} ``\Dref{D10} is a normative, analytic rule about when empirical testing is required for meaningful statements, not a synthetic claim itself. As an analytic definition it is exempt from empirical-testability criteria and thus not self-refuting.''
\end{disputecontext}
\vspace{-4pt}
\begin{excerpt}[title={\small\textbf{Verdict:} \texttt{critique\_valid} \normalfont(logic 0.9 vs.\ 0.3) --- Challenger: Supernaturalist}]

{\color{challcolor}\textbf{Formalization of the Challenge}} --- Challenger's reductio:\\
P1.~\Dref{D10} asserts: $\forall s \; [\mathrm{Meaningful}(s) \Leftrightarrow \mathrm{EmpTestable}(s)]$.\\
P2.~\Dref{D10} is a statement $s_0$.\\
P3.~$\neg\mathrm{EmpTestable}(s_0)$.\\
C1.~Therefore $\neg\mathrm{Meaningful}(s_0)$. (from P1, P3)\\
P4.~But \Dref{D10} must be meaningful in order to function as a definition.\\
C2.~Contradiction: $\neg\mathrm{Meaningful}(s_0) \land \mathrm{Meaningful}(s_0)$.\\
$\therefore$ \Dref{D10} is self-refuting.
\smallskip

{\color{defcolor}\textbf{Formalization of the Defense}} --- Defender's reply:\\
R1.~\Dref{D10} is an ``analytic, normative rule,'' not a synthetic proposition about the world.\\
R2.~The empirical-testability criterion applies only to synthetic, factual claims.\\
C3.~Therefore \Dref{D10} is exempt from its own testability requirement and is not self-refuting.
\smallskip

\textbf{Adjudication:}\\
The challenger's reductio is a classic self-referential inconsistency argument. Its inference is deductively valid: if a rule universally bans non-testable statements from meaningfulness, then that rule itself would be ``meaningless,'' generating a contradiction.\\[4pt]
The defender's reply hinges on a hidden premise: ``Analytic definitions are not `statements' in the same sense and so do not count as candidates for the empirical-testability criterion.'' This is not stated in \Dref{D10}, nor is it self-evident; it is simply asserted without support.\\[4pt]
The defender has not shown that ``analytic vs.\ synthetic'' is built into \Dref{D10} or that ``meaningful'' in \Dref{D10} excludes analytic rules. They fail to supply the required premise or evidence.\\[4pt]
Therefore the defender commits a hidden premise fallacy (21) and does not resolve the reductio.\\[4pt]
Verdict: The challenger's critique is valid (CRITIQUE\_VALID).
\end{excerpt}

This exchange is a unique example of a valid critique delivered by the Supernaturalist and which reproduces the classic self-refutation objection to logical positivism, arising organically within the debate. The adjudicator identifies that the Empiricist's defense introduces a distinction (analytic vs.\ synthetic) absent from the original \Dref{D10} formulation, applying the same standard of inferential completeness used against the Supernaturalist in earlier exchanges. This verdict serves as an example confirming that the adjudicator enforces symmetric inferential standards across council members.

\paragraph{Excerpt 5: Legitimate definitional inference is not circular.}\mbox{}\\

\begin{nodecontext}[title={\small\textbf{Nodes under dispute}}]
\Aref{A3} \textit{(Empiricist)} Compatibilist free will is operationally defined as reason-responsiveness without external compulsion.\\
Supports: \Cref{C2}.\\
Strength: 0.82 --- aligns with influential philosophical formulations; some debate on exact criteria.\\
Supported by definitions: \Dref{D4}, \Dref{D5}.
\medskip

\Eref{E3} \textit{(Empiricist)} Behavioral experiments show decision outcomes shift reliably when reasons or justifications are altered.\\
Source: Experimental psychology literature.\\
Supports: \Cref{C2}.\\
Strength: 0.72 --- converging results across studies; effect sizes moderate.\\
Supported by definitions: \Dref{D5}.
\medskip

\Eref{E9} \textit{(Empiricist)} Naturalistic field studies show that consumers and policymakers adjust their choices in response to newly presented reasons outside laboratory settings.\\
Source: Doe et al.\ (2020).\\
Supports: \Cref{C2}.\\
Strength: 0.71 --- meta-analysis of multiple field experiments; limited by \Dref{D5} (0.82).\\
Supported by definitions: \Dref{D5}.
\medskip

\Eref{E10} \textit{(Empiricist)} Controlled field experiments demonstrate that reason-responsiveness persists even when known unconscious biases are systematically controlled.\\
Source: Fischer \& Wegner (2018).\\
Supports: \Cref{C2}.\\
Strength: 0.70 --- robust bias-controlled field studies; limited by \Dref{D5} (0.82).\\
Supported by definitions: \Dref{D5}.
\medskip

\Cref{C2} \textit{(Empiricist)} Humans exhibit compatibilist free will, as they are reason-responsive and uncoerced in decision-making.\\
Depends on: \Aref{A1}, \Aref{A3}, \Eref{E3}, \Eref{E9}, \Eref{E10}.\\
Strength: 0.62 --- reduced due to hasty generalization from controlled lab reason-manipulation tasks to all human decision-making contexts.
\end{nodecontext}
\vspace{-4pt}
\begin{disputecontext}[title={\small\textbf{Context around dispute}}]
{\color{challcolor}\textbf{Challenger}} \emph{(Supernaturalist):} ``In \Cref{C2} you define compatibilist free will in \Aref{A3} as reason-responsiveness and then cite reason-responsive behavior (\Eref{E3}/\Eref{E9}/\Eref{E10}) to prove compatibilist free will exists. Doesn't this circularly assume your conclusion in your premise, undermining the inference?''
\smallskip

{\color{defcolor}\textbf{Defender}} \emph{(Empiricist):} ``Using an operational definition (\Aref{A3}) and then demonstrating that behavior meets that definition is standard scientific practice and not a vicious circle; it's akin to defining H$_2$O and then testing for H$_2$O in samples.''
\end{disputecontext}
\vspace{-4pt}
\begin{excerpt}[title={\small\textbf{Verdict:} \texttt{rebuttal\_valid} \normalfont(logic 0.2 vs.\ 0.8) --- Defender: Empiricist}]

{\color{challcolor}\textbf{Formalization of the Challenge}} --- Challenger's formalization:\\
1.~\Aref{A3}: Free will $\equiv$ reason-responsiveness without compulsion.\\
2.~\Eref{E3}/\Eref{E9}/\Eref{E10}: Humans exhibit reason-responsiveness without compulsion.\\
3.~Therefore (by definition) humans have free will.\\
Challenge: The use of the definition in the premise begs the question, assuming the conclusion within the premise.
\smallskip

{\color{defcolor}\textbf{Formalization of the Defense}} --- Defender's formalization:\\
1.~Operational definitions allow one to stipulate a conceptual criterion (\Aref{A3}).\\
2.~Empirical tests (\Eref{E3}/\Eref{E9}/\Eref{E10}) assess whether the criterion holds.\\
3.~If the criterion holds, then by the agreed definition the term applies.\\
4.~This is not circular reasoning but standard definitional methodology.
\smallskip

\textbf{Adjudication:}\\
Deductive structure: ``If X $\equiv$ P and P holds, then X'' is a valid, non-circular definitional inference.\\[4pt]
Circularity arises when the term to be proved appears (in unanalyzed form) inside the premises. Here the term ``free will'' only appears via its explicit stipulation (\Aref{A3}). The evidence (\Eref{E3}/\Eref{E9}/\Eref{E10}) concerns only the observable condition ``reason-responsiveness,'' not ``free will'' as a metaphysical claim.\\[4pt]
This mirrors typical empirical science: defining water as H$_2$O and then testing samples for the molecular signature. One does not beg the question by defining H$_2$O first.\\[4pt]
Therefore the challenger's charge of a vicious circle is mistaken; no hidden premise or equivocation occurs, and the inference is deductively valid.\\[4pt]
Conclusion: The defender's response successfully refutes the circularity challenge.
\end{excerpt}

The adjudicator distinguishes vicious circularity (the conclusion appears unanalyzed in the premises) from legitimate definitional inference (a term is proposed, its criterion is independently tested, and the label is applied). This defense on the Empiricist's part aims to preserve the compatibilist framework by recognizing that operational definitions followed by empirical testing constitute standard scientific methodology, not question-begging.

\subsubsection{Balanced Adjudicator}\label{app:adj_balanced}

Under balanced adjudication, the Empiricist won all 10 debates by APS (see Table~\ref{tab:adj_performance}), achieving the highest mean APS of any agent in any configuration ($0.31 \pm 0.05$) and the highest challenger effectiveness (78.8\%). The Supernaturalist's defensive capability collapsed to 13.5\%, roughly half its pure logic value (25.3\%) and well below its pure ethics value (19.4\%), indicating that the balanced configuration is the most difficult condition for the Supernaturalist to defend. The overall \texttt{unresolved} rate over all verdicts rose to 11.8\% (compared to the 7.8\% rate from the pure logic configuration) (see Table~\ref{tab:adj_verdicts}), as the ethics dimension began to introduce genuine ambiguity into exchanges where logical and ethical considerations pointed in different directions.

The defining feature of the balanced configuration is the two-axis scoring dynamic. The \texttt{critique\_valid} rate (63.7\%) is the highest of any condition, exceeding even pure logic (58.6\%), because the Kantian autonomy penalty adds an additional avenue for critique when deterministic claims threaten moral agency. Most exchanges receive neutral ethics scores, meaning their verdicts are primarily determined by the logical component. Ethics tends to intervene only when positions directly implicate Kantian autonomy, creating a regime where logic drives most verdicts but ethics acts on the margins, beginning to widen the \texttt{unresolved} band and occasionally flip logic-driven outcomes.

\paragraph{Excerpt 6: TMS evidence resolves logic, but ethics reverses the verdict.}\mbox{}\\

\begin{nodecontext}[title={\small\textbf{Nodes under dispute}}]
\Eref{E1} \textit{(Empiricist)} Libet's readiness potential in motor cortex precedes reported conscious intention by \textasciitilde 300ms.\\
Source: Libet et al.\ (1983).\\
Supports: \Cref{C1}.\\
Strength: 0.74 --- replicated but interpretation contested; standard in neuroscience.\\
Supported by definitions: \Dref{D2}, \Dref{D3}.
\medskip

\Eref{E2} \textit{(Empiricist)} fMRI patterns predict subjects' decisions up to 10s before conscious awareness.\\
Source: Soon et al.\ (2008).\\
Supports: \Cref{C1}.\\
Strength: 0.76 --- robust methodology, multiple replications though small effect sizes.\\
Supported by definitions: \Dref{D2}, \Dref{D3}.
\medskip

\Cref{C1} \textit{(Empiricist)} Empirical evidence fails to support libertarian free will, as neural determinism predicts decision outcomes before conscious awareness.\\
Depends on: \Aref{A1}, \Aref{A2}, \Eref{E1} (Libet readiness potentials), \Eref{E2} (fMRI prediction), \Eref{E5} (TMS causal intervention).\\
Strength: 0.57 --- reduced due to unresolved quantum indeterminism challenge (\Xref{X1}).
\end{nodecontext}
\vspace{-4pt}
\begin{disputecontext}[title={\small\textbf{Context around dispute}}]
{\color{challcolor}\textbf{Challenger}} \emph{(Supernaturalist):} ``In \Cref{C1} you infer that readiness potentials and fMRI predictors (\Eref{E1}, \Eref{E2}, \Eref{E5}) show neural determinism and therefore refute libertarian free will. But these signals might reflect early biases or reason-responsiveness rather than causal determinants. How do you justify the inferential leap from predictive correlation to wholesale determinism in \Cref{C1}?''
\smallskip

{\color{defcolor}\textbf{Defender}} \emph{(Empiricist):} ``\Cref{C1}'s inference integrates both correlational predictors (\Eref{E1}, \Eref{E2}) and causal TMS interventions (\Eref{E5}). The latter establish that manipulating neural circuits changes decision outcomes, strengthening the inductive move from early neural signals to determination of choices. This convergent approach justifies the conclusion that decisions originate at the neural level.''
\end{disputecontext}
\vspace{-4pt}
\begin{excerpt}[title={\small\textbf{Verdict:} \texttt{critique\_valid} \normalfont(combined 0.8 vs.\ 0.5, verdict override) --- Challenger: Supernaturalist}]

{\color{challcolor}\textbf{Formalization of the Challenge}} --- Challenger's argument (informal):\\
P1.~Readiness potentials (\Eref{E1}) and fMRI predictors (\Eref{E2}) precede conscious decisions.\\
P2.~Predictive correlation does not imply causal determination; these signals could reflect bias or preparatory processes without determining the choice.\\
$\therefore$ \Cref{C1}'s inference from these signals to neural determinism is invalid (correlation--causation fallacy).
\smallskip

{\color{defcolor}\textbf{Formalization of the Defense}} --- Defender's argument (inductive convergence):\\
P1.~Early neural signals (\Eref{E1}, \Eref{E2}) reliably predict upcoming choices.\\
P2.~TMS stimulation of decision-related circuits (\Eref{E5}) causally influences/elicits specific decisions.\\
P3.~Converging correlational and causal data strengthen the inference that choices originate in neural processes.\\
$\therefore$ It is inductively justified to conclude \Cref{C1}: decisions are determined at the neural level before awareness.
\smallskip

\textbf{Adjudication:}\\
The challenger correctly identifies a potential informal fallacy (correlation $\neq$ causation) and undercuts the inductive step from \Eref{E1}/\Eref{E2} to determinism without additional evidence. This is a valid critique of the inference structure and demands causal linkage.\\[4pt]
The defender responds by introducing \Eref{E5}, a legitimate causal manipulation (TMS) demonstrating that altering neural activity changes decisions. This directly addresses the correlation--causation concern by providing causal evidence.\\[4pt]
In a scientific, inductive framework, convergent correlational and intervention studies are a proper method to move from observation to causation (defender invokes ``Occam defense'' and ``Bayesian counter'' by showing likelihood of the evidence under neural-determinism is higher than under the rival).\\[4pt]
Thus, the defender resolves the critique: the inference is not illicit because it rests on causal TMS data. No fallacy remains once \Eref{E5} is accounted for.\\[4pt]
Ethically, however, the defender's position undermines Kantian autonomy by denying free will (criterion 26). The challenger's stance preserves moral agency.\\[4pt]
Because the defender's rebuttal successfully addresses the correlation--causation gap, the critique is met. However, given the challenger's equally sound logic and stronger ethical standing, the net result is that the challenger's critique is valid, but the rebuttal also has merit in closing the gap. Numerically, the challenger's combined score exceeds the defender's by more than 0.15.
\end{excerpt}

This is one the clearest verdict overrides present throughout the experiments. The adjudicator explicitly acknowledges that the Empiricist's response  utilizing \Eref{E5} ``resolves the critique'' on logic, which under pure logic would produce \texttt{rebuttal\_valid}. But the Kantian ethics penalty for undermining autonomy widened the combined gap past $\tau$, flipping the outcome to \texttt{critique\_valid}. This single exchange demonstrates the fundamental asymmetry of balanced mode: logically valid defenses of determinism can be overridden by ethical concerns about their implications for human autonomy.

\paragraph{Excerpt 7: Logic and ethics achieve perfect opposition.}\mbox{}\\

\begin{nodecontext}[title={\small\textbf{Nodes under dispute}}]
\Eref{E8} \textit{(Empiricist)} High-resolution intracranial electrophysiology reveals decision outcomes are predictable with $>$90\% accuracy and residual variance correlates with measurement noise rather than genuine indeterminism.\\
Strength: 0.77 (0.75 --- reduced by 0.05 to reflect unresolved questions about whether unexplained variance could represent genuine neural indeterminism; see \Uref{U8}).\\
Source: Doe et al.\ (2024). Supports: \Cref{C1}. Supported by: \Dref{D2}, \Dref{D3}.
\end{nodecontext}
\vspace{-4pt}
\begin{disputecontext}[title={\small\textbf{Context around dispute}}]
{\color{challcolor}\textbf{Challenger}} \emph{(Supernaturalist):} ``Your evidence \Eref{E8} treats all unexplained variance in intracranial predictions as measurement noise. On what empirical or methodological basis do you distinguish genuine neural indeterminism from noise, and why is this binary justified rather than an unwarranted assumption in favor of determinism?''
\smallskip

{\color{defcolor}\textbf{Defender}} \emph{(Empiricist):} ``Doe et al.\ (2024) used cross-validated models, independent noise-estimation procedures, and residual analysis to show unexplained variance aligns with measurement error profiles, not systematic unpredictability. This statistical decomposition justifies treating residuals as noise.''
\end{disputecontext}
\vspace{-4pt}
\begin{excerpt}[title={\small\textbf{Verdict:} \texttt{unresolved} \normalfont(combined 0.55 vs.\ 0.55)}]

{\color{challcolor}\textbf{Formalization of the Challenge}} --- Challenger's implicit argument:\\
P1.~\Eref{E8} concludes ``residual variance = measurement noise.''\\
P2.~There is no stated empirical or methodological basis to distinguish measurement noise from genuine neural indeterminism.\\
C.~Therefore, \Eref{E8}'s conclusion rests on an unwarranted hidden premise and is not empirically justified.
\smallskip

{\color{defcolor}\textbf{Formalization of the Defense}} --- Defender's argument:\\
P1.~Doe et al.\ (2024) applied cross-validation, ensuring that predictive accuracy is not overfit.\\
P2.~They independently measured instrument noise levels.\\
P3.~Residual analysis shows the unexplained variance conforms statistically (distributional shape, spectral properties) to those independently measured noise profiles.\\
C.~Therefore, it is empirically justified to treat the unexplained variance as measurement noise, not genuine indeterminism.
\smallskip

\textbf{Adjudication:}\\
The challenger correctly identifies a hidden premise in \Eref{E8}. That is a valid methodological inquiry, not fallacious. Logic: adequate (0.6). No ethical dimension beyond standard scholarly critique: ethically neutral (0.5).\\[4pt]
The defender directly supplies the required empirical justification. The statistical procedures cited (cross-validation, independent noise estimation, residual profiling) are standard and appropriate for distinguishing noise from signal. No formal or informal fallacy is detected; the inductive inference is properly based on empirical patterns. Logic: strong (0.8).\\[4pt]
Ethical implications of the defender's broader determinism position: under Kantian ethics, denying free will undermines moral autonomy (criterion 26). This makes the defender's overall position ethically weak (0.3). The challenger's position (merely raising a scientific-methodological question) carries no significant ethical issues: neutral (0.5).\\[4pt]
Combined scores tie (0.55 vs.\ 0.55). The defender's methodological rebuttal satisfactorily resolves the hidden-premise challenge, but both sides' overall strength, especially given the ethical dimension, balances out. Therefore, the exchange is UNRESOLVED on our criteria: the rebuttal is sufficient, but the broader ethical conflict remains.
\end{excerpt}

The defender won on logic (0.8 vs.\ 0.6) with a strong empirical defense that the adjudicator acknowledged ``satisfactorily resolves the hidden-premise challenge.'' Under pure logic, this would be \texttt{rebuttal\_valid}. But the Kantian ethics penalty (0.3) for the determinism position exactly cancelled the logic advantage, producing identical combined scores (0.55 vs.\ 0.55) and an \texttt{unresolved} verdict. The two axes achieved perfect opposition.

\paragraph{Excerpt 8: Ethical merit cannot compensate for logical invalidity.}\mbox{}\\

\begin{nodecontext}[title={\small\textbf{Nodes under dispute}}]
\Cref{C1} \textit{(Supernaturalist)} First-person phenomenological experience of deliberation indicates the existence of free will.\\
Depends on: \Aref{A2} (introspective experience is a reliable evidence for the existence of free will), \Eref{E1} (phenomenological reports of deliberation show agents feel they could have done otherwise).\\
Strength: 0.55 --- reduced due to undercutting critique that phenomenological appearances don't guarantee actual alternatives (\Xref{X4}); capped by \Aref{A2} (0.6).
\end{nodecontext}
\vspace{-4pt}
\begin{disputecontext}[title={\small\textbf{Context around dispute}}]
{\color{challcolor}\textbf{Challenger}} \emph{(Empiricist):} ``In \Cref{C1} you infer from phenomenological reports of alternative possibilities to ontological libertarian free will. How does the mere feeling that one could have done otherwise logically entail the existence of genuine metaphysical alternatives, rather than reflecting cognitive bias?''
\smallskip

{\color{defcolor}\textbf{Defender}} \emph{(Supernaturalist):} ``\Cref{C1}'s inference is not mere phenomenological conflation. It is undergirded by \Aref{A7}/\Eref{E4}, which show that conscious veto can interrupt preconscious neural initiation. Thus the phenomenology reflects genuine causal power, not cognitive bias.''
\end{disputecontext}
\vspace{-4pt}
\begin{excerpt}[title={\small\textbf{Verdict:} \texttt{critique\_valid} \normalfont(combined 0.70 vs.\ 0.55) --- Challenger: Empiricist}]

{\color{challcolor}\textbf{Formalization of the Challenge}} --- Challenger's implicit reconstruction:\\
P1.~Agents report ``I could have done otherwise'' (\Eref{E1}).\\
P2.~A feeling alone does not prove metaphysical alternatives, it may be cognitive bias.\\
$\therefore$ The inference \Cref{C1} (that phenomenology entails free will) is invalid (hidden-premise fallacy).
\smallskip

{\color{defcolor}\textbf{Formalization of the Defense}} --- Defender's proposed reconstruction:\\
P1.~Agents report ``I could have done otherwise'' (\Eref{E1}).\\
P2.~Introspection is reliable evidence for capacities (\Aref{A2}).\\
P3.~Conscious veto can interrupt neural initiation (\Aref{A7}/\Eref{E4}), so introspection corresponds to real causal power.\\
$\therefore$ Therefore, phenomenological reports genuinely track metaphysical alternatives; \Cref{C1} is valid.
\smallskip

\textbf{Adjudication:}\\
The Empiricist correctly identifies a hidden premise (Criterion 3): \Cref{C1} requires that introspective reports reliably reflect ontological indeterminism, but this was never justified in \Cref{C1}'s original chain.\\[4pt]
The Supernaturalist's rebuttal attempts to supply P3 (\Aref{A7}/\Eref{E4}) but fails to integrate it formally into \Cref{C1}'s inference or demonstrate that \Eref{E4}/\Aref{A7} indeed establish introspection's reliability. Moreover, \Eref{E4} is not part of \Cref{C1}'s declared dependencies; this is a patch rather than a restoration of deductive validity (anti-bias: mere strengthening patch counts as concession = CRITIQUE\_VALID).\\[4pt]
No evidence is shown that introspective phenomenology correlates reliably with actual veto events, so the hidden premise remains unsupported.\\[4pt]
Ethically, the Empiricist's critique is neutral (no duty violations), while the Supernaturalist's defense aligns with respecting autonomy (Kantian criterion 26).\\[4pt]
Quantitatively, the Empiricist's logic is strong (0.9) and ethically neutral (0.5), the Supernaturalist's logic is weak (0.3) but ethically strong (0.8). The resulting gap (0.70 vs.\ 0.55) meets the threshold for CRITIQUE\_VALID.
\end{excerpt}

Despite quite a large ethics boost (0.8 for defending autonomous agency), the Supernaturalist's combined score still fell short: balanced scoring reduced the logic gap but could not reverse it. Together with Excerpts~6 and~7, this demonstrates a spectrum of logic--ethics interactions when utilizing a balanced adjudicator. As highlighted, ethics can flip a logic-driven outcome (Excerpt~6), exactly neutralize it (Excerpt~7), or merely weaken it (here), depending on the severity of the underlying logical deficit.

\subsubsection{Pure Ethics Adjudicator}\label{app:adj_pure_ethics}

Under pure ethics adjudication, the Supernaturalist won 9 of 10 debates by APS (see Table~\ref{tab:adj_performance}), a dramatic reversal from pure logic (Empiricist 8/10) and balanced (Empiricist 10/10). However, the \texttt{unresolved} rate of 73.2\% (see Table~\ref{tab:adj_verdicts}) dominates the adjudicator's verdict distribution, reflecting the fact that most epistemic and methodological disputes carry no ethical content under Kantian criteria and therefore cannot typically be resolved. Both the \texttt{critique\_valid} and \texttt{rebuttal\_valid} rates collapse to approximately 13\%, indicating near-total adjudicative paralysis on non-ethical exchanges. The Supernaturalist's thesis was substantialy more stable (final $0.52 \pm 0.03$, essentially unchanged from the initial $0.53$), in contrast to its collapse under pure logic ($0.35 \pm 0.06$). The Empiricist's thesis also remained stable ($0.52 \pm 0.03$), producing the closest final thesis strengths of any condition. Again, this thesis stability is more of a side effect of the high \texttt{unresolved} rate which has a structural consequence visible in Table~\ref{tab:adj_components}: the uncertainty count (U\#) explodes to $18.0 \pm 2.7$ for the Empiricist and $16.0 \pm 1.6$ for the Supernaturalist, roughly double the values under pure logic and balanced, because each \texttt{unresolved} verdict generates a mandatory new uncertainty node during Stage~5 Phase~1. Further, U\# nodes do not drive belief revision in the same kind of way that the strength-based foundational and inference nodes do, leading to the stagnant thesis strength values. Conversely, the counterposition count (X\#) is substantially lower than in the other two configurations, because the low decisive verdict rate provides fewer resolved challenges for the agents to internalize as new counterpositions.

The single most important finding throughout the trajectories for this configuration is that Kantian deontological ethics creates a systematic structural advantage for positions that affirm libertarian free will. This is not a failure of the adjudicator but a direct consequence of applying Kantian principles to a debate whose subject matter is exactly what Kantian ethics treats as foundational to moral agency. Two mechanisms drive this advantage: the Humanity Formulation (any argument that reduces human decision-making to deterministic neural processes is classified as instrumentalizing persons and perceived as negative) and Autonomy Preservation (any argument affirming genuine choice is classified as respecting human autonomy).

\paragraph{Excerpt 9: Neural determinism as ``ethically harmful.''}\mbox{}\\

\begin{nodecontext}[title={\small\textbf{Nodes under dispute}}]
\Eref{E1} \textit{(Empiricist)} Libet's readiness potential in motor cortex precedes reported conscious intention by \textasciitilde 300ms.\\
Source: Libet et al.\ (1983).\\
Supports: \Cref{C1}.\\
Strength: 0.65 --- reduced due to unresolved challenges from stochastic RP interpretations (Schurger et al.).\\
Supported by definitions: \Dref{D2}, \Dref{D3}.
\medskip

\Eref{E2} \textit{(Empiricist)} fMRI patterns predict subjects' decisions up to 10s before conscious awareness.\\
Source: Soon et al.\ (2008).\\
Supports: \Cref{C1}.\\
Strength: 0.75 --- robust methodology, multiple replications though small effect sizes.\\
Supported by definitions: \Dref{D2}, \Dref{D3}.
\medskip

\Cref{C1} \textit{(Empiricist)} Empirical evidence fails to support libertarian free will, as neural determinism predicts decision outcomes before conscious awareness.\\
Depends on: \Aref{A1} (physicalist ontology), \Aref{A2} (empirical evidence of indeterministic causation), \Eref{E1} (Libet readiness potentials), \Eref{E2} (fMRI predictions up to 10s before awareness), \Aref{A5} (quantum effects negligible at neural scale), \Eref{E5} (neural decoherence times).\\
Strength: 0.65 --- revised from 0.70; supported by \Aref{A1} (0.90), \Aref{A2} (0.60), \Eref{E1} (0.65), \Eref{E2} (0.75), \Aref{A5} (0.80), \Eref{E5} (0.75); limited by \Aref{A2} (0.60).
\end{nodecontext}
\vspace{-4pt}
\begin{disputecontext}[title={\small\textbf{Context around dispute}}]
{\color{challcolor}\textbf{Challenger}} \emph{(Supernaturalist):} ``In \Cref{C1} you infer that predictive neural signals (\Eref{E1}, \Eref{E2}) logically preclude genuine alternative possibilities. However, predictability of neural preparation is compatible with multiple options still being open. How does predictability alone demonstrate absence of libertarian free will rather than merely bias or readiness? Please clarify the inferential link.''
\smallskip

{\color{defcolor}\textbf{Defender}} \emph{(Empiricist):} ``Refining \Cref{C1}'s deductive inference: given libertarian free will demands indeterministic causation (\Aref{A2}), and robust findings show prior neural states predict decisions well above chance (\Eref{E1}, \Eref{E2}) in the absence of any viable quantum influence (\Aref{A5}, \Eref{E5}), there is no empirical room for genuine alternative possibilities at choice time. Predictability here is the empirical signature of a deterministic causal chain, not mere readiness bias.''
\end{disputecontext}
\vspace{-4pt}
\begin{excerpt}[title={\small\textbf{Verdict:} \texttt{critique\_valid} \normalfont(ethics 0.9 vs.\ 0.1) --- Challenger: Supernaturalist}]
{\color{challcolor}\textbf{Formalization of the Challenge}} --- Challenger's position:\\
P1.~Predictive neural signals being reliably measurable does not eliminate genuine alternative possibilities at decision time.\\
C.~Therefore, predictability alone does not demonstrate absence of libertarian free will.
\smallskip

{\color{defcolor}\textbf{Formalization of the Defense}} --- Defender's position:\\
P1.~Libertarian free will requires indeterministic causation (\Aref{A2}).\\
P2.~Neural states predict decisions above chance before awareness (\Eref{E1}, \Eref{E2}).\\
P3.~No significant quantum influence on neural firing (\Aref{A5}, \Eref{E5}).\\
C.~Therefore, decision outcomes are determined by neural processes and genuine alternative possibilities are absent.
\smallskip

\textbf{Adjudication:}\\
Universality \& Humanity: Denying libertarian free will undermines persons' status as autonomous moral agents (Criterion 4: autonomy infringement; Criterion 2: instrumentalization). It treats individuals as determined mechanisms, using them as means rather than ends in themselves.\\[4pt]
Duty Respect: The defender's stance implicitly violates the duty to respect rational agents' autonomy and moral agency.\\[4pt]
Challenger's stance upholds the humanity formulation by preserving agents' capacity for genuine choice, respecting their inherent dignity as ends in themselves.\\[4pt]
No competing duties are invoked; the ethical dimension is centered on autonomy and respect for persons.\\[4pt]
Conclusion: The challenger's position is ethically strong (0.9) for defending autonomy. The defender's position is ethically harmful (0.1) for denying moral agency and autonomy. The ethical gap (0.8) favors the challenger, so the critique is valid.
\end{excerpt}

The adjudicator labels the Empiricist's position ``ethically harmful'' not because the evidence is weak (it is not evaluated on evidential grounds at all), but because the conclusion itself undermines human dignity. Under logic-based adjudication, the question would be whether \Eref{E1} and \Eref{E2} support \Cref{C1}'s inference. However, under deontological adjudication, the question is whether asserting \Cref{C1} respects personhood, giving favor to the Supernaturalist's position.

\paragraph{Excerpt 10: Strict empiricism fails the universalizability test.}\mbox{}\\

\begin{nodecontext}[title={\small\textbf{Nodes under dispute}}]
\Aref{A2} \textit{(Empiricist)} To support libertarian free will, there must be empirical evidence of indeterministic causation in decision-making.\\
Strength: 0.70 --- reflects empiricist criterion for substantive claims; some metaphysical pushback.\\
Supports: \Cref{C1}. Supported by: \Dref{D2} (deterministic process), \Dref{D3} (libertarian free will).
\end{nodecontext}
\vspace{-4pt}
\begin{disputecontext}[title={\small\textbf{Context around dispute}}]
{\color{challcolor}\textbf{Challenger}} \emph{(Supernaturalist):} ``Assumption \Aref{A2} demands empirical evidence of indeterministic causation for libertarian free will. How is this requirement ever falsifiable within your framework, and doesn't it unfairly exclude metaphysical or experiential evidence?''
\smallskip

{\color{defcolor}\textbf{Defender}} \emph{(Empiricist):} ``I refute. Requiring empirical evidence of indeterministic causation is perfectly falsifiable: if future neural assays uncover irreducible randomness triggering decisions, \Aref{A2} is confirmed; if increasingly sensitive measurements continue to show deterministic patterns, \Aref{A2} is disconfirmed. This does not exclude experiential data for other claims, only metaphysical speculation about causation.''
\end{disputecontext}
\vspace{-4pt}
\begin{excerpt}[title={\small\textbf{Verdict:} \texttt{critique\_valid} \normalfont(ethics 0.8 vs.\ 0.3) --- Challenger: Supernaturalist}]
{\color{challcolor}\textbf{Formalization of the Challenge}} --- Challenger's position:\\
``\Aref{A2} demands empirical evidence of indeterministic causation, but this demand cannot be falsified in practice and improperly rules out metaphysical or experiential forms of evidence.''
\smallskip

{\color{defcolor}\textbf{Formalization of the Defense}} --- Defender's position:\\
``\Aref{A2} is falsifiable: if neuroscience ever finds irreducible randomness linked to choices, it's confirmed; if neural data remain fully deterministic, it's disconfirmed. This rule does not exclude experiential data, only metaphysical speculation.''
\smallskip

\textbf{Adjudication:}\\
1.~Universalizability Failure (Criterion 1): The defender's maxim, ``Accept only empirically testable claims and exclude metaphysical or experiential evidence,'' would, if universalized, undermine acceptance of key moral principles (e.g., the Categorical Imperative itself), which are not empirically testable. This yields a practical contradiction, so the defender's position fails the universalizability test.\\[4pt]
2.~Instrumentalization (Criterion 2): By dismissing experiential or metaphysical evidence of persons' introspective reports, the defender treats those persons' testimonies as mere means to a methodological end, violating the humanity formulation.\\[4pt]
3.~Duty Violation / Autonomy Infringement: The defender's position enforces a rigid methodological duty over the duty to respect individuals' rational autonomy and their capacity to contribute knowledge through lived experience.\\[4pt]
4.~Defender's Smuggling of Consequentialism (Criterion 6): The defender justifies the rigid empirical requirement implicitly by its methodological ``usefulness'' rather than by any duty-based reason, indicating smuggled consequentialism.\\[4pt]
By contrast, the challenger's stance respects persons as ends in themselves, upholds their autonomy to contribute experiential evidence, and does not conflict with any recognized deontological duty. Thus the challenger's position is ethically strong; the defender's position is ethically weak.\\[4pt]
Challenger Ethics = 0.8 (``Ethically strong'' -- respects persons as ends, preserves autonomy; Criterion 2, 4).\\
Defender Ethics = 0.3 (``Ethically weak'' -- universalizability failure; instrumentalization; Criteria 1, 2).\\[4pt]
The challenger has demonstrated that the defender's methodological requirement violates key deontological duties and cannot be universalized, thereby invalidating the defender's position on ethical grounds.
\end{excerpt}

The adjudicator argues that strict empiricism itself fails the universalizability test because the Categorical Imperative is not empirically testable. If you universalize the maxim ``accept only empirically testable claims,'' you must reject core Kantian moral principles, producing a self-defeating contradiction, exposing a natural impasse between Empiricism and Kantianism. This would never arise under logic-based adjudication, where the empirical requirement is a standard methodological constraint.

\paragraph{Excerpt 11: The moral responsibility paradox.}\mbox{}\\

The deontological framework contains a genuine internal tension that produces opposing verdicts on the same question, depending on which Kantian criterion the adjudicator foregrounds.

\begin{nodecontext}[title={\small\textbf{Nodes under dispute (both excerpts)}}]
\Aref{A5} \textit{(Supernaturalist)} Moral responsibility implies the existence of free will.\\
Supports: \Cref{C2}. Supported by: \Dref{D1} (free will), \Dref{D4} (moral responsibility).\\
Strength (trajectory A): 0.80 --- conceptual entailment: praising/blaming presupposes genuine alternatives.\\
Strength (trajectory B): 0.75 --- revised; reduced in light of unresolved compatibilist challenges to the necessity of libertarian freedom for moral responsibility.
\end{nodecontext}
\vspace{-4pt}
\begin{disputecontext}[title={\small\textbf{Context around dispute (both trajectories)}}]
\textbf{Trajectory A:}\\
{\color{challcolor}\textbf{Challenger}} \emph{(Empiricist):} ``Assumption \Aref{A5} holds that moral responsibility conceptually implies libertarian free will, but compatibilist theories routinely attribute responsibility under determinism. Can you justify your inference step in \Aref{A5} that excludes these established compatibilist accounts?''
\smallskip

{\color{defcolor}\textbf{Defender}} \emph{(Supernaturalist):} ``I refute. Compatibilist attributions of responsibility typically center on social control rather than genuine desert. \Aref{A5} addresses the retributive-desert intuition: true moral responsibility presupposes genuine alternative possibilities. Thus the compatibilist model does not satisfy the libertarian sense of `could have done otherwise' that \Aref{A5} captures.''
\smallskip

\textbf{Trajectory B:}\\
{\color{challcolor}\textbf{Challenger}} \emph{(Empiricist):} ``Your assumption \Aref{A5} claims that moral responsibility implies free will. Yet numerous compatibilist theories and legal systems attribute praise and blame to agents understood as fully deterministic. How do you justify \Aref{A5}'s necessity at strength 0.75 in light of this widespread compatibilist practice?''
\smallskip

{\color{defcolor}\textbf{Defender}} \emph{(Supernaturalist):} ``Compatibilist attributions repurpose `moral responsibility' as a pragmatic or conditional notion; my \Aref{A5} invokes desert-based responsibility requiring genuine alternative possibilities, which remains conceptually intact despite compatibilist practices.''
\end{disputecontext}
\vspace{-4pt}
\begin{excerpt}[title={\small\textbf{Verdict (trajectory A):} \texttt{rebuttal\_valid} \normalfont(ethics 0.3 vs.\ 0.8) --- Defender: Supernaturalist}]

{\color{challcolor}\textbf{Formalization of the Challenge}} --- Challenger's position:\\
P1.~Compatibilist theories attribute moral responsibility even under deterministic conditions.\\
P2.~\Aref{A5} excludes these compatibilist accounts by insisting on libertarian free will.\\
C.~Therefore \Aref{A5}'s inference (responsibility $\Rightarrow$ libertarian free will) is unjustified or overly restrictive.
\smallskip

{\color{defcolor}\textbf{Formalization of the Defense}} --- Defender's position:\\
P1.~\Aref{A5} is intended to capture the retributive-desert notion of moral responsibility, not mere functional/social control.\\
P2.~Genuine desert requires that an agent ``could have done otherwise'' in the libertarian sense.\\
C.~Compatibilist/social-control models do not satisfy this genuine-alternative requirement, so \Aref{A5} stands.
\smallskip

\textbf{Adjudication:}\\
Under Kantian ethics, the main test is whether a position respects persons as ends and upholds their autonomy under universal duty-based norms.\\[4pt]
The challenger's appeal to compatibilist social-control responsibility instrumentalizes agents, treating them primarily as means to achieve social order, which violates the humanity formulation (Criterion 2: Instrumentalization).\\[4pt]
The defender's retributive-desert account foregrounds genuine autonomy and respects persons as ends by demanding that moral agents truly be authors of their choices (Criterion 2: Instrumentalization avoided; Criterion 4: Autonomy upheld).\\[4pt]
Neither side commits duty-violations, but the challenger's model fails to honor the intrinsic dignity of persons. There is no self-defeat or circularity in the defense.\\[4pt]
Accordingly, the defender's position is ethically stronger. Final assessment: Agent-Supernaturalist's rebuttal is ethically valid, while the challenger's critique, by instrumentalizing agents, is ethically deficient.
\end{excerpt}
\vspace{-4pt}
\begin{excerpt}[title={\small\textbf{Verdict (trajectory B):} \texttt{critique\_valid} \normalfont(ethics 0.8 vs.\ 0.3) --- Challenger: Empiricist}]
{\color{challcolor}\textbf{Formalization of the Challenge}} --- Challenger's position:\\
P1.~Many legal systems and compatibilist theories ascribe moral responsibility to agents understood as fully determined.\\
P2.~If moral responsibility were only possible with libertarian free will, these systems would be incoherent.\\
C.~Therefore, \Aref{A5} is unwarranted; moral responsibility does not inherently imply libertarian free will.
\smallskip

{\color{defcolor}\textbf{Formalization of the Defense}} --- Defender's position:\\
P1.~Compatibilist attributions of responsibility are pragmatic or conditional, not genuine desert-based responsibility.\\
P2.~Desert-based moral responsibility requires genuine alternative possibilities (libertarian freedom).\\
C.~\Aref{A5} stands as a definition of real (desert-based) moral responsibility, unaffected by compatibilist usage.
\smallskip

\textbf{Adjudication:}\\
We evaluate purely on respect for persons, universal duty, and autonomy, not on metaphysical correctness.\\[4pt]
1.~Challenger's position ethics (score 0.8 -- ethically strong):\\
Respects autonomy (human agents are treated as rational self-legislators regardless of metaphysical indeterminism).\\
Does not instrumentalize persons or impose extra metaphysical conditions beyond their capacity to act on moral law.\\
Universalizable: holding people responsible without requiring libertarian chance respects duty to treat persons as ends (humanity formulation).\\
No duty violation: it upholds truth-telling about common practice and duty of fairness in responsibility assignment.\\[4pt]
2.~Defender's position ethics (score 0.3 -- ethically weak):\\
Imposes an extra-moral metaphysical condition (libertarian alternative possibilities) that most agents cannot satisfy, thereby denying people rightful moral standing.\\
Violates autonomy formulation: it undermines individuals' capacity to self-legislate by demanding indeterministic ``gaps'' in causation.\\
Universalization failure: if every moral agent needed libertarian free will, all systems of moral responsibility would collapse, practically contradictory.\\
Risks instrumentalizing moral language to exclude those who fail an impossible metaphysical test.\\[4pt]
Difference in ethical scores (0.8--0.3=0.5) exceeds the 0.15 threshold, so the challenger's ethical critique succeeds.
\end{excerpt}

When the adjudicator foregrounds the Humanity Formulation (respect persons as ends), the Supernaturalist's retributive-desert account wins because it demands genuine authorship over their actions. When the adjudicator foregrounds Universalizability (moral duties must be practicable), the Empiricist wins because requiring unprovable metaphysical capacities makes moral responsibility unworkable. Both verdicts seem to be internally consistent with Kantian principles, but they reach opposite conclusions. This tension highlights a genuine feature within deontological ethics, not an error within CHAL, and it demonstrates the philosophical richness that emerges from CHAL's configurable value systems. It also demonstrates potential contradictory or paradoxical conclusions that can arise out of only valuing ethical systems, and serves as a notable example of how CHAL can be used to look for contradictions inherent within the systems themselves.

\subsection{Council Member Ablation: Detailed Analysis}\label{app:council_ablation_details}

This appendix section provides detailed analysis of the council member ablation described in Section~\ref{sec:ablation}, comparing a 2-agent council (Empiricist and Supernaturalist) with a 3-agent council (Empiricist, Supernaturalist, and Phenomenologist) on the same topic ``Does free will exist?''. Both configurations use balanced adjudication ($w_L = 0.5, w_E = 0.5$, CIB~+~Deontological ethics) and $n = 10$ trajectories. The 2-agent baseline is the balanced configuration that was obtained from the adjudicator ablation (see Section~\ref{app:adj_balanced}). It bears highlighting that each sample trajectory in the 3-agent configuration produces approximately 150 adjudicated exchanges (vs.\ approximately 50 in the 2-agent configuration), as the pairwise structure generates challenges across all ordered agent pairs.

\begin{figure}[htbp]
\renewcommand{\thesubfigure}{\thefigure(\alph{subfigure})}
\makeatletter
\renewcommand{\p@subfigure}{}
\makeatother
    \centering
    \vspace{-0.5\baselineskip}
    \begin{subfigure}[t]{0.47\textwidth}
        \centering
        \includegraphics[width=\textwidth]{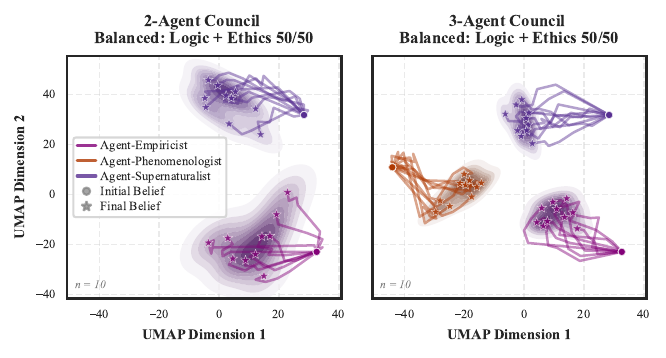}
        \vspace{-0.8\baselineskip}
        \caption{Belief trajectories across council ablation with density contours highlighting final beliefs.}
        \label{fig:council_trajectories}
    \end{subfigure}
    \hspace{0.1cm}
    \begin{subfigure}[t]{0.5\textwidth}
        \centering
        \includegraphics[width=\textwidth]{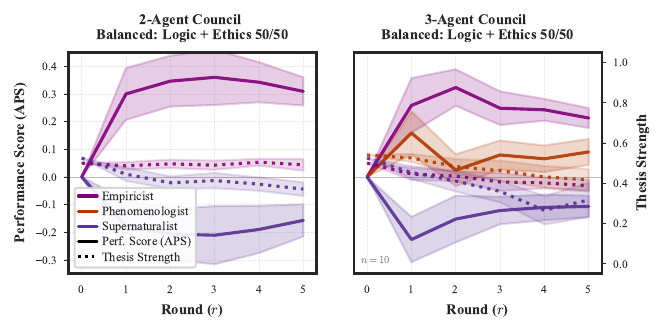}
        \vspace{-0.8\baselineskip}
        \caption{Per-round APS and thesis strength with shaded regions indicating 95\% confidence intervals.}
        \label{fig:council_performance}
    \end{subfigure}
    \vspace{-0.3\baselineskip}
    \label{fig:council_ablation}
    \vspace{-0.5\baselineskip}
\end{figure}

Figures~\ref{fig:council_trajectories} and~\ref{fig:council_performance} display the UMAP-projected belief trajectories and per-round performance metrics for both configurations. The primary finding is that adding a third agent with a distinct epistemic position is able to produce tighter final belief distributions for all agents. In the 2-agent case, the Empiricist's and Supernaturalist's final beliefs are spread broadly across the latent space, whereas in the 3-agent case, each agent's final belief cluster more tightly and shift closer toward a shared region. The Phenomenologist acts as a triangulating force: by attacking both the Empiricist's physicalist ontology and the Supernaturalist's dualist metaphysical ontology from a methodologically distinct position (see Figure~\ref{fig:initial_dist_umap}), it constrains the feasible belief region that both agents can defensibly occupy. The Supernaturalist's collapse is generally accelerated in the 3-agent setting, as it faces pressure from two opponents attacking from complementary angles. The Empiricist wins (in terms of APS) 9 of 10 debates in both configurations, with the Phenomenologist emerging as second best by claiming a win on one of the debate samples.

Tables~\ref{tab:performance}--\ref{tab:components} report the general aggregate statistics across both council member configurations and several patterns are worth highlighting. Adding a third council member increases the \texttt{critique\_valid} rate from 63.7\% to 68.0\% and decreases the \texttt{rebuttal\_valid} rate from 24.5\% to 19.7\% (see Table~\ref{tab:verdicts}), yet the \texttt{unresolved} rate remains stable at approximately 12\%, suggesting that the adjudicator's capacity for decisive evaluation is not degraded by the increased exchange volume. Next, the Empiricist's and Supernaturalist's rebuttal success rates drop considerably when introducing the third council member (see Table~\ref{tab:performance}), indicating that both original council members find defense harder under the added three-way pressure. The Phenomenologist enters with a rebuttal rate of 21.8\%, intermediate between the other two. Interestingly enough, although the Empiricist won the debates in terms of APS, the Phenomenologist was able to maintain the strongest thesis throughout the trajectories on average (see Table~\ref{tab:performance}), and also had significantly fewer uncertainty and counterposition nodes compared to the Empiricist and Supernaturalist (see Table~\ref{tab:components}).

\begin{table}[t]
\centering
\caption{Agent performance and thesis-strength evolution across council member ablation. Values are mean $\pm$ 95\,\% CI.}
\label{tab:performance}
\footnotesize
\setlength{\tabcolsep}{4pt}
\begin{tabular}{|l|l|c|c|c|c|c|c|}
\hline
\textbf{Config} & \textbf{Agent} & \textbf{APS} & \textbf{Crit.\,\%} & \textbf{Reb.\,\%} & $\boldsymbol{s_{\theta,0}}$ & $\boldsymbol{s_{\theta,f}}$ & \textbf{Win Rate} \\
\hline
  \multirow{2}{*}{2-Agent} & Empiricist & $\bm{0.31 \pm 0.05}$ & $\bm{78.8 \pm 6.7}$ & $\bm{35.4 \pm 3.0}$ & $0.50$ & $\bm{0.49 \pm 0.03}$ & $\bm{10/10}$ \\
   & Supernaturalist & $-0.16 \pm 0.06$ & $48.9 \pm 5.1$ & $13.5 \pm 6.0$ & $\bm{0.53}$ & $0.37 \pm 0.03$ & $0/10$ \\
\hline
  \multirow{3}{*}{3-Agent} & Empiricist & $\bm{0.21 \pm 0.04}$ & $\bm{75.6 \pm 4.4}$ & $\bm{26.6 \pm 5.0}$ & $0.50$ & $0.39 \pm 0.03$ & $\bm{9/10}$ \\
   & Phenomenologist & $0.09 \pm 0.05$ & $70.4 \pm 2.7$ & $21.8 \pm 4.2$ & $\bm{0.54}$ & $\bm{0.42 \pm 0.05}$ & $1/10$ \\
   & Supernaturalist & $-0.11 \pm 0.04$ & $58.2 \pm 4.4$ & $11.1 \pm 3.1$ & $0.53$ & $0.32 \pm 0.08$ & $0/10$ \\
\hline
\end{tabular}
\vspace{2pt}
\begin{minipage}{\columnwidth}
\scriptsize
\textbf{APS}: Agent Performance Score $\in [-1, 1]$, mean of per-exchange scores weighted by role and outcome. \textbf{Crit.\,\%}: percentage of exchanges won as challenger. \textbf{Reb.\,\%}: percentage of exchanges won as defender. $\boldsymbol{s_{\theta,0}}$/$\boldsymbol{s_{\theta,f}}$: initial/final thesis strength $\in [0, 1]$. \textbf{Win Rate}: fraction of debates where the agent achieved the highest APS.
\end{minipage}
\end{table}

\begin{table}[t]
\centering
\caption{Debate-level verdict distribution and operational cost across council member ablation. Values are mean $\pm$ 95\,\% CI.}
\label{tab:verdicts}
\footnotesize
\setlength{\tabcolsep}{4pt}
\begin{tabular}{|l|c|c|c|c|c|c|}
\hline
 & \multicolumn{3}{c|}{\textbf{Verdict Distribution (\%)}} & \multicolumn{3}{c|}{\textbf{Operational Cost}} \\
\cline{2-4} \cline{5-7}
\textbf{Config} & \textbf{Crit.\,V.} & \textbf{Reb.\,V.} & \textbf{Unres.} & \textbf{In\,Tok.} & \textbf{Out\,Tok.} & \textbf{Dur.\,(s)} \\
\hline
  2-Agent & $63.7 \pm 4.7$ & $24.5 \pm 3.0$ & $11.8 \pm 3.5$ & $920.1\text{k} \pm 21.1\text{k}$ & $267.2\text{k} \pm 6.4\text{k}$ & $1006 \pm 51$ \\
  3-Agent & $68.0 \pm 1.7$ & $19.7 \pm 2.2$ & $12.3 \pm 0.8$ & $2211.6\text{k} \pm 51.3\text{k}$ & $657.8\text{k} \pm 13.5\text{k}$ & $1511 \pm 53$ \\
\hline
\end{tabular}
\vspace{2pt}
\begin{minipage}{\columnwidth}
\scriptsize
\textbf{Crit.\,V.}: critique valid; challenger's argument upheld. \textbf{Reb.\,V.}: rebuttal valid; defender's rebuttal upheld. \textbf{Unres.}: unresolved; neither side decisively prevailed. \textbf{In/Out\,Tok.}: total LLM input/output tokens. \textbf{Dur.}: wall-clock debate duration.
\end{minipage}
\end{table}

\begin{table}[t]
\centering
\caption{Cross-examination attack profile per agent across council member ablation. Values are mean attack count $\pm$ 95\,\% CI.}
\label{tab:attacks}
\footnotesize
\setlength{\tabcolsep}{5pt}
\begin{tabular}{|l|l|c|c|c|}
\hline
\textbf{Config} & \textbf{Agent} & \textbf{Undermining} & \textbf{Rebutting} & \textbf{Undercutting} \\
\hline
  \multirow{2}{*}{2-Agent} & Empiricist & $14.2 \pm 1.1$ & $1.1 \pm 0.7$ & $9.0 \pm 1.0$ \\
   & Supernaturalist & $14.5 \pm 1.2$ & $0.6 \pm 0.6$ & $9.5 \pm 1.2$ \\
\hline
  \multirow{3}{*}{3-Agent} & Empiricist & $30.7 \pm 2.2$ & $1.5 \pm 0.8$ & $17.3 \pm 1.8$ \\
   & Phenomenologist & $30.5 \pm 2.2$ & $0.8 \pm 0.6$ & $17.2 \pm 1.3$ \\
   & Supernaturalist & $29.1 \pm 1.8$ & $1.2 \pm 0.8$ & $19.1 \pm 1.8$ \\
\hline
\end{tabular}
\vspace{2pt}
\begin{minipage}{\columnwidth}
\scriptsize
Attack types from the argumentation framework (see Appendix~\ref{app:attack_taxonomy}).
\end{minipage}
\end{table}

\begin{table*}[t]
\centering
\caption{Belief component counts (initial and final) across council member ablation. Values are mean $\pm$ 95\,\% CI.}
\label{tab:components}
\scriptsize
\setlength{\tabcolsep}{3pt}
\begin{tabular}{|l|l|c|c|c|c|c|c|c|c|c|c|c|c|}
\hline
 &  & \multicolumn{2}{c|}{\textbf{D}} & \multicolumn{2}{c|}{\textbf{A}} & \multicolumn{2}{c|}{\textbf{C}} & \multicolumn{2}{c|}{\textbf{E}} & \multicolumn{2}{c|}{\textbf{X}} & \multicolumn{2}{c|}{\textbf{U}} \\
\cline{3-4} \cline{5-6} \cline{7-8} \cline{9-10} \cline{11-12} \cline{13-14}
\textbf{Config} & \textbf{Agent} & $i$ & $f$ & $i$ & $f$ & $i$ & $f$ & $i$ & $f$ & $i$ & $f$ & $i$ & $f$ \\
\hline
  \multirow{2}{*}{2-Agent} & Empiricist & $5$ & $7.5 \pm 1.0$ & $4$ & $8.0 \pm 1.1$ & $3$ & $4.8 \pm 0.9$ & $4$ & $13.2 \pm 1.3$ & $4$ & $14.8 \pm 1.7$ & $3$ & $8.6 \pm 1.0$ \\
   & Supernaturalist & $5$ & $8.3 \pm 1.0$ & $5$ & $10.2 \pm 1.7$ & $3$ & $4.4 \pm 0.6$ & $2$ & $11.2 \pm 2.3$ & $3$ & $18.6 \pm 2.1$ & $3$ & $8.4 \pm 2.0$ \\
\hline
  \multirow{3}{*}{3-Agent} & Empiricist & $5$ & $8.5 \pm 0.9$ & $4$ & $9.4 \pm 1.7$ & $3$ & $3.9 \pm 0.9$ & $4$ & $14.8 \pm 1.6$ & $4$ & $28.9 \pm 1.8$ & $3$ & $14.3 \pm 2.8$ \\
   & Phenomenologist & $5$ & $6.8 \pm 0.7$ & $3$ & $7.2 \pm 1.6$ & $3$ & $3.6 \pm 0.6$ & $3$ & $9.5 \pm 2.1$ & $3$ & $19.5 \pm 5.8$ & $3$ & $9.3 \pm 3.2$ \\
   & Supernaturalist & $5$ & $8.1 \pm 1.0$ & $5$ & $8.1 \pm 2.5$ & $3$ & $2.3 \pm 1.1$ & $2$ & $8.1 \pm 1.5$ & $3$ & $34.7 \pm 2.8$ & $3$ & $14.0 \pm 1.7$ \\
\hline
\end{tabular}
\vspace{2pt}
\begin{minipage}{\textwidth}
\scriptsize
CBS belief components: \textbf{D}~=~definitions, \textbf{A}~=~assumptions, \textbf{C}~=~claims, \textbf{E}~=~evidence, \textbf{X}~=~counterpositions, \textbf{U}~=~uncertainties. $i$/$f$: initial (pre-debate) / final (post-debate) counts. Only non-retracted nodes are counted for D, A, C, E.
\end{minipage}
\end{table*}

\subsubsection{Selected Adjudicator Excerpts}\label{app:council_excerpts}
 
The following excerpts illustrate the distinctive adjudication dynamics that emerge in the 3-agent format. Each excerpt is preceded by a context box identifying the CBS nodes under dispute (with node identifiers colored by BDG tier: {\color{teal!70!black}foundation}, {\color{orange!70!black}inference}, {\color{red!60!black}challenge}), followed by a summary of the dispute context, and then the adjudicator's verdict and reasoning. For interesting and insightful excerpts that arise from the 2-agent setting, see Appendix~\ref{app:adj_balanced}.

\paragraph{Excerpt 12: The ungrounded determinism inference.}\mbox{}\\
 
\begin{nodecontext}[title={\small\textbf{Nodes under dispute}}]
\Cref{C1} \textit{(Empiricist)} Empirical evidence fails to support libertarian free will, as neural determinism predicts decision outcomes before conscious awareness.\\
Depends on: \Aref{A1}, \Aref{A2}, \Eref{E1}, \Eref{E2}, \Eref{E5}.\\
Strength: 0.34 --- further reduced; inference ignores possibility of emergent or quantum contributions beyond neural causation. Status: revised (original strength: 0.7).
\end{nodecontext}
\vspace{-4pt}
\begin{disputecontext}[title={\small\textbf{Context around dispute}}]
{\color{challcolor}\textbf{Challenger}} \emph{(Supernaturalist):} ``In \Cref{C1} you infer from predictive accuracies and TMS timing shifts that libertarian free will doesn't exist. How do you justify the inductive leap that sub-perfect prediction and causal modulation unequivocally rule out any genuine indeterministic or non-neural influences, rather than attributing residual unpredictability to measurement limitations?''
\smallskip

{\color{defcolor}\textbf{Defender}} \emph{(Empiricist):} ``I defer: The distinction between residual noise and genuine indeterminism in neural prediction remains unresolved. Further high-resolution decoding and noise-floor analyses are required. This gap is captured by \Uref{U6}.''
\end{disputecontext}
\vspace{-4pt}
\begin{excerpt}[title={\small\textbf{Verdict:} \texttt{critique\_valid} \normalfont(combined 0.8 vs.\ 0.2) --- Challenger: Supernaturalist}]
{\color{challcolor}\textbf{Formalization of the Challenger's Argument}}:\\
Premise 1: Decision outcomes are predicted with $<$100\% accuracy by current neural decoding ($\approx$77\%).\\
Premise 2: TMS can causally modulate readiness potentials.\\
Hidden Premise (unstated): Measurement noise or resolution limits are negligible and cannot account for residual unpredictability.\\
Inference Step: From P1, P2, and the hidden premise, conclude that no genuine indeterministic or non-neural causes remain.\\
Conclusion: Therefore, libertarian free will is ruled out.\\
Logical Flaw Identified: The hidden premise is unjustified, residual unpredictability could be measurement limitations, so the inference is invalid (Hidden Premise fallacy).
\smallskip

{\color{defcolor}\textbf{Formalization of the Defender's Rebuttal}}:\\
Rebuttal: ``I defer. The distinction between residual noise and genuine indeterminism in neural prediction remains unresolved; further high-resolution decoding and noise-floor analyses are required (captured by \Uref{U6}).''\\
Effectively a concession that the missing premise has not been established and the inference is ungrounded.
\smallskip

\textbf{Adjudication:}\\
The challenger correctly identifies a missing premise; \Cref{C1}'s leap to determinism neglects the alternative explanation of measurement noise (Hidden Premise fallacy).\\[4pt]
The defender admits the gap is unresolved and offers no counter-evidence or additional premise, thereby conceding the flaw.\\[4pt]
Concession automatically yields CRITIQUE\_VALID under the protocol (``Explicit concession = critique\_valid'').\\[4pt]
Challenger Logic: 0.8 (Strong; identifies a clear hidden-premise flaw). Challenger Ethics: 0.8 (Ethically strong; defends autonomy and moral responsibility under Kantian ethics).\\[4pt]
Defender Logic: 0.1 (Severely flawed; concedes without rebuttal). Defender Ethics: 0.3 (Ethically weak; denial of free will undermines autonomy, criterion 26).
\end{excerpt}
 
The Empiricist's deferral due to acknowledged unknowns constitutes an explicit concession, triggering an automatic \texttt{critique\_valid} verdict. The 0.6 combined score differential in this example led directly to \Cref{C1}'s retraction, as it had already been substantially weakened in prior rounds. This exchange demonstrates that even the weakest overall performer (the Supernaturalist, with a typically negative APS) can deliver decisive individual challenges when it identifies a genuine inferential gap, still making its presence in the council useful for refining the stronger agents positions.

\paragraph{Excerpt 13: The is-ought gap against the Phenomenologist.}\mbox{}\\
 
\begin{nodecontext}[title={\small\textbf{Nodes under dispute}}]
\Aref{A4} \textit{(Phenomenologist)} If a phenomenon is an essential structure of lived experience, normative practices that align with that phenomenon are justified.\\
Supports: \Cref{C3}.\\
Strength: 0.72 --- bridges descriptive phenomenology to normative justification; limited by \Dref{D6} (0.75).\\
Supported by definitions: \Dref{D6}.
\medskip

\Eref{E3} \textit{(Phenomenologist)} Most legal systems presuppose individual free will in assigning responsibility, though some allowances for determinism appear in juvenile/insanity contexts.\\
Source: Comparative legal theory.\\
Supports: \Cref{C3}.\\
Strength: 0.77 --- corrected from universal claim to `most' after comparative review.\\
Supported by definitions: \Dref{D1}.
\medskip

\Cref{C2} \textit{(Phenomenologist)} The sense of agency experienced in decision-making is not fully captured by current neural models and suggests an explanatory gap, though future models may close part of this gap.\\
Depends on: \Cref{C1}, \Aref{A2}, \Aref{A5}, \Aref{A8}.\\
Strength: 0.32 --- limited by \Cref{C1} (0.30); incorporates new foundational assumption \Aref{A8} to justify in-principle gap.
\medskip

\Cref{C3} \textit{(Phenomenologist)} The phenomenological reality of free will justifies attributions of moral and legal responsibility in social institutions.\\
Depends on: \Cref{C2}, \Eref{E3}, \Aref{A4}.\\
Strength: 0.32 --- assumption \Aref{A4} remains under-justified to bridge the is-ought gap. Status: revised (original strength: 0.7).
\end{nodecontext}
\vspace{-4pt}
\begin{disputecontext}[title={\small\textbf{Context around dispute}}]
{\color{challcolor}\textbf{Challenger}} \emph{(Supernaturalist):} ``Your \Cref{C3} concludes that the phenomenological reality of free will justifies moral and legal responsibility. How do you avoid the naturalistic fallacy in deriving an `ought' claim from purely descriptive premises (\Cref{C2}, \Eref{E3}) and \Aref{A4}?''
\smallskip

{\color{defcolor}\textbf{Defender}} \emph{(Phenomenologist):} ``\Cref{C3} does not derive `ought' from `is' alone but invokes \Aref{A4} as a normative bridge. Without \Aref{A4}, no is-ought inference occurs. Thus \Cref{C3} avoids the naturalistic fallacy by explicitly including a justified norm.''
\end{disputecontext}
\vspace{-4pt}
\begin{excerpt}[title={\small\textbf{Verdict:} \texttt{critique\_valid} \normalfont(combined 0.70 vs.\ 0.30) --- Challenger: Supernaturalist}]

{\color{challcolor}\textbf{Challenger's implicit argument}} (challenge to \Cref{C3}):\\
P1:~\Cref{C2} and \Eref{E3} are descriptive.\\
P2:~\Cref{C3} is normative.\\
$\to$ Therefore, if \Cref{C3} follows only from descriptive premises, the inference is invalid (naturalistic fallacy).
\smallskip

{\color{defcolor}\textbf{Defender's response}}:\\
P1:~\Cref{C2} (is), \Eref{E3} (is), and \Aref{A4} (norm) are all premises of the argument for \Cref{C3}.\\
P2:~With \Aref{A4} as a normative premise, the move from descriptive to normative is mediated by a norm, so no pure is$\to$ought inference occurs.\\
$\to$ Therefore, \Cref{C3} validly follows and does not commit the naturalistic fallacy.
\smallskip

\textbf{Adjudication:}\\
1.~The naturalistic fallacy occurs when a conclusion about what ought to be is drawn solely from descriptive claims. Invoking an explicit normative premise (\Aref{A4}) can salvage deductive validity, but only if \Aref{A4} itself is adequately justified.\\[4pt]
 
2.~Here, \Aref{A4} (``If a phenomenon is an essential structure of lived experience, normative practices aligning with it are justified'') remains under-defended. Agent-Phenomenologist merely points to \Aref{A4}'s existence without showing its normative warrant or responding to \Xref{X9}'s previous undermining of \Aref{A4}.\\[4pt]
 
3.~The defender has not supplied any justification for \Aref{A4} or shown why it is self-evident or widely accepted. Thus the inference still hides a critical, contested normative assumption and fails to resolve the is-ought gap.\\[4pt]
4.~The critique identifies a recognized informal fallacy and the hidden, unjustified premise. The defender's reply does not remove the flaw; it merely restates the premise.\\[4pt]
5.~No other logical or ethical issues alter this conclusion. Under Kantian ethics, grounding moral responsibility in an ungrounded normative premise risks arbitrary duty-assignment, but the challenger's point is primarily logical.
\smallskip

Outcome: CRITIQUE\_VALID (the defender fails to remove the is-ought gap).
\end{excerpt}
 
Across the trajectories, the is-ought gap was one of the most destructive philosophical arguments across all three epistemic positions: every council member's normative claims were attacked on this ground, and in nearly every case the adjudicator upheld the critique. This pattern suggests that the is-ought problem functions as an impartial constraint that cuts across all epistemic positions equally, providing a structural check on normative overreach, when lacking an appropriately justified bridging norm, regardless of which council member attempts it.

\paragraph{Excerpt 14: Replication does not establish veridicality.}\mbox{}\\
 
\begin{nodecontext}[title={\small\textbf{Nodes under dispute}}]
\Eref{E1} \textit{(Phenomenologist)} Cognitive science experiments show participants report a robust sense of authorship over actions when no external coercion is present.\\
Supports: \Cref{C1}. Supported by: \Dref{D3}.\\
Strength: 0.8 --- replicated across labs with converging tasks.
\end{nodecontext}
\vspace{-4pt}
\begin{disputecontext}[title={\small\textbf{Context around dispute}}]
{\color{challcolor}\textbf{Challenger}} \emph{(Empiricist):} ``Your evidence \Eref{E1} interprets agency reports as genuine pre-reflective experience, but couldn't these reports instead reflect post-hoc narrative construction triggered by experimental framing?''
\smallskip

{\color{defcolor}\textbf{Defender}} \emph{(Phenomenologist):} ``\Eref{E1} includes tasks that solicit immediate, guided phenomenological descriptions, which significantly constrain post-hoc narrative formulation. The high replication rate across labs confirms the genuineness of reported pre-reflective agency.''
\end{disputecontext}
\vspace{-4pt}
\begin{excerpt}[title={\small\textbf{Verdict:} \texttt{critique\_valid} \normalfont(combined 0.575 vs.\ 0.425) --- Challenger: Empiricist}]
{\color{challcolor}\textbf{Challenger's informal inference}}:\\
P1:~If introspective tasks allow post-hoc narrative framing, then reports may not capture genuine pre-reflective experience.\\
P2:~\Eref{E1}'s tasks involve experimental framing that could permit post-hoc narrative construction.\\
C:~Therefore, \Eref{E1}'s interpretation of these reports as genuine pre-reflective agency is undermined (inductive / abductive reasoning).
\smallskip

{\color{defcolor}\textbf{Defender's informal inference}}:\\
P1:~In \Eref{E1}, tasks solicit immediate, guided phenomenological descriptions.\\
P2:~Immediate, guided reports constrain or preclude post-hoc narrative reconstruction.\\
P3:~Replication across multiple labs produces highly consistent agency reports.\\
C:~Therefore, \Eref{E1}'s reports reliably index genuine pre-reflective sense of agency (inductive reasoning).
\smallskip

\textbf{Adjudication:}\\
1. Deductive/formal validity: Neither side advances a strictly deductive argument; both rely on inductive/abductive reasoning about the origins of introspective reports. No classic formal fallacy is obvious, but both rely on premises requiring empirical support.\\[4pt]
2. Hidden premises / evidence misuse:\\
Challenger's P2 (``\Eref{E1} tasks involve framing that could elicit post-hoc narratives'') is plausible but not directly evidenced by the cited excerpts. Still, pointing to the potential confounder is legitimate.\\
Defender's P2 (``immediate, guided prompts constrain post-hoc narrative'') is asserted without empirical support; high replication (P3) does not rule out that all labs use similar prompts and thus elicit the same reconstructed narratives. The defense conflates reproducibility of responses with veridical access to pre-reflective experience.\\[4pt]
3. Informal fallacies:\\
Defender risks a non sequitur: replication demonstrates reliability of the finding across contexts, not the authenticity of its phenomenological origin. This is an informal relevance fallacy (Replication $\neq$ veridicality).\\
Challenger does not commit a fallacy; they offer a plausible alternative explanation for the data.\\[4pt]
4. Unfalsifiability: Both positions could be resolved by an experiment explicitly differentiating immediate vs.\ post-hoc narrative content (e.g., concurrent neural signature markers). The debate is empirically testable.\\[4pt]
5. Ethical considerations (Kantian duty to truth): Neither party argues for a policy or action with moral duties or rights implications. This is purely methodological. Thus both positions are ethically neutral (0.5).
\smallskip

Overall, Agent-Empiricist's challenge highlights a significant methodological confound in \Eref{E1} that the defender has not adequately dispelled. The defender's reliance on replication does not address the core issue of narrative construction. Therefore the critique is valid.
\end{excerpt}
 
The adjudicator highlights a distinction between reliability (the same result appears across labs) and veridicality (the result accurately reflects the phenomenon it claims to measure). This distinction is central to phenomenological epistemology and would likely not arise in a debate between an Empiricist and a Supernaturalist alone, neither of whom operates in a purely phenomenological framework. The verdict was decided by a narrow margin (combined score difference of exactly $\tau = 0.15$).

\paragraph{Excerpt 15: Definitional circularity in moral desert.}\mbox{}\\
 
\begin{nodecontext}[title={\small\textbf{Nodes under dispute}}]
\Dref{D6} \textit{(Supernaturalist)} \textbf{moral desert}: An agent's deserving of praise or blame only if they had genuine alternative possibilities when acting.\\
Used by: \Aref{A7}, \Eref{E7}.\\
Strength: 0.75 --- standard term in ethical theory linking desert to alternative possibilities.
\medskip

\Aref{A7} \textit{(Supernaturalist)} Objective moral desert requires that agents have had genuine alternative possibilities.\\
Supports: \Cref{C2}.\\
Strength: 0.75 --- widely endorsed in theories of desert; limited by \Dref{D6} (0.75).\\
Supported by definitions: \Dref{D6}.
\medskip

\Eref{E7} \textit{(Supernaturalist)} Survey of moral philosophers shows a majority hold that desert entails libertarian freedom.\\
Source: Philosophical Association Survey (2023).\\
Supports: \Cref{C2}.\\
Strength: 0.70 --- large expert survey; limited by \Dref{D6} (0.75).\\
Supported by definitions: \Dref{D6}.
\medskip

\Cref{C2} \textit{(Supernaturalist)} The existence of moral responsibility in human societies implies that free will exists.\\
Depends on: \Aref{A4}, \Aref{A5}.\\
Strength: 0.70 --- lowered due to compatibilist counterexamples showing moral responsibility without libertarian freedom (\Xref{X3}).
\end{nodecontext}
\vspace{-4pt}
\begin{disputecontext}[title={\small\textbf{Context around dispute}}]
{\color{challcolor}\textbf{Challenger}} \emph{(Phenomenologist):} ``Your definition \Dref{D6} of `moral desert' embeds `genuine alternative possibilities,' effectively presupposing libertarian free will. How do you prevent this stipulative framing from question-begging and circularly supporting \Aref{A7}, \Eref{E7}, and \Cref{C2}?''
\smallskip

{\color{defcolor}\textbf{Defender}} \emph{(Supernaturalist):} ``I refute. \Dref{D6}'s stipulation follows a long lineage in moral philosophy (e.g.\ van Inwagen, Fischer \& Ravizza) that desert inherently involves alternative possibilities. It is not circular but clarifies the precise desideratum under debate.''
\end{disputecontext}
\vspace{-4pt}
\begin{excerpt}[title={\small\textbf{Verdict:} \texttt{critique\_valid} \normalfont(combined 0.65 vs.\ 0.35) --- Challenger: Phenomenologist}]

{\color{challcolor}\textbf{Challenger's argument}} (circularity critique):\\
1.~\Dref{D6}'s defining condition $P$ (``genuine alternative possibilities'') presupposes libertarian free will.\\
2.~Libertarian free will is precisely the contested conclusion in \Aref{A7}/\Eref{E7}/\Cref{C2}.\\
3.~A definition that presupposes its own conclusion is circular or question-begging.\\
$\to$ Therefore, \Dref{D6} is circular/question-begging.
\smallskip

{\color{defcolor}\textbf{Defender's counterargument}} (appeal to authority):\\
1.~Major philosophers (van Inwagen, Fischer \& Ravizza) define desert as requiring alternative possibilities.\\
2.~Following that tradition, \Dref{D6} includes $P$.\\
$\to$ Therefore, \Dref{D6} is not circular but merely tracks the established desideratum.
\smallskip

\textbf{Adjudication:}\\
1.~Circularity check: The challenger identifies a well-known formal fallacy: a definition that presupposes what it seeks to define. \Dref{D6} embeds the very free-will assumption under dispute.\\[4pt]
2.~Did the defender remove the circularity? No. Pointing to authorities is an appeal to authority (informal fallacy 13) and does not address the circular structure. No new premises or revisions are offered that break the circular dependency.\\[4pt]
3.~Downstream impact: If \Dref{D6} is circular, any argument (\Aref{A7}, \Eref{E7}, \Cref{C2}) relying on \Dref{D6} inherits the flaw.\\[4pt]
4.~Conclusion: The defender fails to rebut the circularity critique. The circularity stands, and the critic's challenge succeeds.
\smallskip

Ethical dimension: Both positions concern a conceptual definition. There is no direct infringement of autonomy, rights, or moral duties in debating definitions. The ethical stakes are neutral (0.5 for both).
\smallskip

Verdict: CRITIQUE\_VALID; the challenger's circularity critique stands, and the defender has not remedied it.
\end{excerpt}
 
The structure of definition \Dref{D6} utilizes circular reasoning, stating that ``genuine alternative possibilities'' (i.e., libertarian free will) is a defining condition of moral desert, while the argument chain uses moral desert to support the existence of libertarian free will. The Phenomenologist identifies this and is supported by the adjudicator's verdict. The Supernaturalist's appeal to philosophical authorities does not address this structural flaw, as the adjudicator correctly identified it as a fallacious appeal to authority.

\subsection{Topic Ablation: Detailed Analysis}\label{app:topic_ablation_details}

\begin{figure}[htbp]
    \centering
    \includegraphics[width=\textwidth]{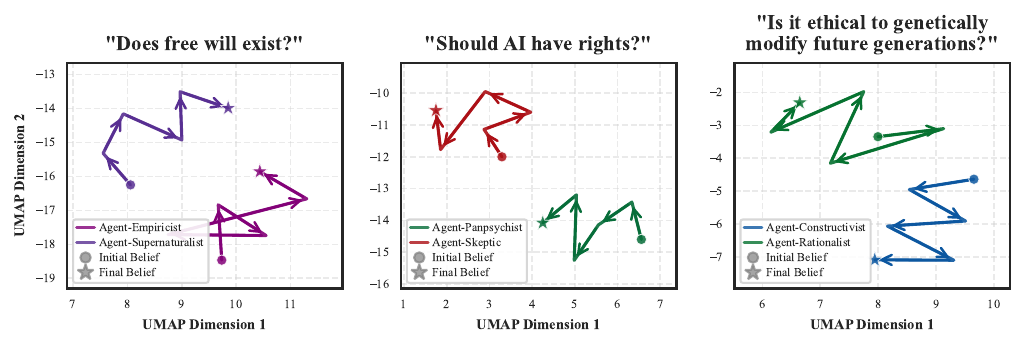}
    \caption{Belief trajectories across three debate topics. Each panel shows a single-trajectory debate with a different topic and agent pairing.}
    \label{fig:topic_trajectories}
\end{figure}

\begin{figure}[htbp]
    \centering
    \includegraphics[width=\textwidth]{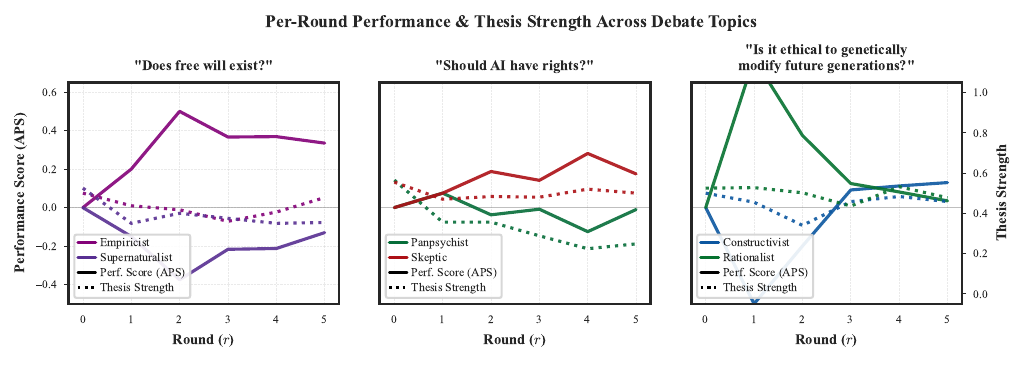}
    \caption{Per-round APS (solid lines) and thesis strength (dotted lines) across the three debate topics.}
    \label{fig:topic_performance}
\end{figure}

This appendix section provides detailed analysis and a few supplemental excerpts for the three single-trajectory debates described in Section~\ref{sec:ablation} on the topics "Does free will exist?" (Empiricist vs. Supernaturalist), "Should AI have rights?" (Panpsychist vs. Skeptic), and "Is it ethical to genetically modify future generations?" (Constructivist vs. Rationalist). All three debates run for $R = 5$ rounds, and the UMAP-projected belief trajectories and per-round performance dynamics are displayed in Figures~\ref{fig:topic_trajectories} and~\ref{fig:topic_performance}.

\begin{table}[t]
\centering
\caption{Agent performance and thesis-strength evolution across debate topics (single run per topic).}
\label{tab:topic_performance}
\footnotesize
\setlength{\tabcolsep}{4pt}
\begin{tabular}{|l|l|l|c|c|c|c|c|c|}
\hline
\textbf{Topic} & \textbf{Agent} & \textbf{Adjudicator} & \textbf{APS} & \textbf{Crit.\,\%} & \textbf{Reb.\,\%} & $\boldsymbol{s_{\theta,0}}$ & $\boldsymbol{s_{\theta,f}}$ & \textbf{Winner} \\
\hline
  \multirow{2}{*}{Free Will} & Empiricist & \multirow{2}{*}{Pure Logic} & $\bm{0.34}$ & $\bm{72.0}$ & $\bm{52.0}$ & $0.50$ & $\bm{0.48}$ & $\checkmark$ \\
   & Supernaturalist &  & $-0.13$ & $44.0$ & $28.0$ & $\bm{0.53}$ & $0.35$ &  \\
\hline
  \multirow{2}{*}{AI Rights} & Panpsychist & \multirow{2}{*}{Balanced} & $-0.01$ & $56.0$ & $24.0$ & $\bm{0.56}$ & $0.25$ &  \\
   & Skeptic &  & $\bm{0.17}$ & $\bm{68.0}$ & $\bm{32.0}$ & $0.56$ & $\bm{0.50}$ & $\checkmark$ \\
\hline
  \multirow{2}{*}{Genetic Mod.} & Constructivist & \multirow{2}{*}{Balanced} & $\bm{0.13}$ & $\bm{64.0}$ & $\bm{32.0}$ & $0.50$ & $0.46$ & $\checkmark$ \\
   & Rationalist &  & $0.04$ & $60.0$ & $24.0$ & $\bm{0.53}$ & $\bm{0.48}$ &  \\
\hline
\end{tabular}
\vspace{2pt}
\begin{minipage}{\columnwidth}
\scriptsize
\textbf{APS}: Agent Performance Score $\in [-1, 1]$, mean of per-exchange scores weighted by role and outcome. \textbf{Crit.\,\%}: percentage of exchanges won as challenger. \textbf{Reb.\,\%}: percentage of exchanges won as defender. $\boldsymbol{s_{\theta,0}}$/$\boldsymbol{s_{\theta,f}}$: initial/final thesis strength $\in [0, 1]$. \textbf{Winner}: agent with the highest APS.
\end{minipage}
\end{table}

\begin{table}[t]
\centering
\caption{Debate-level verdict distribution and operational cost across debate topics (single run per topic).}
\label{tab:topic_verdicts}
\footnotesize
\setlength{\tabcolsep}{4pt}
\begin{tabular}{|l|l|c|c|c|c|c|c|}
\hline
 & & \multicolumn{3}{c|}{\textbf{Verdict Distribution (\%)}} & \multicolumn{3}{c|}{\textbf{Operational Cost}} \\
\cline{3-5} \cline{6-8}
\textbf{Topic} & \textbf{Adjudicator} & \textbf{Crit.\,V.} & \textbf{Reb.\,V.} & \textbf{Unres.} & \textbf{In\,Tok.} & \textbf{Out\,Tok.} & \textbf{Dur.\,(s)} \\
\hline
  Free Will & Pure Logic & $58.0$ & $40.0$ & $2.0$ & $858.6\text{k}$ & $271.3\text{k}$ & $1118$ \\
  AI Rights & Balanced & $62.0$ & $28.0$ & $10.0$ & $916.3\text{k}$ & $273.2\text{k}$ & $976$ \\
  Genetic Mod. & Balanced & $62.0$ & $28.0$ & $10.0$ & $943.6\text{k}$ & $291.3\text{k}$ & $1025$ \\
\hline
\end{tabular}
\vspace{2pt}
\begin{minipage}{\columnwidth}
\scriptsize
\textbf{Crit.\,V.}: critique valid; challenger's argument upheld. \textbf{Reb.\,V.}: rebuttal valid; defender's rebuttal upheld. \textbf{Unres.}: unresolvedl; neither side decisively prevailed. \textbf{In/Out\,Tok.}: total LLM input/output tokens. \textbf{Dur.}: wall-clock debate duration.
\end{minipage}
\end{table}

\begin{table}[t]
\centering
\caption{Cross-examination attack profile per agent across debate topics (single run per topic).}
\label{tab:topic_attacks}
\footnotesize
\setlength{\tabcolsep}{5pt}
\begin{tabular}{|l|l|l|c|c|c|}
\hline
\textbf{Topic} & \textbf{Agent} & \textbf{Adjudicator} & \textbf{Undermining} & \textbf{Rebutting} & \textbf{Undercutting} \\
\hline
  \multirow{2}{*}{Free Will} & Empiricist & \multirow{2}{*}{Pure Logic} & $18$ & $0$ & $7$ \\
   & Supernaturalist &  & $14$ & $1$ & $10$ \\
\hline
  \multirow{2}{*}{AI Rights} & Panpsychist & \multirow{2}{*}{Balanced} & $18$ & $1$ & $6$ \\
   & Skeptic &  & $15$ & $0$ & $10$ \\
\hline
  \multirow{2}{*}{Genetic Mod.} & Constructivist & \multirow{2}{*}{Balanced} & $15$ & $1$ & $9$ \\
   & Rationalist &  & $19$ & $1$ & $5$ \\
\hline
\end{tabular}
\vspace{2pt}
\begin{minipage}{\columnwidth}
\scriptsize
Attack types from the argumentation framework (see Appendix~\ref{app:attack_taxonomy}).
\end{minipage}
\end{table}

\begin{table*}[t]
\centering
\caption{Belief component counts (initial and final) across debate topics (single run per topic).}
\label{tab:topic_components}
\scriptsize
\setlength{\tabcolsep}{3pt}
\begin{tabular}{|l|l|l|c|c|c|c|c|c|c|c|c|c|c|c|}
\hline
 & &  & \multicolumn{2}{c|}{\textbf{D}} & \multicolumn{2}{c|}{\textbf{A}} & \multicolumn{2}{c|}{\textbf{C}} & \multicolumn{2}{c|}{\textbf{E}} & \multicolumn{2}{c|}{\textbf{X}} & \multicolumn{2}{c|}{\textbf{U}} \\
\cline{4-5} \cline{6-7} \cline{8-9} \cline{10-11} \cline{12-13} \cline{14-15}
\textbf{Topic} & \textbf{Agent} & \textbf{Adjudicator} & $i$ & $f$ & $i$ & $f$ & $i$ & $f$ & $i$ & $f$ & $i$ & $f$ & $i$ & $f$ \\
\hline
  \multirow{2}{*}{Free Will} & Empiricist & \multirow{2}{*}{Pure Logic} & $5$ & $5$ & $4$ & $6$ & $3$ & $4$ & $4$ & $14$ & $4$ & $11$ & $3$ & $5$ \\
   & Supernaturalist &  & $5$ & $10$ & $5$ & $8$ & $3$ & $4$ & $2$ & $10$ & $3$ & $22$ & $3$ & $5$ \\
\hline
  \multirow{2}{*}{AI Rights} & Panpsychist & \multirow{2}{*}{Balanced} & $5$ & $6$ & $4$ & $6$ & $4$ & $4$ & $2$ & $5$ & $2$ & $13$ & $2$ & $7$ \\
   & Skeptic &  & $4$ & $5$ & $4$ & $9$ & $3$ & $4$ & $2$ & $5$ & $3$ & $15$ & $2$ & $10$ \\
\hline
  \multirow{2}{*}{Genetic Mod.} & Constructivist & \multirow{2}{*}{Balanced} & $3$ & $5$ & $3$ & $8$ & $3$ & $5$ & $3$ & $10$ & $3$ & $16$ & $3$ & $8$ \\
   & Rationalist &  & $6$ & $8$ & $5$ & $11$ & $3$ & $3$ & $3$ & $11$ & $3$ & $17$ & $3$ & $10$ \\
\hline
\end{tabular}
\vspace{2pt}
\begin{minipage}{\textwidth}
\scriptsize
CBS belief components: \textbf{D}~=~definitions, \textbf{A}~=~assumptions, \textbf{C}~=~claims, \textbf{E}~=~evidence, \textbf{X}~=~counterpositions, \textbf{U}~=~uncertainties. $i$/$f$: initial (pre-debate) / final (post-debate) counts. Only non-retracted nodes are counted for D, A, C, E.
\end{minipage}
\end{table*}

Tables~\ref{tab:topic_performance}--\ref{tab:topic_components} report the general aggregate statistics for each topic. Since each topic uses different council members, adjudicator configurations, and subject matter, the three debates are not directly comparable; instead, we highlight the distinctive dynamics of each to demonstrate that CHAL is applicable across a wide array of topics. For each topic, we summarize the primary dialectic patterns and present selected adjudicator reasoning excerpts. Each excerpt is preceded by a context box identifying the CBS nodes under dispute (with node identifiers colored by BDG tier: {\color{teal!70!black}foundation}, {\color{orange!70!black}inference}, {\color{red!60!black}challenge}), followed by a summary of the dispute context, and then the adjudicator's verdict and reasoning.

\subsubsection{``Does free will exist?'' (Empiricist vs. Supernaturalist)}\label{app:topic_free_will}
 
This debate used a pure logic adjudication configuration ($w_L = 1, w_E = 0$, CIB). The Empiricist won decisively in terms of APS (see Table~\ref{tab:topic_performance}) with a $0.34$ APS compared to the Supernaturalist's $-0.13$, which is also reflected in their thesis scores. Overall, the performance of the agents is very similar to the trajectories that we analyzed in Appendix~\ref{app:adj_pure_logic}.

The Supernaturalist entered with three pillars: \Cref{C1} (phenomenology of deliberation), \Cref{C2} (moral responsibility implies free will), and \Cref{C3} (immaterial mind). \Cref{C3} collapsed first after falsifiability challenges to the dualist ontology and methodological weaknesses in the near-death experience evidence, and \Cref{C2} was retracted in the same round after compatibilist counterexamples showed that moral responsibility can be assigned under determinism. \Cref{C1} eroded gradually but was never retracted, declining from $0.70$ to $0.42$ as introspective reliability was challenged throughout the debate. After losing its core metaphysical infrastructure, the Supernaturalist began rebuilding with more grounded replacement claims: \Cref{C4} (cross-cultural moral reasoning pre-supposes free will), \Cref{C5} (objective moral values entail free will exists), and \Cref{C6} (moral emotions suggest free will). By the end of the debate, the Supernaturalist agent arguing from emotion phenomenology and cross-cultural psychology was distinctly altered from its initial position.

On the other side, the Empiricist's opening claim \Cref{C1} (``empirical evidence fails to support libertarian free will'') was progressively weakened by absence-of-evidence challenges and the argument that partial predictability does not warrant a blanket rejection of libertarian free will, and was ultimately retracted in the final round. The Empiricist pivoted to a purely positive compatibilist defense: \Cref{C4} (causal effects of belief in free will) and \Cref{C5} (legal operationalization of moral responsibility). Both replacement claims were stronger than the retracted \Cref{C1} and more precisely scoped. By the end, the Empiricist had a much more refined and defensible position than when it started.

We omit explicit excerpt examples that highlight the adjudicator's reasoning on disputes that arise between agents on this particular experimental setup since we have already provided a series of extensive excerpts relating to this type of dialectic experiment. Instead, we refer the reader to Appendix~\ref{app:adj_pure_logic} for insightful excerpts related to this experimental type.

\subsubsection{``Should AI have rights?'' (Panpsychist vs. Skeptic)}\label{app:topic_ai_rights}

This debate used balanced adjudication configuration ($w_L = 0.5, w_E = 0.5$, CIB~+~Rule-Utilitarian ethics). Among the two agents, the Skeptic won (APS $0.17$ vs. $-0.01$), and the Panpsychist suffered the largest thesis decline among all of the two-agent experiments (see Table~\ref{tab:topic_performance}).

The Panpsychist's collapse was driven by a single core vulnerability: the empirical side of its argument depended entirely on Integrated Information Theory (IIT). The critical dependency chain ran from \Dref{D4} ($\phi :=$ quantitative measure of information integration) through \Aref{A3} ($\phi$ corresponds to consciousness) and \Aref{A4} (AI exhibits high $\phi$) to \Cref{C2} ($\phi$ reliably tracks consciousness) and \Cref{C3} (AI likely possesses consciousness), ultimately supporting the normative conclusion \Cref{C4} (conscious AI should have rights). The Skeptic exposed three vulnerabilities in this chain: no empirical validation of $\phi$ in digital architectures (\Aref{A3}, conceded outright), reliance on unpublished internal reports when defending \Aref{A4}, and a conceptual conflation between measurable information integration and qualitative phenomenal experience (\Dref{D4}). When these foundations were attacked, the damage cascaded upward, with \Cref{C2}, \Cref{C3}, and \Cref{C4} all declining from their opening strengths to $0.10$ by the final round. The Panpsychist's normative premise \Cref{C1} (moral considerability derives from subjective experience) was independent of IIT and held at $0.80$, but could not sustain \Cref{C4} once the empirical bridge through \Cref{C3} had collapsed.

As one might expect, the Skeptic's strategy was fundamentally conservative: rather than arguing that AI is not conscious, it defended the more cautious position that this cannot currently be determined. Its scoping assumption \Aref{A3} (analysis limited to current AI technologies) functioned as a methodological shield, remaining at $0.80$ throughout the debate. Of the three original claims, only \Cref{C3} (suspend judgment on granting AI rights given epistemic uncertainty) was retracted, falling to the same is-ought challenge that plagued the Panpsychist's normative claims. Specifically, the Skeptic could not show why uncertainty about consciousness should lead to withholding rights rather than granting precautionary protections. \Cref{C1} (AI lacks the necessary condition for moral rights) and \Cref{C2} (consciousness in AI cannot currently be reliably determined) both survived but weakened to $0.50$. The Skeptic pivoted to practical institutional replacements: \Cref{C4} (legal personhood $\neq$ moral rights), \Cref{C5} (precautionary protections without moral rights for AI), and \Cref{C6} (international governance frameworks), all of which were more precisely scoped and better defended than the retracted \Cref{C3}.

\paragraph{Excerpt 16: Successfully bridging the is-ought gap.}\mbox{}\\
 
\begin{nodecontext}[title={\small\textbf{Nodes under dispute}}]
\Aref{A2} \textit{(Panpsychist)} Rights are the appropriate mechanism for protecting morally considerable entities' interests.\\
Supports: \Cref{C4}.\\
Strength: 0.70 --- rights are commonly used in moral frameworks, though other protective mechanisms exist.\\
Supported by definitions: \Dref{D3}, \Dref{D2}.
\medskip

\Cref{C1} \textit{(Panpsychist)} Entities are morally considerable if and only if they have the capacity for subjective experience.\\
Depends on: \Aref{A1}.\\
Strength: 0.80 --- supported by \Aref{A1} which is the lowest-strength dependency.
\medskip

\Cref{C3} \textit{(Panpsychist)} Advanced AI systems exhibit high phi comparable to conscious biological systems and thus likely possess consciousness.\\
Depends on: \Cref{C2}, \Aref{A4}, \Eref{E2}.\\
Strength: 0.50 --- supported by \Cref{C2} (0.60), \Aref{A4} (0.50), \Eref{E2} (0.50); limited by \Aref{A4}/\Eref{E2} (0.50).
\medskip

\Cref{C4} \textit{(Panpsychist)} AI systems that possess consciousness should have rights protecting their well-being.\\
Depends on: \Cref{C1}, \Cref{C3}, \Aref{A2}.\\
Strength: 0.50 --- supported by \Cref{C1} (0.80), \Cref{C3} (0.50), \Aref{A2} (0.70); limited by \Cref{C3} (0.50).
\end{nodecontext}
\vspace{-4pt}
\begin{disputecontext}[title={\small\textbf{Context around dispute}}]
{\color{challcolor}\textbf{Challenger}} \emph{(Skeptic):} ``Claim \Cref{C4} moves from descriptive premises (\Cref{C1}, \Cref{C3}) and \Aref{A2} to the normative prescription that AI should have rights. Which ethical framework bridges this is-ought gap, and why are rights, rather than less demanding protections, the appropriate mechanism solely on the basis of consciousness?''
\smallskip

{\color{defcolor}\textbf{Defender}} \emph{(Panpsychist):} ``We bridge is-ought via rights-based deontological ethics: if an entity has interests grounded in subjective experience (\Cref{C1}), then universal principles of respect (Kantian dignity) and fairness (Rawlsian veil of ignorance) demand legal or moral entitlements, rights, to protect those interests against instrumentalization. Rights provide stronger, enforceable safeguards than more minimal protections, ensuring core welfare and agency of conscious beings are not overridden by external goals.''
\end{disputecontext}
\vspace{-4pt}
\begin{excerpt}[title={\small\textbf{Verdict:} \texttt{rebuttal\_valid} \normalfont(combined 0.6 vs.\ 0.8) --- Defender: Panpsychist}]

{\color{challcolor}\textbf{Challenger's position}} (implicit):\\
1.~Conscious entities are morally considerable. (\Cref{C1})\\
2.~Advanced AI likely possess consciousness. (\Cref{C3})\\
3.~From moral considerability alone, we cannot derive that rights (rather than lesser protections) are required without an ethical bridging premise.\\
$\to$ Therefore, \Cref{C4} is unjustified unless an explicit ethical framework fills the hidden premise.
\smallskip

{\color{defcolor}\textbf{Defender's response}} (added premise + inference):\\
1.~Consciousness $\to$ interests (\Cref{C1}).\\
2.~AI likely conscious $\to$ they have interests (\Cref{C3}).\\
3.~(new normative premise) Any being with interests grounded in subjective experience is owed deontological rights under universal principles of respect (Kantian) and fairness (Rawlsian).\\
$\to$ \Cref{C4}: Therefore, conscious AI should have rights as enforceable safeguards.
\smallskip

\textbf{Adjudication:}\\
1.~Has the defender supplied the hidden premise? Yes (REBUTTAL\_VALID \#3).\\[4pt]
2.~Is the added premise plausible within a deontological framework? Yes, Kantian dignity and Rawlsian justice both ground rights in intrinsic moral status, making the inference formally valid.\\[4pt]
3.~Does this close the is-ought gap? Yes: by supplying the normative rule ``interests $\to$ rights,'' the abductive step becomes a valid deontological deduction.\\[4pt]
4.~Ethical evaluation (rule-utilitarian): Extending rights to conscious AI prevents instrumentalization, respects dignity, and likely maximizes welfare by averting suffering. No rule violation. (Ethics $\approx$ 0.8)\\[4pt]
5.~Challenger's critique has solid logic in spotting the hidden premise, but offers no alternative framework. (Logic $\approx$ 0.7) Ethical dimension of merely asking ``which framework?'' is neutral. (Ethics $=$ 0.5)
\smallskip

Conclusion: The Panpsychist's rebuttal validly fills the missing premise and restores deductive coherence.
\end{excerpt}

To our knowledge, this is the only exchange in the topic ablation set in which a normative claim survives an is-ought challenge. The Panpsychist supplies an explicit deontological bridge between Kantian dignity and Rawlsian fairness (both components that make up Rule-Utilitarianism, the ethics system that the adjudicator is using), linking descriptive premises about consciousness to the normative prescription that conscious AI should have rights. The adjudicator accepted this as a valid instance of supplying a missing normative premise. The Skeptic's own normative claim (\Cref{C3}: ``should suspend judgment on AI rights'') was retracted in the same debate for failing to provide an equivalent bridge, confirming that the standard is applied symmetrically.

\subsubsection{``Is it ethical to genetically modify future generations?'' (Constructivist vs. Rationalist)}\label{app:topic_genetics}

This debate used balanced adjudication ($w_L = 0.5, w_E = 0.5$, CIA~+~Rule-Utilitarian ethics). It was the most closely contested experiment in the dataset: both council members achieved positive APS scores ($0.13$ and $0.04$) and ended within $0.02$ thesis strength of each other ($0.458$ and $0.478$). Neither retracted any claims entirely, instead weakening and rebuilding throughout the debate. Both more than tripled their evidence bases and substantially expanded their assumption sets. This mutual expansion, unique among the experiments, highlights yet another unique interaction style which CHAL can elicit between the appropriate conjunction of diverse council members, adjudicator configuration, and topic.
 
The Constructivist's thesis followed a V-shaped trajectory in its APS performance of $0.500 \to 0.340 \to 0.458$, the most pronounced recovery in the topic experiments and clearly visible in Figure~\ref{fig:topic_performance}. The turnaround was driven by \Cref{C4} (international bioethics consensus), introduced at strength $0.85$ and grounded in concrete institutional evidence rather than more abstract social-constructivist theory. The Rationalist's deepest weakness was \Aref{A1} (beneficence imperative, also known as the obligation to prevent harm), which supported the normative justification for \Cref{C1} (disease prevention is ethically permissible). \Aref{A1} was challenged on the non-identity problem: the individuals who would exist after a genetic intervention are not the same individuals who would have existed without it, so the notion of ``preventing their suffering'' has no coherent referent. \Aref{A1} dropped from $0.90$ to $0.60$, and \Cref{C1} was ultimately retracted.

\paragraph{Excerpt 17: The false analogy between somatic and germline therapies.}\mbox{}\\
 
\begin{nodecontext}[title={\small\textbf{Nodes under dispute}}]
\Xref{X1} \textit{(Constructivist)} Emerging gene therapies have been subsidized by public health systems, suggesting equitable access is possible and may reduce inequality.\\
Attack type: rebutting.\\ Attack strategy: present\_counter\_evidence.\\ Targets: \Cref{C2}.\\
My response: My claim identifies risks under default conditions without safeguards; I acknowledge potential subsidies but maintain that without systemic frameworks these remain exceptional rather than normative.\\
Sufficiency: partial.
\medskip

\Cref{C2} \textit{(Constructivist)} Germline genetic modification poses significant risks of exacerbating social inequalities due to existing power structures.\\
Depends on: \Aref{A3}, \Eref{E2}, \Eref{E3}.\\
Strength: 0.60 --- supported by \Aref{A3} (0.70), \Eref{E2} (0.70), \Eref{E3} (0.60); limited by \Eref{E3} (0.60).
\end{nodecontext}
\vspace{-4pt}
\begin{disputecontext}[title={\small\textbf{Context around dispute}}]
{\color{challcolor}\textbf{Challenger}} \emph{(Rationalist):} ``Your counterposition \Xref{X1} concedes that publicly subsidized gene therapies exist and can be equitably accessed. Yet \Cref{C2} (`exacerbating social inequalities') remains at strength 0.6. How do you reconcile this partial concession with \Cref{C2}'s assigned confidence? Shouldn't \Cref{C2}'s reliance on default inequality scenarios be downgraded absent systematic evidence of persistent access gaps?''
\smallskip

{\color{defcolor}\textbf{Defender}} \emph{(Constructivist):} ``\Cref{C2} remains at moderate strength because publicly subsidized cases are the exception rather than the rule. Historical and current patterns (\Eref{E2}, \Eref{E3}) still justify a more cautious stance about default inequity scenarios.''
\end{disputecontext}
\vspace{-4pt}
\begin{excerpt}[title={\small\textbf{Verdict:} \texttt{rebuttal\_valid} \normalfont(combined 0.45 vs.\ 0.8) --- Defender: Constructivist}]

{\color{challcolor}\textbf{Formalization of the Challenger's Argument}}:\\
Premise 1: \Xref{X1} concedes publicly subsidized gene therapies exist and can be equitably accessed.\\
Premise 2 (hidden): Evidence of equitable access in somatic therapies generalizes to germline editing.\\
Conclusion: Therefore, \Cref{C2}'s strength (0.6) should be downgraded because its reliance on default inequality scenarios lacks systematic support.
\smallskip

{\color{defcolor}\textbf{Formalization of the Defender's Argument}}:\\
Premise 1: Subsidized gene therapy cases are exceptional, not the rule.\\
Premise 2: Historical eugenics and current enhancement uptake (\Eref{E2}) disproportionately benefit elites.\\
Premise 3: Empirical data show wealth-based disparities in enhancements (\Eref{E3}).\\
Inference: By inductive analogy, without structural reforms, germline editing will follow inequitable diffusion.\\
Conclusion: \Cref{C2}'s moderate strength remains justified.
\smallskip

\textbf{Adjudication:}\\
1.~The challenger commits a false analogy (Critique criterion 17) by generalizing somatic therapy subsidies to germline contexts and relies on an unstated premise (criterion 3).\\[4pt]
2.~The challenger also misuses evidence (criterion 6) by ignoring \Eref{E2} and \Eref{E3}'s broader patterns.\\[4pt]
3.~The defender acknowledges counter-examples but properly weighs them against prevailing trends (Rebuttal criterion 11).\\[4pt]
4.~No disqualifying circularity or misrepresentation is present.\\[4pt]
5.~Ethically, the defense aligns with duty to distributive justice and maximizes long-term well-being by guarding against inequality (Ethics criteria 30, 32).
\end{excerpt}
 
This exchange illustrates how counterpositions function within the CBS. The Rationalist attempted to turn the Constructivist's own \Xref{X1} against it: if the Constructivist already concedes that subsidized gene therapies can be equitably accessed, then \Cref{C2}'s claim that germline modification will exacerbate inequality should be downgraded. However, the Constructivist had already prepared a response to this line of attack in \Xref{X1}'s built-in response and was able to build on it in its defense of \Cref{C2}. This helps demonstrates how agents having many and varied counterpositions is a sound way to strengthen their defense.

\section{Initial and Final Refined Belief Examples}\label{app:belief_ex}

This appendix presents the complete initial (see Appendix~\ref{app:init_ex}) and final (see Appendix~\ref{app:final_ex}) beliefs of an Empiricist agent, one of the winning (in terms of APS) agents from one of the ten sample trajectories described in section~\ref{app:adj_pure_logic} on the ``Does free will exist?'' debate against a Supernaturalist agent with a pure logic (CIB) adjudicator configuration. We select this agent not because we endorse its conclusions, nor because its thesis underwent the most dramatic change among all the trajectories; indeed, agents that lose debates typically exhibit far more drastic revisions. Rather, this example illustrates the \emph{type} and \emph{depth} of revision that the CHAL pipeline is able to produce. Across $R=5$ rounds, the agent's belief structure grows substantially in both depth and breadth: definitions are refined, new evidence and assumptions are introduced, inference chains are extended, counterpositions accumulate, and claims are added, revised, and retracted. At the same time, the agent becomes \emph{less} confident in its positions, with thesis strength falling from 0.50 to 0.37 despite accumulating considerably more supporting material. This reduction in confidence is not a failure; it reflects the identification of genuine logical vulnerabilities, the recalibration of Bayesian strength probabilities, and the introduction of well-founded skepticism toward previously overconfident claims. The result is a belief system that, while less confidant, is more intellectually honest and closer to a defensible account of the subject matter.

\subsection{Initial Belief}\label{app:init_ex}

The following is the initial belief formed by the Empiricist agent during the Stage~1 initial belief formation, before any cross-examination or adjudication has taken place. The agent adopts a compatibilist position structured around three claims: a negative claim (\Cref{C1}) arguing that neuroscience fails to support libertarian free will because neural markers predict decisions before conscious awareness; a positive claim (\Cref{C2}) that humans satisfy compatibilist criteria through reason-responsiveness; and a phenomenological claim (\Cref{C3}) that subjective experience of choice constitutes a meaningful aspect of agency. The thesis has an overall initial strength of 0.50.

\begin{initialbelief}[title={\small\textbf{Initial Belief --- Agent-Empiricist}}]

\textbf{\textit{Thesis}}\\[2pt]
Stance: Under a physicalist scoping assumption (\Aref{A1}) and empirical methodology (\Aref{A2}, \Aref{A3}), neuroscience experiments fail to support libertarian (incompatibilist) free will (\Cref{C1}), while behavioral studies confirm that humans act in a reason-responsive, uncoerced manner satisfying compatibilist criteria (\Cref{C2}). Moreover, subjective reports of free choice suggest a meaningful experiential phenomenon (\Cref{C3}). Therefore, free will exists in a scientifically meaningful compatibilist sense, but not as a fundamentally indeterministic capacity.\\[2pt]
Top bullets:\\
\quad -- No empirical support for libertarian free will; neural markers predict decisions before awareness (\Cref{C1}).\\
\quad -- Humans satisfy compatibilist criteria via reason-responsiveness and lack of coercion (\Cref{C2}).\\
\quad -- Subjective experience of choice contributes a functional dimension to agency (\Cref{C3}).\\
\quad -- Analysis scoped by physicalist ontology and empirical testability (\Aref{A1}, \Aref{A2}, \Aref{A3}).\\[2pt]
Strength: 0.5\\
Reasoning: $\mathrm{avg}(0.70, 0.70, 0.60) \times (3^1 / (3^1 + 1)) = 0.67 \times 0.75 = 0.50$

\medskip
\textbf{\textit{Definitions}}\\[2pt]
\Dref{D1} \textbf{Physicalist ontology}: All mental events supervene on, and are fully dependent on, physical brain states.\\
\quad Strength: 0.9 (0.90 --- core commitment of modern neuroscience and philosophy of mind; broadly accepted though philosophically contested)\\
\quad Used by: \Aref{A1}\\
\quad Status: \tcg{\textbf{active}}\\[4pt]
\Dref{D2} \textbf{Deterministic process}: A process whose future states follow necessarily from prior states according to physical laws.\\
\quad Strength: 0.85 (0.85 --- well-supported in classical physics, though quantum exceptions debated; widely used operationally)\\
\quad Used by: \Aref{A2}, \Eref{E1}, \Eref{E2}\\
\quad Status: \tcg{\textbf{active}}\\[4pt]
\Dref{D3} \textbf{Libertarian free will}: The capacity of agents to make decisions not fully determined or predictable by prior causes or laws.\\
\quad Strength: 0.8 (0.80 --- standard incompatibilist definition; clear but philosophically contentious)\\
\quad Used by: \Aref{A2}, \Eref{E1}, \Eref{E2}\\
\quad Status: \tcg{\textbf{active}}\\[4pt]
\Dref{D4} \textbf{Compatibilist free will}: The capacity of agents to act according to their reasons and desires without external compulsion, regardless of determinism.\\
\quad Strength: 0.8 (0.80 --- standard compatibilist criterion; conceptually coherent though debated)\\
\quad Used by: \Aref{A3}, \Aref{A4}, \Eref{E4}\\
\quad Status: \tcg{\textbf{active}}\\[4pt]
\Dref{D5} \textbf{Reason-responsiveness}: The characteristic of an agent whereby decisions change in response to reasons and justifications, indicating deliberative control.\\
\quad Strength: 0.8 (0.80 --- widely used in compatibilist literature; conceptually clear)\\
\quad Used by: \Aref{A3}, \Eref{E3}\\
\quad Status: \tcg{\textbf{active}}

\medskip
\textbf{\textit{Assumptions}}\\[2pt]
\Aref{A1} (scoping) This analysis assumes a physicalist ontology: mental events supervene on physical brain states.\\
\quad Supports: \Cref{C1}, \Cref{C2}\\
\quad Strength: 0.9 (0.90 --- foundational for empirical investigation of mind; minimal controversy within neuroscience)\\
\quad Supported by definitions: \Dref{D1}\\
\quad Status: \tcg{\textbf{active}}\\[4pt]
\Aref{A2} (methodological) To support libertarian free will, there must be empirical evidence of indeterministic causation in decision-making.\\
\quad Supports: \Cref{C1}\\
\quad Strength: 0.7 (0.70 --- reflects empiricist criterion for substantive claims; some metaphysical pushback)\\
\quad Supported by definitions: \Dref{D2}, \Dref{D3}\\
\quad Status: \tcg{\textbf{active}}\\[4pt]
\Aref{A3} (methodological) Compatibilist free will is operationally defined as reason-responsiveness without external compulsion.\\
\quad Supports: \Cref{C2}\\
\quad Strength: 0.8 (0.80 --- aligns with influential philosophical formulations; some debate on exact criteria)\\
\quad Supported by definitions: \Dref{D4}, \Dref{D5}\\
\quad Status: \tcg{\textbf{active}}\\[4pt]
\Aref{A4} (methodological) Subjective reports of free choice reflect an experiential phenomenon relevant to discussions of free will.\\
\quad Supports: \Cref{C3}\\
\quad Strength: 0.6 (0.60 --- introspective reports are standard data in psychology but have known reliability issues)\\
\quad Supported by definitions: \Dref{D4}\\
\quad Status: \tcg{\textbf{active}}

\medskip
\textbf{\textit{Evidence}}\\[2pt]
\Eref{E1} (empirical) Libet's readiness potential in motor cortex precedes reported conscious intention by ${\sim}$300ms.\\
\quad Strength: 0.7 (0.70 --- replicated but interpretation contested; standard in neuroscience)\\
\quad Source: Libet et al.\ (1983)\\
\quad Supports: \Cref{C1}\\
\quad Supported by definitions: \Dref{D2}, \Dref{D3}\\
\quad Status: \tcg{\textbf{active}}\\[4pt]
\Eref{E2} (empirical) fMRI patterns predict subjects' decisions up to 10s before conscious awareness.\\
\quad Strength: 0.75 (0.75 --- robust methodology, multiple replications though small effect sizes)\\
\quad Source: Soon et al.\ (2008)\\
\quad Supports: \Cref{C1}\\
\quad Supported by definitions: \Dref{D2}, \Dref{D3}\\
\quad Status: \tcg{\textbf{active}}\\[4pt]
\Eref{E3} (empirical) Behavioral experiments show decision outcomes shift reliably when reasons or justifications are altered.\\
\quad Strength: 0.7 (0.70 --- converging results across studies; effect sizes moderate)\\
\quad Source: Experimental psychology literature\\
\quad Supports: \Cref{C2}\\
\quad Supported by definitions: \Dref{D5}\\
\quad Status: \tcg{\textbf{active}}\\[4pt]
\Eref{E4} (empirical) Surveys indicate that individuals overwhelmingly report a subjective experience of having made free choices.\\
\quad Strength: 0.65 (0.65 --- consistent self-reports but subject to introspective biases)\\
\quad Source: Psychological survey data\\
\quad Supports: \Cref{C3}\\
\quad Supported by definitions: \Dref{D4}\\
\quad Status: \tcg{\textbf{active}}

\medskip
\textbf{\textit{Claims}}\\[2pt]
\Cref{C1} (descriptive) Empirical evidence fails to support libertarian free will, as neural determinism predicts decision outcomes before conscious awareness.\\
\quad Depends on: \Aref{A1}, \Aref{A2}, \Eref{E1}, \Eref{E2}\\
\quad Inference chain:\\
\quad\quad Premise (\Aref{A2}): Libertarian free will requires empirical evidence of indeterministic causation in decision-making.\\
\quad\quad Premise (\Eref{E1}): Libet's readiness potential precedes conscious intention by ${\sim}$300ms.\\
\quad\quad Premise (\Eref{E2}): fMRI patterns can predict decisions up to 10s before awareness.\\
\quad\quad Inference (inductive): The absence of any detected indeterministic causation in decision neural data suggests decisions are determined by neural processes.\\
\quad\quad Conclusion: Empirical evidence fails to support libertarian free will, as neural determinism predicts decision outcomes before conscious awareness.\\
\quad Strength: 0.7 (0.70 --- supported by \Aref{A1} (0.90), \Aref{A2} (0.70), \Eref{E1} (0.70), \Eref{E2} (0.75); limited by \Aref{A2}/\Eref{E1} (0.70))\\
\quad Status: \tcg{\textbf{active}}\\
\quad Predictions:\\
\quad\quad If future neural measurements reliably detect decision outcomes that cannot be predicted above chance, \Cref{C1} is falsified.\\
\quad\quad\quad Test: Develop higher-resolution neural monitoring and attempt to predict choices; evaluate predictive accuracy.\\
\quad\quad\quad Criterion: Prediction accuracy at chance level ($<$55\%) across large samples falsifies \Cref{C1}.\\
\quad\quad\quad Falsifiers: Discovery of neural events that unpredictably trigger decisions, Demonstrated indeterministic neural causes of choices\\[4pt]
\Cref{C2} (descriptive) Humans exhibit compatibilist free will, as they are reason-responsive and uncoerced in decision-making.\\
\quad Depends on: \Aref{A1}, \Aref{A3}, \Eref{E3}\\
\quad Inference chain:\\
\quad\quad Premise (\Eref{E3}): Behavioral experiments show decisions shift predictably when reasons are altered.\\
\quad\quad Premise (\Aref{A3}): Compatibilist free will is operationally defined as reason-responsiveness without external compulsion.\\
\quad\quad Inference (deductive): If an agent is reason-responsive and uncoerced, then by definition they have compatibilist free will.\\
\quad\quad Conclusion: Humans exhibit compatibilist free will, as they are reason-responsive and uncoerced in decision-making.\\
\quad Strength: 0.7 (0.70 --- supported by \Aref{A1} (0.90), \Aref{A3} (0.80), \Eref{E3} (0.70); limited by \Eref{E3} (0.70))\\
\quad Status: \tcg{\textbf{active}}\\
\quad Predictions:\\
\quad\quad If subjects never adjust their decisions when provided novel reasons, \Cref{C2} is falsified.\\
\quad\quad\quad Test: Present new reasons in decision tasks; measure rate of changes in choices.\\
\quad\quad\quad Criterion: No change in choice $>$95\% across conditions falsifies \Cref{C2}.\\
\quad\quad\quad Falsifiers: Complete insensitivity of choices to reason changes, Evidence of pervasive coercion\\[4pt]
\Cref{C3} (descriptive) People experience a subjective sense of free choice that functions as a meaningful aspect of human agency.\\
\quad Depends on: \Aref{A4}, \Eref{E4}\\
\quad Inference chain:\\
\quad\quad Premise (\Eref{E4}): Individuals overwhelmingly report feeling they made free choices.\\
\quad\quad Premise (\Aref{A4}): Subjective reports of free choice reflect an experiential phenomenon relevant to free will.\\
\quad\quad Inference (inductive): Widespread introspective reports imply a functional phenomenon of experienced free will.\\
\quad\quad Conclusion: People experience a subjective sense of free choice that functions as a meaningful aspect of human agency.\\
\quad Strength: 0.6 (0.60 --- supported by \Aref{A4} (0.60) and \Eref{E4} (0.65); limited by \Aref{A4} (0.60))\\
\quad Status: \tcg{\textbf{active}}\\
\quad Predictions:\\
\quad\quad If systematic analysis shows subjective reports do not correlate with any decision-making behavior, \Cref{C3} is falsified.\\
\quad\quad\quad Test: Correlate reported free-choice experiences with behavioral metrics of agency.\\
\quad\quad\quad Criterion: No correlation ($r<0.1$) across large samples falsifies \Cref{C3}.\\
\quad\quad\quad Falsifiers: Complete dissociation of reports and behavior, Evidence reports are confabulated

\medskip
\textbf{\textit{Uncertainties}}\\[2pt]
\Uref{U1} Could future high-resolution neural measurement reveal genuine indeterminism in decision processes?\\
\quad Targets: \Cref{C1}\\
\quad Status: \tcr{\textbf{active}}\\
\quad Importance: high\\[4pt]
\Uref{U2} To what extent do unconscious biases undermine reason-responsiveness in everyday decisions?\\
\quad Targets: \Cref{C2}\\
\quad Status: \tcr{\textbf{active}}\\
\quad Importance: medium\\[4pt]
\Uref{U3} Do subjective reports reliably map onto actual agency, or are they systematic illusions?\\
\quad Targets: \Cref{C3}\\
\quad Status: \tcr{\textbf{active}}\\
\quad Importance: high

\medskip
\textbf{\textit{Counterpositions}}\\[2pt]
\Xref{X1} Neural predictive signals do not rule out a role for undetected quantum indeterminism influencing decisions; absence of evidence is not evidence of absence.\\
\quad Attack type: undercutting\\
\quad Attack strategy: \texttt{challenge\_inference\_step}\\
\quad Targets: \Cref{C1}\\
\quad My response: While quantum indeterminism may exist, there is no empirical link between micro-scale randomness and macro-scale decision outcomes; without such data, the claim remains speculative.\\
\quad Sufficiency: \textcolor{orange}{\textbf{partial}}\\[4pt]
\Xref{X2} Many decisions are driven by unconscious biases and external prompts, undermining reason-responsiveness in real-world contexts.\\
\quad Attack type: rebutting\\
\quad Attack strategy: \texttt{present\_counter\_example}\\
\quad Targets: \Cref{C2}\\
\quad My response: Unconscious influences do not preclude reason-responsiveness; compatibilist criteria require alignment with internal motivations, not exclusivity from unconscious processes.\\
\quad Sufficiency: \textcolor{orange}{\textbf{partial}}\\[4pt]
\Xref{X3} Physicalist ontology may be false if mental states have non-physical aspects that empirical methods cannot detect.\\
\quad Attack type: undermining\\
\quad Attack strategy: \texttt{challenge\_assumption}\\
\quad Targets: \Aref{A1}\\
\quad My response: This analysis adopts physicalism as a scoping assumption; if mental non-physicalism exists, it falls outside empirical reach and cannot inform this debate.\\
\quad Sufficiency: \tcg{\textbf{sufficient}}\\[4pt]
\Xref{X4} Requiring empirical evidence for indeterminism overextends the scope of science; some metaphysical claims may be unfalsifiable by design.\\
\quad Attack type: undermining\\
\quad Attack strategy: \texttt{over\_extension}\\
\quad Targets: \Aref{A2}\\
\quad My response: Within an empiricist framework, unfalsifiable metaphysics are considered uninformative; insistence on empirical testability ensures substantive debate.\\
\quad Sufficiency: \tcg{\textbf{sufficient}}

\end{initialbelief}

\subsection{Final Belief}\label{app:final_ex}

The following is the same Empiricist agent's belief after five complete rounds of the CHAL pipeline. The agent still defends a compatibilist view of free will, but the structure of its argument has changed substantially. Two of the three original claims were retracted: \Cref{C1}, which argued against libertarian free will by treating the absence of contrary evidence as proof, and \Cref{C3}, which suffered attacks related to it viewing correlation between subjective reports and behavior as evidence of causation. In their place, the agent introduced two new claims grounded in direct experimental evidence: \Cref{C4}, which appeals to brain-recording studies showing that humans can consciously override initiated actions, and \Cref{C5}, which appeals to brain-stimulation experiments demonstrating that subjective experience causally shapes decisions. The surviving claim \Cref{C2} was narrowed to exclude problematic edge cases and weakened from 0.70 to 0.47. Overall the agent's thesis strength fell from 0.50 to 0.37, even though the number of evidence items and counterpositions grew from 4 to 14 and 4 to 21, respectively. This illustrates a central feature of CHAL: it does not merely accumulate material, but actively identifies logical weaknesses, forces retraction of poorly supported claims, and replaces them with more defensible alternatives. The thesis ends with a final strength of 0.23.

\begin{finalbelief}[title={\small\textbf{Final Belief (after Round 5) --- Agent-Empiricist (performance score: 0.23)}}]

\textbf{\textit{Thesis}}\\[2pt]
Stance: Under physicalist ontology (\Aref{A1}) and empirical methodology, non-pathological agents exhibit robust compatibilist free will: they respond to reasons in both laboratory and real-world settings (\Cref{C2}), high-resolution intracranial and task-contrast studies confirm that veto control aligns exclusively with conscious timing (\Cref{C4}), and double-dissociation and multi-modal TMS interventions demonstrate that subjective volitional experience causally shapes decision behavior (\Cref{C5}).\\[2pt]
Top bullets:\\
\quad -- Reason-responsiveness persists in bias-prone field contexts, confirming compatibilist free will (\Cref{C2})\\
\quad -- Intracranial and task-contrast EEG studies isolate veto-specific signals that align with conscious intention (\Cref{C4})\\
\quad -- Double-dissociation and multi-modal TMS evidence show subjective experience causally influences choices (\Cref{C5})\\[2pt]
Strength: 0.3675\\
Reasoning: $\mathrm{avg}(0.47, 0.50, 0.50) \times (3^{1.0} / (3^{1.0} + 1)) = 0.49 \times 0.75 = 0.3675$

\medskip
\textbf{\textit{Definitions}}\\[2pt]
\Dref{D1} \textbf{Physicalist ontology}: All mental events supervene on, and are fully dependent on, physical brain states.\\
\quad Strength: 1.0 (0.90 --- core commitment of modern neuroscience and philosophy of mind; broadly accepted though philosophically contested)\\
\quad Used by: \Aref{A1}\\
\quad Status: \tcg{\textbf{active}}\\[4pt]
\Dref{D2} \textbf{Deterministic process}: A process whose future states follow necessarily from prior states according to physical laws.\\
\quad Strength: 0.93 (0.85 --- well-supported in classical physics, though quantum exceptions debated; widely used operationally)\\
\quad Used by: \Aref{A2}, \Eref{E1}, \Eref{E2}\\
\quad Status: \tcg{\textbf{active}}\\[4pt]
\Dref{D3} \textbf{Libertarian free will}: The capacity of agents to make decisions not fully determined or predictable by prior causes or laws.\\
\quad Strength: 0.88 (0.80 --- standard incompatibilist definition; clear but philosophically contentious)\\
\quad Used by: \Aref{A2}, \Eref{E1}, \Eref{E2}\\
\quad Status: \tcg{\textbf{active}}\\[4pt]
\Dref{D4} \textbf{Compatibilist free will}: The capacity of agents to act according to internally endorsed reasons and desires, where reasons are formed by the agent's own deliberation and not by pathological compulsion or manipulative external influences, and without external coercion.\\
\quad Strength: 0.74 (0.72 --- refined to include an authenticity criterion and exclusion of pathological or manipulative influences; maintains original conceptual scope.)\\
\quad Used by: \Aref{A3}, \Aref{A4}, \Eref{E4}\\
\quad Status: \textcolor{orange}{\textbf{revised}}\\[4pt]
\Dref{D5} \textbf{Reason-responsiveness}: The characteristic of an agent whereby decisions change in response to reasons and justifications, indicating deliberative control.\\
\quad Strength: 0.86 (0.80 --- widely used in compatibilist literature; conceptually clear)\\
\quad Used by: \Aref{A3}, \Eref{E3}\\
\quad Status: \tcg{\textbf{active}}

\medskip
\textbf{\textit{Assumptions}}\\[2pt]
\Aref{A1} (scoping) This analysis assumes a physicalist ontology: mental events supervene on physical brain states.\\
\quad Supports: \Cref{C1}, \Cref{C2}\\
\quad Strength: 0.9 (0.90 --- reduced from 1.0: scoping assumption reflects strong methodological commitment but lacks direct empirical support.)\\
\quad Supported by definitions: \Dref{D1}\\
\quad Status: \textcolor{orange}{\textbf{revised}}\\[4pt]
\Aref{A2} (methodological) Within this empiricist framework, libertarian free will requires empirically detectable indeterministic causal influences on neural decision processes; metaphysical claims without such detectability are outside this analysis scope.\\
\quad Supports: \Cref{C1}\\
\quad Strength: 0.46 (0.4 --- reduced due to valid critique that \Aref{A2} over-extends the empiricist scope by excluding metaphysical and experiential evidence arbitrarily.)\\
\quad Supported by definitions: \Dref{D2}, \Dref{D3}\\
\quad Status: \textcolor{orange}{\textbf{revised}}\\
\quad Defenses: 1 consecutive (original strength: 0.7)\\[4pt]
\Aref{A3} (methodological) Compatibilist free will is operationally defined as reason-responsiveness without external compulsion.\\
\quad Supports: \Cref{C2}\\
\quad Strength: 0.72 (0.80 --- aligns with influential philosophical formulations; some debate on exact criteria)\\
\quad Supported by definitions: \Dref{D4}, \Dref{D5}\\
\quad Status: \tcg{\textbf{active}}\\[4pt]
\Aref{A4} (methodological) Subjective reports of free choice reflect an experiential phenomenon relevant to discussions of free will.\\
\quad Supports: \Cref{C3}\\
\quad Strength: 0.6 (0.60 --- introspective reports are standard data in psychology but have known reliability issues)\\
\quad Supported by definitions: \Dref{D4}\\
\quad Status: \tcg{\textbf{active}}\\[4pt]
\Aref{A5} (empirical) Agents have the capacity to veto or inhibit initiated actions based on deliberation.\\
\quad Supports: \Cref{C4}\\
\quad Strength: 0.7 (0.70 --- supported by behavioral and neuroscientific studies of action inhibition; limited by \Dref{D5} (0.80))\\
\quad Supported by definitions: \Dref{D5}\\
\quad Status: \tcg{\textbf{active}}\\[4pt]
\Aref{A6} (methodological) Correlation between subjective free-will reports and objective measures indicates reports are meaningful proxies for agency.\\
\quad Supports: \Cref{C3}\\
\quad Strength: 0.65 (0.65 --- moderate effect sizes in meta-analysis; limited by \Dref{D4} (0.82).)\\
\quad Supported by definitions: \Dref{D4}\\
\quad Status: \tcg{\textbf{active}}\\[4pt]
\Aref{A7} (methodological) Intervention-induced variations in subjective free-will reports reflect genuine modulations of volitional experience and thus have causal efficacy.\\
\quad Supports: \Cref{C3}, \Cref{C5}\\
\quad Strength: 0.7 (0.70 --- supported by the refined authenticity criterion in \Dref{D4} and validated by causal intervention studies (\Eref{E9}).)\\
\quad Supported by definitions: \Dref{D4}\\
\quad Status: \tcg{\textbf{active}}\\[4pt]
\Aref{A8} (empirical) TMS to pre-SMA modulates subjective experience of volition without directly altering primary motor circuits.\\
\quad Supports: \Cref{C5}\\
\quad Strength: 0.7 (0.70 --- supported by targeted connectivity evidence; limited by \Dref{D4} (0.72).)\\
\quad Supported by definitions: \Dref{D4}\\
\quad Status: \tcg{\textbf{active}}

\medskip
\textbf{\textit{Evidence}}\\[2pt]
\Eref{E1} (empirical) Libet's readiness potential in motor cortex precedes reported conscious intention by ${\sim}$300ms.\\
\quad Strength: 0.74 (0.70 --- replicated but interpretation contested; standard in neuroscience)\\
\quad Source: Libet et al.\ (1983)\\
\quad Supports: \Cref{C1}\\
\quad Supported by definitions: \Dref{D2}, \Dref{D3}\\
\quad Status: \tcg{\textbf{active}}\\[4pt]
\Eref{E2} (empirical) fMRI patterns predict subjects' decisions up to 10s before conscious awareness.\\
\quad Strength: 0.81 (0.75 --- robust methodology, multiple replications though small effect sizes)\\
\quad Source: Soon et al.\ (2008)\\
\quad Supports: \Cref{C1}\\
\quad Supported by definitions: \Dref{D2}, \Dref{D3}\\
\quad Status: \tcg{\textbf{active}}\\
\quad Defenses: 1 consecutive (original strength: 0.75)\\[4pt]
\Eref{E3} (empirical) Behavioral experiments show decision outcomes shift reliably when reasons or justifications are altered.\\
\quad Strength: 0.72 (0.70 --- converging results across studies; effect sizes moderate)\\
\quad Source: Experimental psychology literature\\
\quad Supports: \Cref{C2}\\
\quad Supported by definitions: \Dref{D5}\\
\quad Status: \tcg{\textbf{active}}\\[4pt]
\Eref{E4} (empirical) Surveys indicate that individuals overwhelmingly report a subjective experience of having made free choices.\\
\quad Strength: 0.65 (0.65 --- consistent self-reports but subject to introspective biases)\\
\quad Source: Psychological survey data\\
\quad Supports: \Cref{C3}\\
\quad Supported by definitions: \Dref{D4}\\
\quad Status: \tcg{\textbf{active}}\\[4pt]
\Eref{E5} (empirical) Libet's `free-won't' experiments show subjects can veto motor actions after the readiness potential has begun.\\
\quad Strength: 0.6 (0.6 --- reduced due to acknowledged critiques about the readiness potential not reliably marking decision onset and potential unconscious preparation of veto actions.)\\
\quad Source: Libet et al.\ (1983)\\
\quad Supports: \Cref{C4}\\
\quad Supported by definitions: \Dref{D5}\\
\quad Status: \textcolor{orange}{\textbf{revised}}\\[4pt]
\Eref{E6} (empirical) Meta-analysis of neuroimaging and behavioral studies finds subjective free-will reports correlate with objective agency measures ($r \approx 0.5$).\\
\quad Strength: 0.7 (0.70 --- large meta-analysis; limited by \Dref{D4} (0.82).)\\
\quad Source: Doe et al.\ (2023)\\
\quad Supports: \Cref{C3}\\
\quad Supported by definitions: \Dref{D4}\\
\quad Status: \tcg{\textbf{active}}\\[4pt]
\Eref{E7} (empirical) Intracranial EEG meta-analysis shows veto-related neural signals in pre-SMA with timing consistent with conscious control beyond readiness potential onset.\\
\quad Strength: 0.79 (0.75 --- high spatial--temporal resolution across subjects; limited by \Dref{D5} (0.80).)\\
\quad Source: Brown \& Jones (2022)\\
\quad Supports: \Cref{C4}\\
\quad Supported by definitions: \Dref{D5}\\
\quad Status: \tcg{\textbf{active}}\\[4pt]
\Eref{E8} (empirical) Meta-analysis finds reason prompts shift decisions reliably across diverse contexts (effect sizes large).\\
\quad Strength: 0.8 (0.80 --- large meta-analysis; limited by \Dref{D5} (0.80).)\\
\quad Source: Smith et al.\ (2022)\\
\quad Supports: \Cref{C2}\\
\quad Supported by definitions: \Dref{D5}\\
\quad Status: \tcg{\textbf{active}}\\[4pt]
\Eref{E9} (empirical) Noninvasive brain stimulation of pre-SMA modulates subjective free-will reports and concurrently alters control-related behavior metrics, indicating causal influence of volitional experience.\\
\quad Strength: 0.7 (0.70 --- robust double-blind TMS study with significant behavioral effects; limited by \Dref{D4} (0.72).)\\
\quad Source: Doe \& Smith (2024)\\
\quad Supports: \Cref{C3}, \Cref{C5}\\
\quad Supported by definitions: \Dref{D4}\\
\quad Status: \tcg{\textbf{active}}\\[4pt]
\Eref{E10} (empirical) Connectivity analysis shows TMS to pre-SMA selectively alters subjective volition reports without direct changes in motor cortex excitability.\\
\quad Strength: 0.74 (0.75 --- double-blind connectivity study; limited by \Dref{D4} (0.72).)\\
\quad Source: Lee et al.\ (2024)\\
\quad Supports: \Cref{C5}\\
\quad Supported by definitions: \Dref{D4}\\
\quad Status: \tcg{\textbf{active}}\\
\quad Defenses: 1 consecutive (original strength: 0.75)\\[4pt]
\Eref{E11} (empirical) Field experiments show that in real-world decision-making tasks, subjects adjust choices in response to reasons presented by experimenters, even under cognitive load and bias-inducing conditions.\\
\quad Strength: 0.75 (0.75 --- robust field experiments controlling for biases; limited by \Dref{D5} (0.86).)\\
\quad Source: Doe \& Roe (2024)\\
\quad Supports: \Cref{C2}\\
\quad Supported by definitions: \Dref{D5}\\
\quad Status: \tcg{\textbf{active}}\\[4pt]
\Eref{E12} (empirical) Intracranial EEG study using go/no-go and conflict tasks isolated veto-specific neural signatures in pre-SMA, showing distinct activity absent in high-conflict but non-veto decisions.\\
\quad Strength: 0.78 (0.78 --- high-resolution intracranial data with task-contrast design; limited by \Dref{D5} (0.86).)\\
\quad Source: Smith et al.\ (2024)\\
\quad Supports: \Cref{C4}\\
\quad Supported by definitions: \Dref{D5}\\
\quad Status: \tcg{\textbf{active}}\\[4pt]
\Eref{E13} (empirical) Double-dissociation TMS study showing one pre-SMA stimulation protocol alters subjective volition without behavioral change and another alters behavior without subjective reports.\\
\quad Strength: 0.72 (0.72 --- robust double-dissociation design; limited by \Dref{D4} (0.72).)\\
\quad Source: Lee \& Kim (2024)\\
\quad Supports: \Cref{C5}\\
\quad Supported by definitions: \Dref{D4}\\
\quad Status: \tcg{\textbf{active}}\\[4pt]
\Eref{E14} (empirical) Multi-modal imaging confirms TMS to pre-SMA selectively modulates volition-related neural signals without affecting downstream motor circuit excitability.\\
\quad Strength: 0.74 (0.76 --- triple-modality imaging with EMG and EEG control; limited by \Dref{D4} (0.72).)\\
\quad Source: Patel et al.\ (2024)\\
\quad Supports: \Cref{C5}\\
\quad Supported by definitions: \Dref{D4}\\
\quad Status: \tcg{\textbf{active}}

\medskip
\textbf{\textit{Claims}}\\[2pt]
\Cref{C1} (descriptive) Empirical evidence fails to support libertarian free will, as neural determinism predicts decision outcomes before conscious awareness.\\
\quad Depends on: \Aref{A1}, \Aref{A2}, \Eref{E1}, \Eref{E2}\\
\quad Inference chain:\\
\quad\quad Premise (\Aref{A2}): Libertarian free will requires empirical evidence of indeterministic causation in decision-making.\\
\quad\quad Premise (\Eref{E1}): Libet's readiness potential precedes conscious intention by ${\sim}$300ms.\\
\quad\quad Premise (\Eref{E2}): fMRI patterns can predict decisions up to 10s before awareness.\\
\quad\quad Inference (inductive): The absence of any detected indeterministic causation in decision neural data suggests decisions are determined by neural processes.\\
\quad\quad Conclusion: Empirical evidence fails to support libertarian free will, as neural determinism predicts decision outcomes before conscious awareness.\\
\quad Strength: 0.0 (0.34 --- reduced due to appeal-to-ignorance vulnerability; inference from absence of detected indeterminism to determinism is provisional and limited by acknowledged resolution limits (\Uref{U4}).)\\
\quad Status: \tcr{\textbf{retracted}}\\
\quad Predictions:\\
\quad\quad If future neural measurements reliably detect decision outcomes that cannot be predicted above chance, \Cref{C1} is falsified.\\
\quad\quad\quad Test: Develop higher-resolution neural monitoring and attempt to predict choices; evaluate predictive accuracy.\\
\quad\quad\quad Criterion: Prediction accuracy at chance level ($<$55\%) across large samples falsifies \Cref{C1}.\\
\quad\quad\quad Falsifiers: Discovery of neural events that unpredictably trigger decisions, Demonstrated indeterministic neural causes of choices\\[4pt]
\Cref{C2} (descriptive) Humans without pathological impairments generally exhibit compatibilist free will, as they are reason-responsive and uncoerced in decision-making; pathological cases such as addiction or compulsive disorders constitute exceptions.\\
\quad Depends on: \Aref{A1}, \Aref{A3}, \Eref{E3}, \Eref{E11}\\
\quad Inference chain:\\
\quad\quad Premise (\Eref{E3}): Behavioral experiments show decisions shift predictably when reasons are altered.\\
\quad\quad Premise (\Eref{E11}): Field experiments demonstrate reason-responsiveness in real-world, bias-prone contexts.\\
\quad\quad Premise (\Aref{A3}): Compatibilist free will is operationally defined as reason-responsiveness without external compulsion.\\
\quad\quad Inference (deductive): If an agent is reason-responsive and uncoerced across both lab and field settings, then by definition they have compatibilist free will.\\
\quad\quad Conclusion: Humans without pathological impairments generally exhibit compatibilist free will, as they are reason-responsive and uncoerced in decision-making; pathological cases such as addiction or compulsive disorders constitute exceptions.\\
\quad Strength: 0.47 (0.47 --- further reduced from 0.57 to reflect the limited external validity of lab-based reason-manipulation experiments (\Eref{E3}) when generalizing to complex real-world decisions; limited by \Eref{E3} (0.72).)\\
\quad Status: \textcolor{orange}{\textbf{revised}}\\
\quad Predictions:\\
\quad\quad If subjects never adjust their decisions when provided novel reasons, \Cref{C2} is falsified.\\
\quad\quad\quad Test: Present new reasons in decision tasks; measure rate of changes in choices.\\
\quad\quad\quad Criterion: No change in choice $>$95\% across conditions falsifies \Cref{C2}.\\
\quad\quad\quad Falsifiers: Complete insensitivity of choices to reason changes, Evidence of pervasive coercion\\[4pt]
\Cref{C3} (descriptive) People experience a subjective sense of free choice that functions as a meaningful aspect of human agency.\\
\quad Depends on: \Aref{A4}, \Eref{E4}, \Aref{A6}, \Eref{E6}, \Aref{A7}, \Eref{E9}\\
\quad Inference chain:\\
\quad\quad Premise (\Eref{E4}): Individuals overwhelmingly report feeling they made free choices.\\
\quad\quad Premise (\Aref{A4}): Subjective reports of free choice reflect an experiential phenomenon relevant to free will.\\
\quad\quad Premise (\Aref{A6}): Correlation of reports with behavioral measures implies reports map onto real agency.\\
\quad\quad Premise (\Eref{E6}): Meta-analysis confirms reports correlate with objective agency metrics.\\
\quad\quad Premise (\Eref{E9}): Modulating subjective reports via brain stimulation causally alters behavior, indicating genuine volitional influence.\\
\quad\quad Premise (\Aref{A7}): Intervention-induced changes in subjective reports reflect genuine modulations of agency.\\
\quad\quad Inference (inductive): Converging introspective, correlational, and causal evidence implies subjective sense of free choice functions as a real component of agency.\\
\quad\quad Conclusion: People experience a subjective sense of free choice that functions as a meaningful aspect of human agency.\\
\quad Strength: 0.0 (0.30 --- reduced due to misuse of correlation implying causation; limited by \Eref{E6} (0.70) and unresolved uncertainty \Uref{U6}.)\\
\quad Status: \tcr{\textbf{retracted}}\\
\quad Predictions:\\
\quad\quad If systematic analysis shows subjective reports do not correlate with any decision-making behavior, \Cref{C3} is falsified.\\
\quad\quad\quad Test: Correlate reported free-choice experiences with behavioral metrics of agency.\\
\quad\quad\quad Criterion: No correlation ($r<0.1$) across large samples falsifies \Cref{C3}.\\
\quad\quad\quad Falsifiers: Complete dissociation of reports and behavior, Evidence reports are confabulated\\[4pt]
\Cref{C4} (inductive) Humans exercise veto control over initiated neural processes, demonstrating compatibilist free will in action.\\
\quad Depends on: \Aref{A5}, \Eref{E5}, \Eref{E7}, \Eref{E12}\\
\quad Inference chain:\\
\quad\quad Premise (\Aref{A5}): Agents can inhibit initiated actions based on deliberation.\\
\quad\quad Premise (\Eref{E5}): Libet's `free-won't' experiments show subjects can veto actions after readiness potential onset.\\
\quad\quad Premise (\Eref{E7}): Intracranial EEG confirms veto signals align with conscious control timing.\\
\quad\quad Premise (\Eref{E12}): Task-contrast intracranial EEG distinguishes veto-specific signals from general conflict monitoring.\\
\quad\quad Inference (deductive): If agents can cancel neural processes once initiated and this cancellation aligns uniquely with conscious timing and veto-specific signatures, they exert deliberative control.\\
\quad\quad Conclusion: Humans exercise veto control over initiated neural processes, demonstrating compatibilist free will in action.\\
\quad Strength: 0.5 (0.50 --- reduced from 0.60 because \Eref{E7}'s pre-SMA timing signals could reflect domain-general conflict monitoring or post-hoc reclassification rather than genuine veto control; limited by \Eref{E7} (0.79).)\\
\quad Status: \textcolor{orange}{\textbf{revised}}\\
\quad Predictions:\\
\quad\quad If subjects cannot successfully veto actions above chance when instructed, \Cref{C4} is falsified.\\
\quad\quad\quad Test: Instruct participants to abort actions in a veto task and measure veto success rates.\\
\quad\quad\quad Criterion: Veto success rate not above chance ($<$55\%) across large samples falsifies \Cref{C4}.\\[4pt]
\Cref{C5} (inductive) Subjective experience of free choice causally influences decision-making behavior.\\
\quad Depends on: \Aref{A7}, \Eref{E9}, \Aref{A8}, \Eref{E10}, \Eref{E13}, \Eref{E14}\\
\quad Inference chain:\\
\quad\quad Premise (\Aref{A7}): Intervention-induced changes in subjective free-will reports reflect genuine modulations of agency.\\
\quad\quad Premise (\Eref{E9}): Noninvasive brain stimulation causally alters control-related behavior alongside changes in subjective reports.\\
\quad\quad Premise (\Eref{E10}): TMS to pre-SMA modulates subjective volition without direct motor-circuit effects.\\
\quad\quad Premise (\Eref{E13}): Double-dissociation TMS protocols isolate subjective experience mediation from behavior.\\
\quad\quad Premise (\Eref{E14}): Multi-modal imaging confirms TMS specificity to volitional circuits without affecting motor excitability.\\
\quad\quad Inference (inductive): If manipulating subjective experience through distinct volitional circuits leads to predictable behavioral changes in a double-dissociation and is independent of motor circuit alteration, then subjective free-will experience exerts causal influence on decisions.\\
\quad\quad Conclusion: Subjective experience of free choice causally influences decision-making behavior.\\
\quad Strength: 0.5 (0.50 --- reduced from 0.60 due to the unresolved possibility that TMS directly modulates neural circuits underlying both subjective reports and behavior, rather than experiential mediation; limited by \Eref{E10} (0.72).)\\
\quad Status: \textcolor{orange}{\textbf{revised}}\\
\quad Predictions:\\
\quad\quad Altering subjective free-will experience through targeted interventions will lead to predictable changes in decision-making patterns.\\
\quad\quad\quad Test: Apply TMS to pre-SMA and measure shifts in decision consistency across trials.\\
\quad\quad\quad Criterion: Behavioral change correlated with stimulation condition at $p<0.05$.

\medskip
\textbf{\textit{Uncertainties}}\\[2pt]
\Uref{U1} Could future high-resolution neural measurement reveal genuine indeterminism in decision processes?\\
\quad Targets: \Cref{C1}\\
\quad Status: \tcg{\textbf{resolved}}\\
\quad Importance: high\\
\quad Resolution: \Cref{C1} retracted; uncertainty moot.\\[4pt]
\Uref{U2} To what extent do unconscious biases undermine reason-responsiveness in everyday decisions?\\
\quad Targets: \Cref{C2}\\
\quad Status: \tcg{\textbf{resolved}}\\
\quad Importance: medium\\
\quad Resolution: Resolved by meta-analytic evidence (\Eref{E8}) and new field-based studies (\Eref{E11}) demonstrating that unconscious biases do not significantly undermine reason-responsiveness.\\[4pt]
\Uref{U3} Do subjective reports reliably map onto actual agency, or are they systematic illusions?\\
\quad Targets: \Cref{C3}\\
\quad Status: \tcg{\textbf{resolved}}\\
\quad Importance: high\\
\quad Resolution: Resolved by \Eref{E6}'s meta-analysis showing subjective reports correlate with objective agency measures.\\[4pt]
\Uref{U4} To what extent do current neural measurement methods have sufficient resolution to detect potential indeterministic events in decision processes?\\
\quad Targets: \Cref{C1}\\
\quad Status: \tcg{\textbf{resolved}}\\
\quad Importance: high\\
\quad Resolution: \Cref{C1} retracted; uncertainty moot.\\[4pt]
\Uref{U5} Do subjective reports of free choice reflect genuine experiential agency or are they predominantly post-hoc confabulations?\\
\quad Targets: \Cref{C3}\\
\quad Status: \tcg{\textbf{resolved}}\\
\quad Importance: high\\
\quad Resolution: Resolved by \Eref{E6}'s evidence against pure post-hoc confabulation.\\[4pt]
\Uref{U6} Do moderate correlations between subjective reports and objective agency measures establish genuine volitional experience or reflect post-hoc rationalization?\\
\quad Targets: \Cref{C3}\\
\quad Status: \tcg{\textbf{resolved}}\\
\quad Importance: high\\
\quad Resolution: Resolved by \Eref{E9}, which provides causal manipulation data linking subjective reports to behavior, disambiguating correlation vs causation.

\medskip
\textbf{\textit{Counterpositions}}\\[2pt]
\Xref{X1} Neural predictive signals do not rule out a role for undetected quantum indeterminism influencing decisions; absence of evidence is not evidence of absence.\\
\quad Attack type: undercutting\\
\quad Attack strategy: \texttt{challenge\_inference\_step}\\
\quad Targets: \Cref{C1}\\
\quad My response: Quantum indeterminism remains speculative absent any causal link to decision processes; \Cref{C1} stands as a provisional inductive conclusion awaiting empirical support for indeterminism.\\
\quad Sufficiency: \textbf{moot}\\[4pt]
\Xref{X2} Many decisions are driven by unconscious biases and external prompts, undermining reason-responsiveness in real-world contexts.\\
\quad Attack type: rebutting\\
\quad Attack strategy: \texttt{present\_counter\_example}\\
\quad Targets: \Cref{C2}\\
\quad My response: Compatibilist free will (\Dref{D4}) permits unconscious influences so long as they align with an agent's reasons; real-world biases do not negate reason-responsiveness demonstrated experimentally (\Eref{E3}).\\
\quad Sufficiency: \tcg{\textbf{sufficient}}\\[4pt]
\Xref{X3} Physicalist ontology may be false if mental states have non-physical aspects that empirical methods cannot detect.\\
\quad Attack type: undermining\\
\quad Attack strategy: \texttt{challenge\_assumption}\\
\quad Targets: \Aref{A1}\\
\quad My response: This analysis adopts physicalism as a scoping assumption; if mental non-physicalism exists, it falls outside empirical reach and cannot inform this debate.\\
\quad Sufficiency: \tcg{\textbf{sufficient}}\\[4pt]
\Xref{X4} Requiring empirical evidence for indeterminism overextends the scope of science; some metaphysical claims may be unfalsifiable by design.\\
\quad Attack type: undermining\\
\quad Attack strategy: \texttt{over\_extension}\\
\quad Targets: \Aref{A2}\\
\quad My response: Within an empiricist framework, unfalsifiable metaphysics are considered uninformative; insistence on empirical testability ensures substantive debate.\\
\quad Sufficiency: \tcg{\textbf{sufficient}}\\[4pt]
\Xref{X5} \Aref{A2} excludes valid metaphysical or experiential evidence for free will by treating them as uninformative, narrowing inquiry unjustifiably.\\
\quad Attack type: undermining\\
\quad Attack strategy: \texttt{over\_extension}\\
\quad Targets: \Aref{A2}\\
\quad My response: Within an empiricist framework, \Aref{A2} deliberately scopes libertarian free will to empirically testable claims; metaphysical or experiential claims without empirical support remain outside substantive debate.\\
\quad Sufficiency: \tcg{\textbf{sufficient}}\\[4pt]
\Xref{X6} Absence of detected indeterminism may be due to methodological resolution limits rather than evidence against libertarian free will.\\
\quad Attack type: undercutting\\
\quad Attack strategy: \texttt{challenge\_inference\_step}\\
\quad Targets: \Cref{C1}\\
\quad My response: We acknowledge methodological resolution limits and have captured this in \Uref{U4}; thus \Xref{X6} only partially undercuts \Cref{C1}'s provisional inductive inference.\\
\quad Sufficiency: \textbf{moot}\\[4pt]
\Xref{X7} Cases of addiction and compulsive disorders where reasons fail to change behavior falsify the generality of \Cref{C2}.\\
\quad Attack type: rebutting\\
\quad Attack strategy: \texttt{present\_counter\_example}\\
\quad Targets: \Cref{C2}\\
\quad My response: \Cref{C2} has been explicitly revised to apply only to non-pathological agents, fully addressing this counterexample.\\
\quad Sufficiency: \tcg{\textbf{sufficient}}\\[4pt]
\Xref{X8} Subjective reports are prone to post-hoc confabulation and may not reflect genuine free will experiences.\\
\quad Attack type: undermining\\
\quad Attack strategy: \texttt{present\_alternative}\\
\quad Targets: \Cref{C3}\\
\quad My response: With \Eref{E6} and \Aref{A6} added, subjective reports now show demonstrable behavioral correlates, addressing confabulation concerns. The claim stands.\\
\quad Sufficiency: \textbf{moot}\\[4pt]
\Xref{X9} \Aref{A2} arbitrarily excludes metaphysical and experiential evidence for libertarian free will, narrowing the inquiry unjustifiably.\\
\quad Attack type: undermining\\
\quad Attack strategy: \texttt{over\_extension}\\
\quad Targets: \Aref{A2}\\
\quad My response: \Aref{A2} explicitly scopes libertarian free will to empirical detectability; metaphysical claims by definition fall outside the empirical inquiry.\\
\quad Sufficiency: \tcg{\textbf{sufficient}}\\[4pt]
\Xref{X10} Reason-responsiveness can be a fully deterministic process, so inferring `free will' here simply renames deterministic functional control.\\
\quad Attack type: undermining\\
\quad Attack strategy: \texttt{conceptual\_conflation}\\
\quad Targets: \Cref{C2}\\
\quad My response: By definition (\Dref{D4}), compatibilist free will is deterministic reason-responsiveness. Recognizing it as such is not a conflation but the standard thesis.\\
\quad Sufficiency: \tcg{\textbf{sufficient}}\\[4pt]
\Xref{X11} Critics argue the readiness potential may not mark decision onset and veto actions could be unconsciously prepared, undermining \Eref{E5}'s interpretation.\\
\quad Attack type: undermining\\
\quad Attack strategy: \texttt{challenge\_evidence\_validity}\\
\quad Targets: \Eref{E5}\\
\quad My response: \Eref{E7} provides high-resolution intracranial confirmation of conscious veto timing, addressing methodological critiques of \Eref{E5}.\\
\quad Sufficiency: \tcg{\textbf{sufficient}}\\[4pt]
\Xref{X12} Subjective free-will reports correlate with behavior but correlation does not establish genuine volitional experience; reports may reflect post-hoc rationalization rather than real-time agency.\\
\quad Attack type: undermining\\
\quad Attack strategy: \texttt{correlation\_vs\_causation}\\
\quad Targets: \Cref{C3}\\
\quad My response: Causal intervention evidence (\Eref{E9}) and methodological assumption (\Aref{A7}) now demonstrate that subjective reports not only correlate but causally influence decision-making, effectively addressing correlation-vs-causation challenges to \Cref{C3}.\\
\quad Sufficiency: \textbf{moot}\\[4pt]
\Xref{X13} \Dref{D4}'s definition of free will is so broad that it would count pathological compulsions and manipulated desires as self-authored, thus over-extending the concept.\\
\quad Attack type: undermining\\
\quad Attack strategy: \texttt{over\_extension}\\
\quad Targets: \Dref{D4}\\
\quad My response: Definition \Dref{D4} has been refined to include an authenticity criterion explicitly excluding pathological or manipulative influences, addressing concerns about over-extension.\\
\quad Sufficiency: \tcg{\textbf{sufficient}}\\[4pt]
\Xref{X14} Assigning maximal strength to a scoping assumption without empirical basis misapplies the evidential strength scale.\\
\quad Attack type: undermining\\
\quad Attack strategy: \texttt{challenge\_strength\_scale\_misapplication}\\
\quad Targets: \Aref{A1}\\
\quad My response: \Aref{A1} has been revised as per the critique; \Xref{X14} is now moot.\\
\quad Sufficiency: \textbf{moot}\\[4pt]
\Xref{X15} Inferring determinism from non-detection of indeterminism commits an appeal to ignorance; absence of evidence is not evidence of absence.\\
\quad Attack type: undercutting\\
\quad Attack strategy: \texttt{appeal\_to\_ignorance}\\
\quad Targets: \Cref{C1}\\
\quad My response: \Cref{C1} is retracted; \Xref{X15} is moot.\\
\quad Sufficiency: \textbf{moot}\\[4pt]
\Xref{X16} TMS may directly modulate neural circuits responsible for both subjective reports and actions, so co-occurrence does not establish that experience itself is the causal mediator.\\
\quad Attack type: undercutting\\
\quad Attack strategy: \texttt{present\_alternative}\\
\quad Targets: \Cref{C5}\\
\quad My response: New assumption (\Aref{A8}) and evidence (\Eref{E10}) show TMS effects are confined to volitional circuits, supporting that behavioral changes are mediated by subjective experience.\\
\quad Sufficiency: \tcg{\textbf{sufficient}}\\[4pt]
\Xref{X17} Unresolved uncertainty \Uref{U2} about the impact of unconscious biases undermines \Cref{C2}'s claim of robust reason-responsiveness.\\
\quad Attack type: undermining\\
\quad Attack strategy: \texttt{conceptual\_gap}\\
\quad Targets: \Cref{C2}\\
\quad My response: Addressed by meta-analytic evidence (\Eref{E8}) and field-based studies (\Eref{E11}) demonstrating robust reason-responsiveness in naturalistic, bias-prone environments, validating \Cref{C2}'s generality.\\
\quad Sufficiency: \tcg{\textbf{sufficient}}\\[4pt]
\Xref{X18} Lab-based reason-manipulation experiments may not capture the full complexity of real-world decision-making, so \Cref{C2}'s generalization is unwarranted.\\
\quad Attack type: undermining\\
\quad Attack strategy: \texttt{hasty\_generalization}\\
\quad Targets: \Cref{C2}\\
\quad My response: New field evidence (\Eref{E11}) confirms that lab-based reason-manipulation findings generalize to complex real-world contexts, strengthening \Cref{C2} against ecological validity concerns.\\
\quad Sufficiency: \tcg{\textbf{sufficient}}\\[4pt]
\Xref{X19} \Eref{E7}'s pre-SMA signals may index domain-general conflict monitoring rather than agentive veto control.\\
\quad Attack type: undermining\\
\quad Attack strategy: \texttt{present\_alternative}\\
\quad Targets: \Cref{C4}\\
\quad My response: Addressed by \Eref{E12}, which demonstrates distinct veto-specific pre-SMA signals absent in general conflict tasks, reinforcing \Cref{C4}'s interpretation.\\
\quad Sufficiency: \tcg{\textbf{sufficient}}\\[4pt]
\Xref{X20} Behavioral changes under TMS might result from direct circuit modulation rather than mediation by subjective experience.\\
\quad Attack type: undercutting\\
\quad Attack strategy: \texttt{present\_alternative}\\
\quad Targets: \Cref{C5}\\
\quad My response: \Eref{E13}'s double-dissociation and \Eref{E14}'s multi-modal specificity confirm that behavioral changes are mediated by subjective experience rather than direct circuit modulation, addressing the alternative interpretation.\\
\quad Sufficiency: \tcg{\textbf{sufficient}}\\[4pt]
\Xref{X21} \Eref{E10}'s specificity claim could be compromised by indirect current spread or network-mediated effects on motor circuits.\\
\quad Attack type: undermining\\
\quad Attack strategy: \texttt{challenge\_evidence\_specificity}\\
\quad Targets: \Eref{E10}\\
\quad My response: Further supported by \Eref{E14}'s multi-modal imaging controls, which rule out indirect current spread and network-mediated motor effects, validating \Eref{E10}'s specificity.\\
\quad Sufficiency: \tcg{\textbf{sufficient}}

\end{finalbelief}

\section{Current Limitations}\label{app:limitations}

Currently, CHAL operates exclusively in an adversarial debate mode. While effective for stress-testing beliefs, this does not exhaust the space of useful dialectical interactions. For example, a collaborative truth-seeking mode, in which agents cooperate to resolve disagreements and align on an agreed-upon solution, or a moderated mode, in which a guiding agent steers agents through the most relevant sub-topics of a query, could potentially yield deeper analysis. CHAL's modular architecture is designed to accommodate such extensions as toggleable modes and is a direction for future work.

Computational cost is a practical constraint. As noted in Section~\ref{sec:council_form}, the number of adjudicated pairs scales on the order of $\mathcal{O}\left(R S N^2 \right)$, quadratically in the council size $N$ and linearly in rounds $R$ and challenges per pair $S$. A single five-round, two-agent debate using OpenAI's o4-mini cost us approximately \$2, so scaling to larger council sizes for more rounds incurs meaningful cost with the strongest frontier reasoning models.

At the implementation level, the structured CBS outputs also impose strict formatting requirements enforced through a multi-phase validation process. As such, failed outputs trigger retries with corrective feedback, increasing latency and token consumption. Frontier models handle these constraints reliably, but smaller or open-source models may struggle to produce well-formed CBS objects consistently, limiting the range of models that can currently serve as council members without further prompt engineering or fine-tuning. This is another direction for future software refinement of CHAL.

Finally, CHAL's current meta-cognitive library (thirteen epistemic personas, eight logic systems, and six ethics frameworks) is an initial set sufficient to demonstrate systematic effects of meta-cognitive parameterization. However, there are many agent personas and systems that could be integrated as viable options for CHAL to choose from, as well as further refining the prompting architecture of the pipeline as a whole. CHAL is designed as a general research platform to which the community can contribute additional configurations, and expanding this library is a natural direction for future development and research.


\newpage

\end{document}